\documentclass{article}
\usepackage{subcaption}
\usepackage[dvipsnames]{xcolor}
\usepackage[colorlinks,citecolor=cvprblue]{hyperref}
\usepackage{tocloft}
\usepackage{multirow}
\usepackage{hhline}
\usepackage{colortbl}
\usepackage{multicol}
\usepackage{microtype}
\usepackage{graphicx}
\usepackage{booktabs} 
\usepackage[font=small, skip=-10pt]{caption}
\usepackage{pifont}

\usepackage{listings}

\definecolor{codegreen}{rgb}{0,0.6,0}
\definecolor{codegray}{rgb}{0.5,0.5,0.5}
\definecolor{codepurple}{rgb}{0.58,0,0.82}
\definecolor{backcolour}{rgb}{0.95,0.95,0.92}

\lstdefinestyle{mystyle}{
    backgroundcolor=\color{backcolour},   
    commentstyle=\color{codegreen},
    keywordstyle=\color{magenta},
    numberstyle=\tiny\color{codegray},
    stringstyle=\color{codepurple},
    basicstyle=\ttfamily\footnotesize,
    breakatwhitespace=false,         
    breaklines=true,                 
    captionpos=b,                    
    keepspaces=true,                 
    numbers=left,                    
    numbersep=5pt,                  
    showspaces=false,                
    showstringspaces=false,
    showtabs=false,                  
    tabsize=2
}

\newcommand{\myparagraph}[1]{\vspace{0pt}\noindent{\bf #1}}
\newcommand{\nocontentsline}[3]{}
\newcommand{\tocless}[2]{\bgroup\let\addcontentsline=\nocontentsline#1{#2}\egroup}
\usepackage{multirow}
\usepackage{makecell}

\usepackage{float}
\usepackage{adjustbox}
\usepackage[accepted]{icml2024}

\usepackage{amsmath}
\usepackage{amssymb}
\usepackage{mathtools}
\usepackage{amsthm}
\usepackage{adjustbox}
\usepackage{colortbl}

\usepackage[capitalize,noabbrev]{cleveref}
\theoremstyle{plain}
\newtheorem{theorem}{Theorem}[section]

\theoremstyle{definition}

\theoremstyle{remark}

\usepackage[disable,textsize=tiny]{todonotes}
\usepackage{tcolorbox}

\icmltitlerunning{DiffuseKronA: A Parameter Efficient Fine-tuning Method for Personalized Diffusion Models}

\usepackage{color}

\begin{document}

\twocolumn[
\icmltitle{DiffuseKronA: A Parameter Efficient Fine-tuning Method for \\ Personalized Diffusion Models}



\icmlsetsymbol{equal}{*}
\begin{icmlauthorlist}
\icmlauthor{Shyam Marjit}{iiitg,equal}
\icmlauthor{Harshit Singh}{iiitg,equal}
\icmlauthor{Nityanand Mathur}{iiitg,equal}
\icmlauthor{Sayak Paul}{huggingface}
\icmlauthor{Chia-Mu Yu}{taiwan}
\icmlauthor{Pin-Yu Chen}{ibm}
\end{icmlauthorlist}
\par
\centering
Project Page: \href{https://diffusekrona.github.io/}{https://diffusekrona.github.io/}
\icmlaffiliation{iiitg}{Indian Institute of Information Technology Guwahati, India}
\icmlaffiliation{huggingface}{Hugging Face}
\icmlaffiliation{taiwan}{National Yang Ming Chiao Tung University, Hsinchu, Taiwan}
\icmlaffiliation{ibm}{IBM Research, New York, USA}

\icmlcorrespondingauthor{Shyam Marjit}{shyam.marjit@iiitg.ac.in}
\icmlcorrespondingauthor{Pin-Yu Chen}{pin-yu.chen@ibm.com}

\icmlkeywords{Machine Learning, ICML}

\vskip 0.3in
\captionsetup{type=figure}
{\begin{center}
    \vspace{-3mm}
    \centering
    \includegraphics[width=0.949\linewidth]{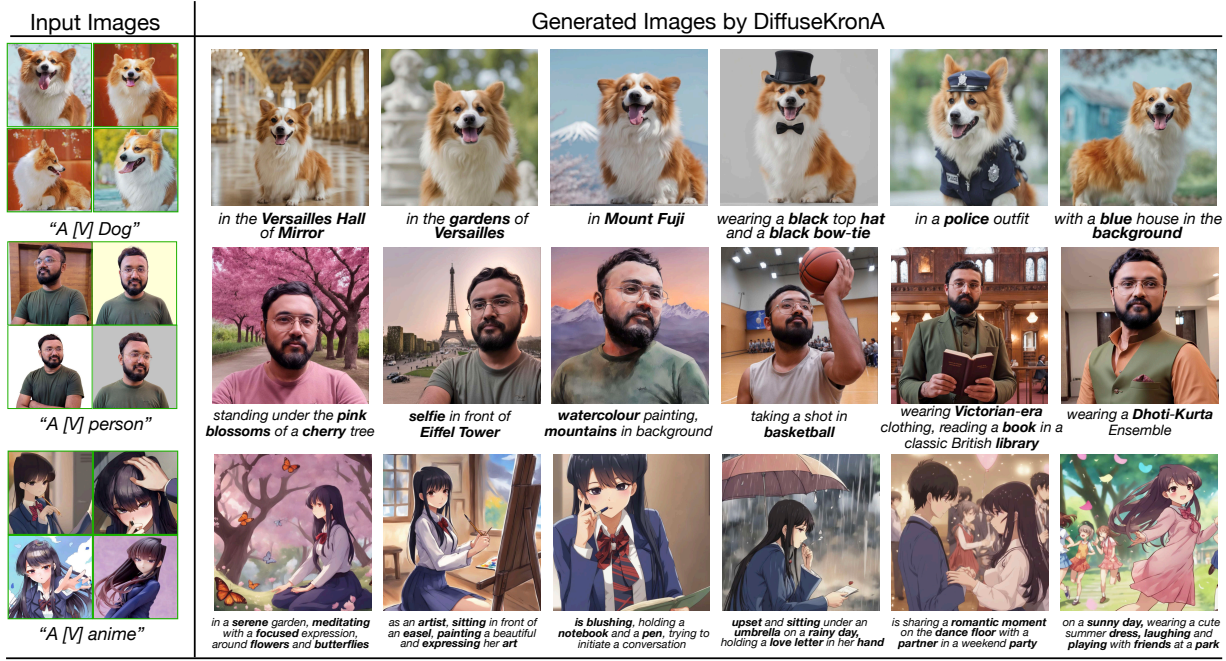}
    \vspace{-2mm}
    \captionof{figure}{\textit{\textbf{DiffuseKronA}} achieves superior image quality and text alignment across diverse input images and prompts, all the while upholding exceptional parameter efficiency. Here, \textit{[V]} denotes a unique token used for fine-tuning a specific subject in the text-to-image diffusion model. We showcase human face editing in Fig.~\ref{fig:human_face_anime}, and car modifications in Fig.~\ref{fig:car}, allowing for a wider range of applications.
    }
    \label{fig:teaser-dig}
\end{center}}
]



\printAffiliationsAndNotice{\icmlEqualContribution} 

\begin{abstract}

In the realm of subject-driven text-to-image (T2I) generative models, recent developments like DreamBooth and BLIP-Diffusion have led to impressive results yet encounter limitations due to their intensive fine-tuning demands and substantial parameter requirements. While the low-rank adaptation (LoRA) module within DreamBooth offers a reduction in trainable parameters, it introduces a pronounced sensitivity to hyperparameters, leading to a compromise between parameter efficiency and the quality of T2I personalized image synthesis. Addressing these constraints, we introduce \textbf{\textit{DiffuseKronA}}, a novel Kronecker product-based adaptation module that not only significantly reduces the parameter count by 35\% and 99.947\% compared to LoRA-DreamBooth and the original DreamBooth, respectively, but also enhances the quality of image synthesis. Crucially, \textit{DiffuseKronA} mitigates the issue of hyperparameter sensitivity, delivering consistent high-quality generations across a wide range of hyperparameters, thereby diminishing the necessity for extensive fine-tuning. Furthermore, a more controllable decomposition makes \textit{DiffuseKronA} more interpretable and even can achieve up to a 50\% reduction with results comparable to LoRA-Dreambooth.
Evaluated against diverse and complex input images and text prompts, \textit{DiffuseKronA} consistently outperforms existing models, producing diverse images of higher quality with improved fidelity and a more accurate color distribution of objects, all the while upholding exceptional parameter efficiency, thus presenting a substantial advancement in the field of T2I generative modeling. 

\end{abstract}


\vspace{-5.5mm}
\tocless{
\section{Introduction}
\label{sec: intro}
}

In recent years, text-to-image (T2I) generation models~\cite{gu2022vector, chang2023muse,rombach2022ldm, sdxl, yu2022scaling} have rapidly evolved, generating intricate and highly detailed images that often defy discernment from real-world photographs. The current state-of-the-art has marked significant progress and demonstrated substantial improvement, which hints at a future where the boundary between human imagination and computational representation becomes increasingly blurred. In this context, subject-driven T2I generative models~\cite{dreambooth, blip_diffusion} unlock creative potential such as image editing, subject-specific property modifications, art renditions, etc. Works like DreamBooth ~\cite{dreambooth}, BLIP-Diffusion~\cite{blip_diffusion} seamlessly introduce new subjects into the pre-trained models while preserving the priors learned by the original model without impacting its generation capabilities. These approaches excel at retaining the essence and subject-specific details across various styles when fine-tuned with few-shot examples, leveraging foundational pre-trained latent diffusion models (LDMs)~\cite{rombach2022ldm}.

However, DreamBooth with Stable Diffusion~\cite{rombach2022ldm} suffers from some primary issues, such as incorrect prompt context synthesis, context appearance entanglement, and hyperparameter sensitivity. 
Additionally, DreamBooth finetunes all parameters of latent diffusion model's~\cite{rombach2022ldm} UNet and text encoder~\cite{clip}, which significantly increases the trainable parameter count, making the finetuning process expensive. Here, the widely used low-rank adaptation module~\cite{lora} (LoRA) within DreamBooth attempts to significantly trim the parameter counts but it magnifies the aforementioned DreamBooth-reported issues, which makes a complete tradeoff between parameter efficiency and satisfactory subject-driven image synthesis. Moreover, it suffers from high sensitivity to hyperparameters, necessitating extensive fine-tuning to achieve desired outputs. This motivates us to design a more robust and effective parameter-efficient fine-tuning (PEFT) method for adapting T2I generative models to subject-driven personalized generation. 

\begin{figure}[h]
    \centering
    \includegraphics[width=0.47\textwidth]{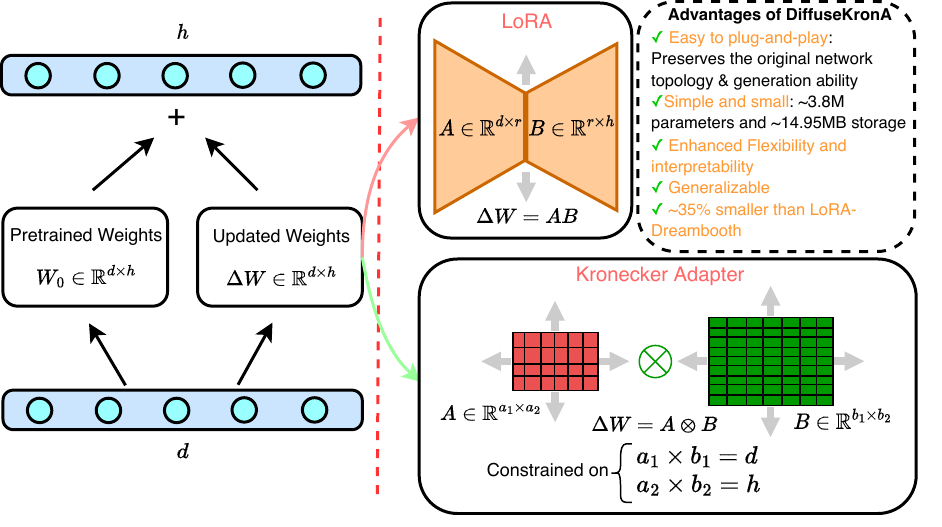}
     \vspace{-5mm}
    \caption{Schematic illustration: LoRA is limited to one controllable parameter, the rank $\mathbf{r}$; while the Kronecker product showcases enhanced interpretability by introducing two controllable parameters $\mathbf{a_1}$ and $\mathbf{a_2}$ (or equivalently $\mathbf{b_1}$ and $\mathbf{b_2}$).}
    \label{fig:interpretability}
     \vspace{+2mm}
\end{figure}

In this paper, we introduce \textit{\textbf{DiffuseKronA}}, a novel parameter-efficient module that leverages the Kronecker product-based adaptation module for fine-tuning T2I diffusion models, focusing on few-shot adaptations. LoRA adheres to a vanilla encoder-decoder type architecture, which learns similar representations within decomposed matrices due to constrained flexibility and similar-sized matrix decomposition~\cite{tahaei2022kroneckerbert}.  In contrast, Kronecker's decomposition exploits patch-specific redundancies, offering a much higher-rank approximation of the original weight matrix with less parameter count and greater flexibility in representation by allowing different-sized decomposed matrices. This fundamental difference is attributed to several improvements including parameter reduction, enhanced stability, and greater flexibility. Moreover, it effectively captures crucial subject-specific spatial features while producing images that closely adhere to the provided prompts. This results in higher quality, improved fidelity, and more accurate color distribution in objects during personalized image generation, achieving comparable results to state-of-the-art techniques.

Our \underline{key contributions} are as follows:\\
\ding{182} \textbf{Parameter Efficiency:}
\textit{DiffuseKronA} significantly reduces trainable parameters by 35\% and 99.947\% as compared to LoRA-DreamBooth and vanilla DreamBooth using SDXL~\cite{sdxl} as detailed in~\cref{tab:model_performance}. By changing Kronecker factors, we can even achieve up to a 50\% reduction with results comparable to state-of-the-art as demonstrated in ~\cref{fig:comp_lora_lokr_loha} in the Appendix.\\
\ding{183} \textbf{Enhanced Stability:} \textit{DiffuseKronA} offers a much more stable image-generation process formed within a fixed spectrum of hyperparameters when fine-tuning, even when working with complicated input images and diverse prompts. In \cref{fig:comparison_stability}, we demonstrate the trends associated with hyperparameter changes in both methods and highlight our superior stability over LoRA-DreamBooth.\\
\ding{184} \textbf{Text Alignment and Fidelity:} On average, \textit{DiffusekronA} captures better subject semantics and large contextual prompts. We refer the readers to ~\cref{fig:comparison_sota} and ~\cref{fig:comparison_sota_quant} for qualitative and quantitative comparisons, respectively.\\ 
\ding{185} \textbf{Interpretablilty:} Notably, we conduct extensive analysis to explore the advantages of the Kronecker product-based adaptation module within personalized diffusion models. More controllable decomposition makes \textit{DiffusekronA} more interpretable as demonstrated in \cref{fig:interpretability}.\\
Extensive experiments on 42 datasets under the few-shot setting demonstrate the aforementioned effectiveness of \textit{DiffuseKronA}, achieving the best trade-off between parameter efficiency and satisfactory image synthesis.

\tocless{
\section{Related Works}
\label{sec: related_work}
}

\myparagraph{Text-to-Image Diffusion Models.} Recent advancements in T2I diffusion models such as Stable Diffusion (SD)~\cite{rombach2022ldm, sdxl}, Imagen~\cite{imagen}, DALL-E2~\cite{ramesh2022dall.e2} \& E3~\cite{betker2023improving}, PixArt-$\alpha$~\cite{chen2023pixartalpha}, Kandinsky~\cite{KANDINSKY}, and eDiff-I~\cite{balaji2022ediffi} have showcased remarkable efficacy in modeling data distributions, yielding impressive results in image synthesis and opening the door for various creative applications across domains. 
Compared to the previous iterations of the SD model, Stable Diffusion XL (SDXL)~\cite{sdxl} represents a significant advancement in T2I synthesis owing to a larger backbone and an improved training procedure.
In this work, we mainly incorporate SDXL due to its impressive capability to generate high-resolution images, prompt adherence, as well as better composition and semantics.

\myparagraph{Subject-driven T2I Personalization.} Given only a few images (typically 3 to 5) of a specific subject, T2I personalization techniques aim to synthesize diverse contextual images of the subject based on textual input. In particular, Textual Inversion~\cite{gal2022textual} and DreamBooth~\cite{dreambooth} were the first lines of work. Textual Inversion fine-tunes text embedding, while DreamBooth fine-tunes the entire network using an additional preservation loss as regularization, resulting in visual quality improvements that show promising outcomes. More recently, BLIP-Diffusion~\cite{blip_diffusion} enables zero-shot subject-driven generation capabilities by performing a two-stage pre-training process leveraging the multimodal BLIP-2~\cite{li2023blip} model.
These studies focus on single-subject generation, with later works~\cite{custom_diffusion, han2023svdiff, ma2023subject, tewel2023key} delving into multi-subject generation. 

\myparagraph{PEFT Methods within T2I Personalization.}
In contrast to foundational models~\cite{dreambooth, blip_diffusion} that fine-tune large pre-trained models at full scale, several seminal works~\cite{custom_diffusion, han2023svdiff, ruiz2023hyperdreambooth, ye2023ip-adapter} in parameter-efficient fine-tuning (PEFT) have emerged as a transformative approach. Within the realm of PEFT techniques, low-rank adaptation methods~\cite{lora, von_Platen_Diffusers_State-of-the-art_diffusion} has become a de-facto way of reducing the parameter count by introducing learnable truncated Singular Value Decomposition (SVD) modules into the original model weights on essential layers. For instance, Custom Diffusion~\cite{custom_diffusion} focuses on fine-tuning the $K$ and $V$ matrices of the cross-attention, introducing multiple concept generation for the first time, and employing LoRA for efficient parameter compression. SVDiff~\cite{han2023svdiff} achieves parameter efficiency by fine-tuning the singular values of the weight matrices with a Cut-Mix-Unmix data augmentation technique to enhance the quality of multi-subject image generation. Hyper-Dreambooth~\cite{ruiz2023hyperdreambooth} proposed a hypernetwork to make DreamBooth rapid and memory-efficient for personalized fidelity-controlled face generation. T2I-Adapters~\cite{mou2023t2i}, a conceptually similar approach to ControlNets~\cite{zhang2023adding}, makes use of an auxiliary network to compute the representations of the additional inputs and mixes that with the activations of the UNet. Mix-of-Show~\cite{gu2023mix}, on the other hand, involves training distinct LoRA models for each subject and subsequently performing fusion.

In context, the LoRA-Dreambooth~\cite{cloneofsimo} technique has encountered difficulties due to its poor representational capacity and low interpretability, and to address these constraints we introduce \textit{DiffuseKronA}. Our method is inspired by the KronA technique initially proposed by~\cite{krona}. However, there are key distinctions: (\textbf{1}) The original paper was centered around language models, whereas our work extends this exploration to LDMs, particularly in the context of T2I generation. (\textbf{2}) Our focus lies on the efficient fine-tuning of various modules within LDMs. (\textbf{3}) More importantly, we investigate the impact of altering Kronecker factors on subject-specific generation, considering interpretability, parameter efficiency, and subject fidelity.
It is also noteworthy to mention that LoKr~\cite{lokr} is a concurrent work, and we discuss the key differences in~\cref{sec:low-rank-other}.




\tocless{
\section{Methodology}
\label{sec: methods}
}

\begin{figure*}[!ht]
  \centering
  \includegraphics[width=0.98\linewidth]{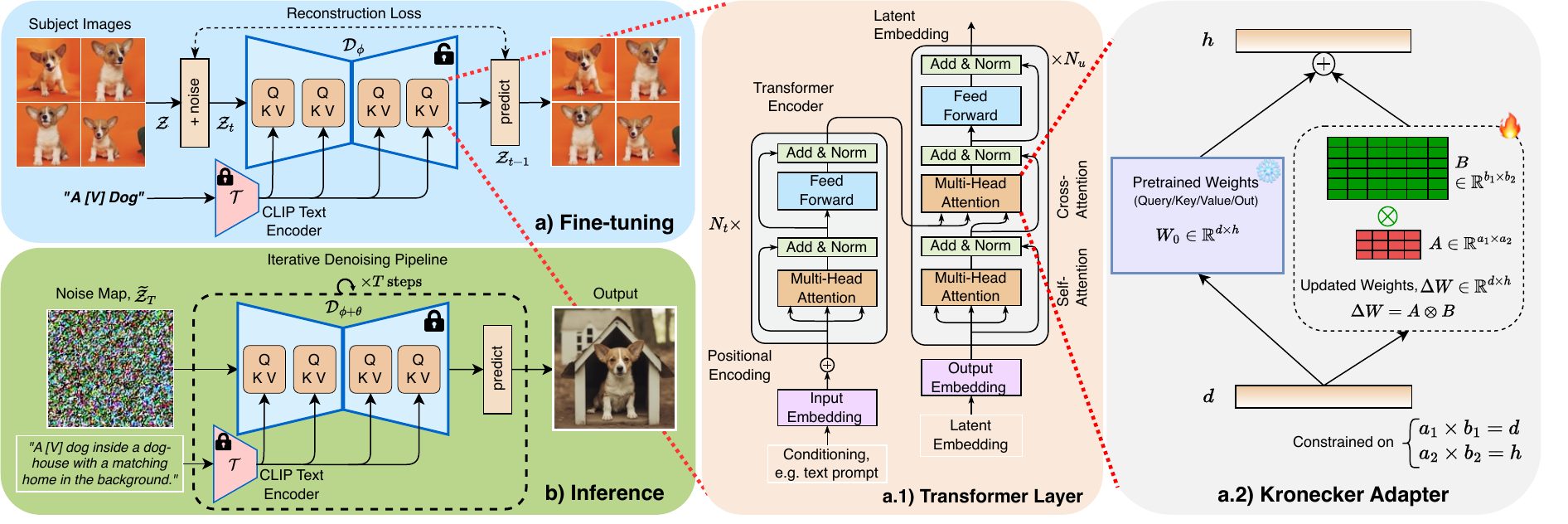}
  \vspace{-4mm}
  \caption{Overview of \textbf{\textit{DiffuseKronA}}: (a) Fine-tuning process involves optimizing the multi-head attention parameters (a.1) using Kronecker Adapter, elaborated in the subsequent block, a.2, (b) During inference, newly trained parameters, denoted as $\theta$, are integrated with the original weights $\mathcal{D}_{\phi}$ and images are synthesized using the updated personalized model $\mathcal{D}_{\phi+\theta}$.}
  \label{fig:model_diagram} \vspace{-3mm} 
\end{figure*}

\myparagraph{Problem Formulation.} Given a pre-trained T2I latent diffusion model $\mathcal{D}_{\phi}$ with size $|\mathcal{D}_{\phi}|$ and weights denoted by $\phi$, we aim to develop a parameter-efficient adaptation technique with trainable parameters $\theta$ of size $m$ such that $m \ll |\mathcal{D}_{\phi}|$ holds (\emph{i.e.} efficiency) while attaining satisfactory and comparable performance with a full fine-tuned model. At inference, newly trained parameters will be integrated with their corresponding original weight matrix, and diverse images can be synthesized from the new personalized model, $\mathcal{D}_{\phi + \theta}$.

\myparagraph{Method Overview.} \cref{fig:model_diagram} shows an overview of our proposed  \textbf{\textit{DiffuseKronA}} for PEFT of T2I diffusion models in subject-driven generation. \textit{DiffuseKronA} only updates parameters in the attention layers of the UNet model while keeping text encoder weights frozen within the SDXL backbone. Here, we first outline a preliminary section in ~\cref{ssec:prelim} followed by a detailed explanation of \textit{DiffuseKronA} in ~\cref{ssec:DiffuseKronA}. Particularly, in ~\cref{ssec:DiffuseKronA}, we provide insights and mathematical explanations of 
\textit{``Why DiffuseKronA is a more parameter-efficient and interpretable way of fine-tuning Diffusion models compared to vanilla LoRA?''}

\tocless{
\subsection{Preliminaries}
\label{ssec:prelim}
}

\myparagraph{T2I Diffusion Models.}
\label{ssec:T2I_Diffusion_Models} LDMs~\cite{rombach2022ldm}, a prominent variant of probabilistic generative Diffusion models denoted as $\mathcal{D}_\phi$, aim to produce an image $\mathbf{x}_{gen} = \mathcal{D}_\phi\left(\boldsymbol{\epsilon}, \mathbf{c} \right)$ by incorporating a noise map $\boldsymbol{\epsilon} \sim \mathcal{N}(\mathbf{0}, \mathbf{I})$ and a conditioning embedding $\mathbf{c} = \mathcal{T}\left(\mathbf{P}\right)$ derived from a text prompt $\mathbf{P}$ using a text encoder, $\mathcal{T}$. LDMs transform the input image $\mathbf{x} \in \mathbb{R}^{H\times W \times 3}$ into a latent representation $\mathbf{z} \in \mathbb{R}^{h\times w \times v}$ through an encoder $\mathcal{E}$, where $\mathbf{z} = \mathcal{E}\left(\mathbf{x}\right)$ and $v$ is the latent feature dimension. In this context, the denoising diffusion process occurs in the latent space, $\mathcal{Z}$, utilizing a conditional UNet~\cite{ronneberger2015u} denoiser $\mathcal{D}_\phi$ to predict noise $\boldsymbol{\epsilon}$ at the current timestep $t$ given the noisy latent $\mathbf{z}_t$ and generation condition $c$. In brief, the denoising training objective of an LDM $\mathcal{D}_\phi$ can be simplified to:
\vspace{-3mm}
\begin{equation}
    \mathbb{E}_{\mathcal{E}\left(\mathbf{x}\right), \mathbf{c}, \boldsymbol{\epsilon} \sim \mathcal{N}\left(\mathbf{0}, \mathbf{I}\right), t \sim \mathcal{U}\left(0, 1\right)}\left[w_t\left\|\mathcal{D}_{\phi} \left(\mathbf{z}_t|\boldsymbol{c}, t\right)-\boldsymbol{\epsilon}\right\|_2^2\right],
\vspace{-2mm}
\end{equation}%
where $\mathcal{U}$ denotes uniform distribution and $w_t$ is a time-dependent weight on the loss.


\myparagraph{Low Rank Adaptation (LoRA).}
\label{ssec:LoRA}
Pre-trained large models exhibit a low ``intrinsic dimension" for task adaptation~\cite{lora, han2023svdiff}, implying efficient learning after subspace projection. Based on this, LoRA~\cite{lora} hypothesizes that weight updates also possess low ``intrinsic rank" during adaptation and inject trainable rank decomposition matrices into essential layers of the model for task adaptations, significantly reducing the number of trainable parameters.\\
In the context of a pre-trained weight matrix \(W_{0} \in \mathbb{R}^{d\times k}\), the update of \(W_{0}\) is subject to constraints imposed through the representation of the matrix as a low-rank decomposition \(W_{0} + \Delta W := W_{0} + AB\), where \(A \in \mathbb{R}^{d\times r}\), \(B \in \mathbb{R}^{r\times h}\), and the rank \(r \ll \min(d, h)\). As a result, the sizes of $A$ and $B$ are significantly smaller than $W_0$, reducing the number of trainable parameters. Throughout the training process, \(W_0\) remains fixed, impervious to gradient updates, while the trainable parameters are contained within \(A\) and \(B\). 
For \(h = W_0x\), the modified forward pass is formulated as follows:

\vspace{-6mm}
\begin{equation}
    f(x)=W_0 x+\Delta W x+b_0:=W_{\text {LoRA }} x+b_0,
    \vspace{-1mm}
\end{equation}
where $b_0$ is the bias term of the pre-trained model.
\myparagraph{LoRA-DreamBooth.}
\label{ssec:LoRA-DreamBooth}
LoRA~\cite{lora} is strategically employed to fine-tune DreamBooth with the primary purpose of reducing the number of trainable parameters. LoRA injects trainable modules with low-rank decomposed matrices in $W_{Q}$, $W_{K}$, $W_{V}$, and $W_{O}$ weight matrices of attention modules within the UNet and text encoder. During training, the weights of the pre-trained UNet and text encoder are frozen and only LoRA modules are tuned. However, during inference, the weights of fine-tuned LoRA modules are annexed to the corresponding pre-trained weights. Moreover, this task does not increase the inference time.


\vspace{2mm}
\tocless{
\subsection{\textit{DiffuseKronA}}
\label{ssec:DiffuseKronA}
}

LoRA demonstrates effectiveness in the realm of diffusion models but is hindered by its limited representation power. In contrast, the Kronecker product offers a more nuanced representation by explicitly capturing pairwise interactions between elements of two matrices. This ability to capture intricate relationships enables the model to learn and represent complex patterns in the data with greater detail. 

\myparagraph{Kronecker Product ($\otimes$)} is a matrix multiplication method that allows multiplication between matrices of different shapes. For two matrices $A \in \mathbb{R}^{a_1 \times a_2}$ and $B \in \mathbb{R}^{b_1 \times b_2}$, each block  of their Kronecker product $A \otimes B \in \mathbb{R}^{a_{1}b_{1} \times a_{2}b_{2}}$ is defined by multiplying the entry $A_{i,j}$ with $B$ such that 
\vspace{-2mm}
\begin{equation}
{ \small
    A \otimes B=\left[\begin{array}{ccc}
    a_{1,1} B & \cdots & a_{1,a_2} B \\
    \vdots & \ddots & \vdots \\
    a_{a_1, 1} B & \cdots & a_{a_1, a_2} B
    \end{array}\right]
    }.
\end{equation}
The Kronecker product can be used to create matrices that represent the relationships between different sets of model parameters. These matrices encode how changes in one set of parameters affect or interact with another set. In \cref{fig:how_krona_works}, we showcase \textit{how Kronecker product works}.
Interestingly, it does not suffer from rank deficiency as low-rank down-projection does, as in the case of techniques such as LoRA and Adapter.
The Kronecker product has several advantageous properties that make it a good option for handling complex data \cite{greenewald2015robust}. 

\myparagraph{Kronecker Adapter (KronA).} Firstly introduced in studying PEFT of language models \cite{krona},
The Kronecker product takes advantage of the structured relationships encoded in the matrices.
Instead of explicitly performing all the multiplications required to compute the product $A \otimes B$, the following equivalent matrix-vector multiplication can be applied, reducing the overall computational cost. This is particularly beneficial when working with large matrices or when computational resources are constrained:


\vspace{-5mm}
\begin{equation}
    (A \otimes B) x=\gamma\left(B \eta_{b_2 \times a_2}(x) A^{\top}\right)
\end{equation}
where  $A^{\top}$ is transposed to $A$. The rationale is that a vector $y \in \mathbb{R}^{m \cdot n}$ can be reshaped into a matrix $Y$ of size $m \times n$ using the mathematical operation $\eta_{m \times n}(\mathbf{y})$. Similarly, $Y \in \mathbb{R}^{m \times n}$ can also be transformed back into a vector by stacking its columns using the $\gamma(Y)$ operation. This approach achieves $\mathcal{O}(b\log{}b)$ computational complexity and $\mathcal{O}(\log{}b)$ space complexity for a $b$-dimensional vector, a drastic improvement over the standard unstructured Kronecker multiplication \cite{KroneckerICCV}.

\myparagraph{Fine-Tuning Diffusion Models with KronA.} In essence, KronA can be applied to any subset of weight matrices in a neural network for parameter-efficient adaptation as specified in the equation below, where $U$ denotes different modules in diffusion models, including Key ($K$), Query ($Q$), Value ($V$), and Linear ($O$) layers. During fine-tuning, KronA modules are applied in parallel to the pre-trained weight matrices. The Kronecker factors are multiplied, scaled, and merged into the original weight matrix after they have been adjusted. Hence, like LoRA, KronA maintains the same inference time.
\vspace{-2mm}

\begin{equation}
\centering
\begin{aligned}
\Delta W^{U} =A^U \otimes B^U,
 U \in \{K, Q, V, O\};\\
 W_{\text{fine-tuned}}=W_{\text{pre-trained}}+\Delta W. \\
\end{aligned}
\end{equation}

Previous studies~\cite{custom_diffusion, von_Platen_Diffusers_State-of-the-art_diffusion, tewel2023key} have conducted extensive experiments to identify the most influential modules in the fine-tuning process. In \cite{custom_diffusion, 10.5555/3495724.3497056}, authors explored the rate of changes in each module during fine-tuning on different datasets, denoted as $\delta_l=\left\|\theta_l^{\prime}-\theta_l\right\| /\left\|\theta_l\right\|$, where $\theta_l^{\prime}$ and $\theta_l$ represent the updated and pre-trained model parameters of layer $l$. Their findings indicated that the cross-attention module exhibited a relatively higher $\delta$, signifying its pivotal role in the fine-tuning process. In light of these studies, we conducted fine-tuning on the attention layers and observed their high effectiveness. Additional details on this topic are available in~\cref{sec:supple_ablation}.

\myparagraph{A closer look at LoRA v.s. \textit{DiffuseKronA}.} Higher-rank matrices are decomposable to a higher number of singular vectors, capturing better expressibility and allowing for a richer capacity for PEFT. In LoRA, the rank of the resultant update matrix $\Delta W_{\text{lora}}$ is bounded by the minimum rank between matrices $A$ and $B$, \emph{i.e.} $rank(\Delta W_{\text{lora}}) = \min(rank(A), rank(B))$. Conversely, in \textit{DiffuseKronA}, the matrix rank $\Delta W_{\text{KronA}} = A \otimes B$ is the product of the ranks of matrices $A$ and $B$, \emph{i.e.}
$ rank(\Delta W_{\text{KronA}}) = rank(A) \cdot rank(B)$, which can be properly configured to produce a higher-rank matrix than LoRA while maintaining lower-rank decomposed matrices than LoRA. Hence, for personalized T2I diffusion models, \textit{DiffuseKronA} is expected to carry more subject-specific information in lesser parameters, as compared in~\cref{tab:model_performance} and~\cref{tab:sdxl_comp}. More details are provided in~\cref{sec:detailed_lora_vs_our}.


\begin{table}[]
    \centering
    \resizebox{0.475\textwidth}{!}{%
    \begin{tabular}{c|ccccc}
    \toprule 
    \multirow{2}{*}{\makecell{\textbf{Decomposed Matrix}\\\textbf{Factor Name}}} & \multirow{2}{*}{\textbf{Notation}} & \multirow{2}{*}{\makecell{\textbf{Module}\\\textbf{Parameters}}} & \multirow{2}{*}{\makecell{\textbf{Factorization} \\\textbf{Constraint}}} \\
    &&&\\ \midrule
    Kronecker down factor & $A \in \mathbb{R}^{a_1 \times a_2}$ & \multirow{2}{*}{$a_1 a_2+b_1 b_2$} & $a_1 b_1=d$ \\
    Kronecker up factor & $B \in \mathbb{R}^{b_1 \times b_2}$ && $a_2 b_2=h$ \\
    \midrule LoRA down projection & $A\in\mathbb{R}^{d \times r}$ & \multirow{2}{*}{$r(d + h)$} &\multirow{2}{*}{$r \ll  \min(d,h)$ } \\
    LoRA up projection & $B \in \mathbb{R}^{r \times h}$ && \\
    \bottomrule
    \end{tabular}}
    \vspace*{-8.5pt}
    \caption{Comparing Kronecker factors and LoRA projections.}
    \vspace{1.5mm}
    \label{tab:comparison}
\end{table}

\tocless{
\section{Experiments}
\label{sec: experiments}
}

\begin{figure*}[!ht]
    \centering
    \includegraphics[width=0.85\textwidth]{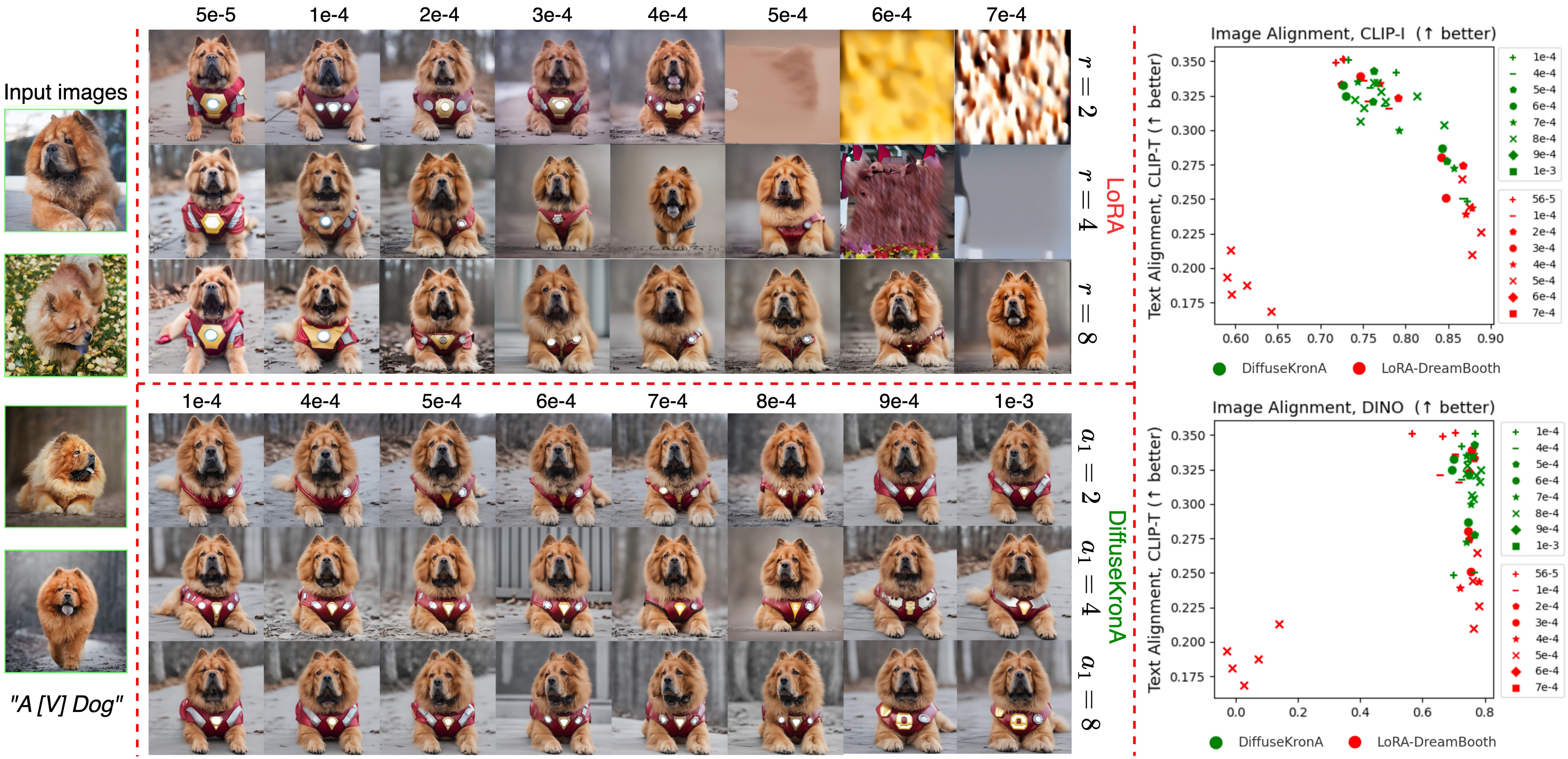}
    \vspace*{-12pt}
    \caption{Comparison between \textit{DiffuseKronA} and LoRA-DreamBooth across varying learning rates on SDXL. In our approach, we set the value of $a_2$ to 64. \textit{DiffuseKronA} produces favorable results across a wider range of learning rates, specifically from $1\times\mathrm{10}^{-4}$ to $1\times\mathrm{10}^{-3}$. In contrast, no discernible patterns are observed in LoRA. The right part of the figure shows plots of Text \& Image Alignment for \textcolor{red}{LoRA-DreamBooth} and \textcolor{green}{\textit{DiffuseKronA}}, where points belonging to \textit{DiffuseKronA} seem to be dense and those of LoRA-DreamBooth seems to be sparse, signifying that \textit{DiffuseKronA} tends to be more \textit{stable} than LoRA-DreamBooth while changing learning rates.}
    \vspace*{-10pt}
  \label{fig:comparison_stability}
\end{figure*}

In this section, we assess the various components of personalization using \textit{DiffuseKronA} through a comprehensive ablation study to confirm their effectiveness, using SDXL~\cite{von_Platen_Diffusers_State-of-the-art_diffusion} and SD~~\cite{compvis} models as backbones. Furthermore, we have conducted an insightful comparison between \textit{DiffuseKronA} and LoRA-DreamBooth in six aspects in \cref{ssec:lora_vs_krona} and also compare \textit{DiffuseKronA} with other related prior works in~\cref{ssec:sota_comp}, highlighting our superiority.

\vspace{1.5mm}
\tocless{
\subsection{Datasets and Evaluation}
\label{sec: dataset_eval}
}

\myparagraph{Datasets.} We have performed extensive experimentation on four types of subject-specific datasets: (i) 12 datasets (9 are from~\cite{dreambooth} and 3 are from~\cite{custom_diffusion}) of living subjects/pets such as stuffed animals, dogs, and cats; (ii) dataset of 21 unique objects including sunglasses, backpacks, etc.; 
(iii) our 5 collected datasets on cartoon characters including Super-Saiyan, Akimi, Kiriko, Shoko Komi, and Hatake Kakashi; (iv) our 4 collected datasets on facial images. More details are given in~\cref{sec:dataset_appendix}.

\myparagraph{Implementation Details.}
We observe that $\sim$ 1000 iterations, employing a learning rate of $5 \times 10^{-4}$, and utilizing an average of 3 training images prove sufficient for generating desirable results. The training process takes $\sim$ 5 minutes for SD~\cite{compvis} and $\sim$ 40 minutes for SDXL~\cite{von_Platen_Diffusers_State-of-the-art_diffusion} on a 24GB NVIDIA RTX-3090 GPU.

\myparagraph{Evaluation metrics.} We evaluate \textit{DiffuseKronA} on (1) \textit{Image-alignment}: we compute the CLIP~\cite{clip} visual similarity (CLIP-I) and DINO~\cite{dino} similarity scores of generated images with the reference concept images, and (2) \textit{Text-alignment}: we quantify the CLIP text-image similarity (CLIP-T) between the generated images and the provided textual prompts. A detailed mathematical explanations are available in~\cref{sec:evaluation_metrics_appendix}.

\vspace{2mm}
\tocless{
\subsection{Unlocking the Optimal Configurations of \textit{DiffuseKronA}}
\label{ssec:optimal_conf} 
}

\hypertarget{4.2}{Throughout} our experimentation, we observed the following trends and found the optimal configuration of hyper-parameters for better image synthesis using \textit{DiffuseKronA}. 


\myparagraph{How to perform Kronecker decomposition?} 
Unlike LoRA, \textit{DiffuseKronA} features two controllable Kronecker factors, as illustrated in Table~\ref{tab:comparison}, providing greater flexibility in decomposition. Our findings reveal that the dimensions of the downward Kronecker matrix $\mathbf{A}$ must be smaller than those of the upward Kronecker matrix $\textbf{B}$. Specifically, we determined the optimal value of $a_2$ to be precisely 64, while $a_1$ falls within the set $\left\{2, 4, 8\right\}$. Remarkably, among all pairs of $\left(a_1, a_2\right)$ values, $\left(4, 64\right)$ yields images with the highest fidelity. Additionally, it has been observed that images exhibit minimal variation with learning rates when $a_2 = 64$, as depicted in~\cref{fig:comparison_stability} and~\cref{fig:kronecker_factors}. Detailed ablation about Kronecker factors, their initializations, and their impact on fine-tuning is provided in~\cref{ssec:Effect_Kronecker_Factors}.

\myparagraph{Effect of learning rate.} \textit{DiffuseKronA} produces consistent results across a wide range of learning rates. Here, we observed that the images generated for a learning rate closer to the optimal learning rate value $5\times\mathrm{10}^{-4}$ generate similar images. However, learning rates exceeding $1\times\mathrm{10}^{-3}$ contribute to model overfitting, resulting in high-fidelity images but with diminished emphasis on input text prompts. Conversely, learning rates below $1\times\mathrm{10}^{-4}$ lead to lower fidelity in generated images, prioritizing input text prompts to a greater extent. 
This pattern is evident in~\cref{fig:comparison_stability}, where our approach produces exceptional images that faithfully capture both the input image and the input text prompt. Additional results are provided in~\cref{ssec:effect_learning_rate} to justify the same.

Additionally, we conducted investigations into model ablations, examining (a) choice of modules to fine-tune the model in~\cref{ssec:without_with_MLP} (b) effects of no training images in~\cref{ssec:effect_training_images} and steps in~\cref{ssec:effect_training_steps}, (c) one-shot model performance in~\cref{sssec:one_shot_performance}, and (d) effect of inference hyperparameters such as the number of inference steps and the guidance score in~\cref{ssec:effect_inference_hyperparameters}.

\begin{table}[t]
\vspace{-2mm}  
\centering
\resizebox{0.475\textwidth}{!}{%
    \begin{tabular}{cc|c|c|c}
        \toprule
        & \textsc{Model} & \textsc{Train. Time} ($\downarrow$) & \textsc{\# Param} ($\downarrow$) & \textsc{Model size} ($\downarrow$)\\
        \midrule
        \multirow{2}{*}{\small \rotatebox{90}{\textbf{SDXL}}}
        
        & \textit{LoRA-DreamBooth} & \textbf{$\sim$ 38 min.} & 5.8 M & 22.32 MB\\

        & \cellcolor{gray!20}\textit{DiffuseKronA} & \cellcolor{gray!20}$\sim$ 40 min. & \cellcolor{gray!20}\textbf{3.8 M} & \cellcolor{gray!20}\textbf{14.95 MB}\\
        
        \midrule
        
        \multirow{2}{*}{\small \rotatebox{90}{\textbf{SD}}}
        
        & LoRA-Dreambooth & $\sim$ \textbf{5.3 min.} & 1.09 M & 4.3 MB\\

        & \cellcolor{gray!20}\textit{DiffuseKronA} & \cellcolor{gray!20}$\sim$ 5.52 min. & \cellcolor{gray!20}\textbf{0.52 M} & \cellcolor{gray!20}\textbf{2.1MB} \\
        
        \bottomrule
    \end{tabular}
}
\vspace*{-7.5pt}
\caption{Exploring model efficiency metrics (\textit{DiffuseKronA} variant used ($a_1=4$ and $a_2=64$).}
\vspace{1.5mm}
\label{tab:model_performance}
\end{table}

\vspace{2mm}
\tocless{
\subsection{Exploring Model Performance: LoRA-Dreambooth vs \textit{DiffuseKronA}}
\label{ssec:lora_vs_krona}
}

\begin{figure*}[!ht]
\centering
\begin{minipage}{.5\textwidth}
  \centering
  \includegraphics[width=0.83\linewidth]{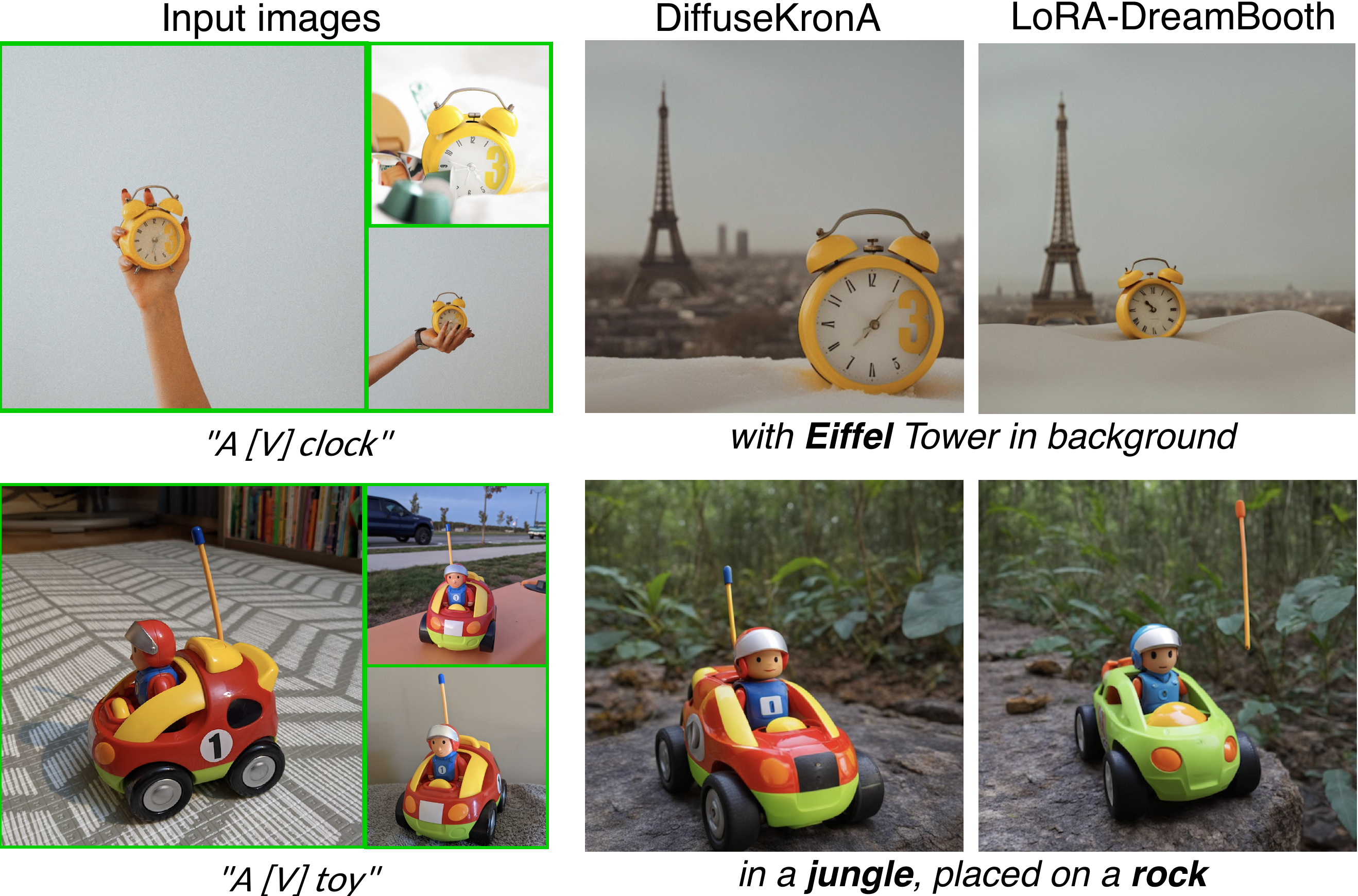}
  \vspace{-3mm}
  \captionof{figure}{\textit{DiffuseKronA} preserving superior fidelity.}
  \label{fig:comparison_fidelity}
\end{minipage}%
\begin{minipage}{.5\textwidth}
  \centering
  \includegraphics[width=0.78\linewidth]{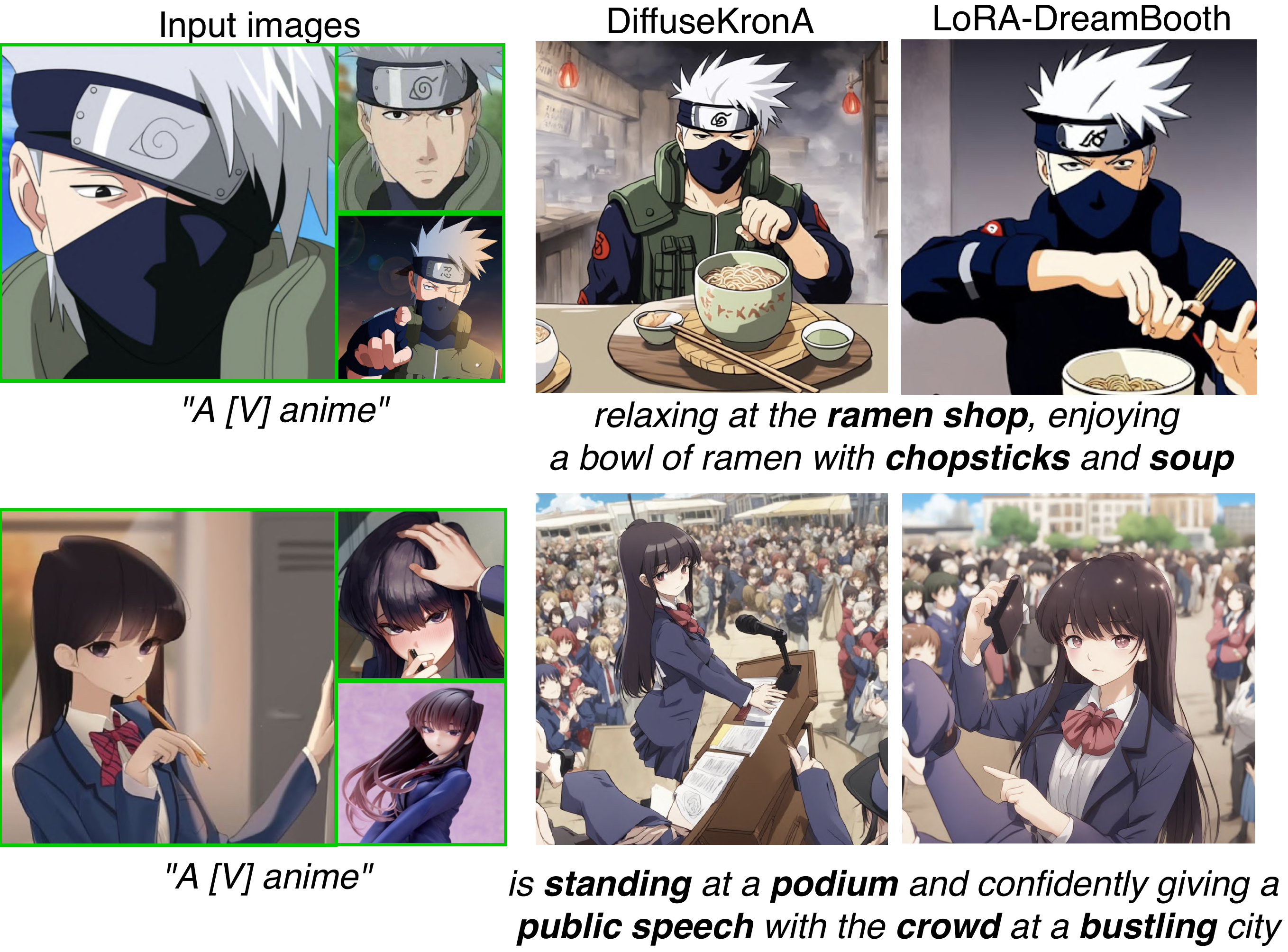}
    \vspace{-4mm}
  \captionof{figure}{\textit{DiffuseKronA} illustrating enhanced text alignment.}
  \label{fig:comparison_text_alignment}
\end{minipage}%
\end{figure*}


We use SDXL and employ our \textit{DiffuseKronA} to generate images from various subjects and text prompts and show its effectiveness in generating images with high fidelity, more accurate color distribution of objects, text alignment, and stability as compared to LoRA-DreamBooth.

\myparagraph{Fidelity \& Color Distribution.}
Our approach consistently produces images of superior fidelity compared to LoRA-DreamBooth, as illustrated in~\cref{fig:comparison_fidelity}. Notably, the \textit{clock} generated by \textit{DiffuseKronA} faithfully reproduces the intricate details, such as the exact depiction of the \textit{\underline{numeral \textbf{3}}}, mirroring the original image. In contrast, the output from LoRA-DreamBooth exhibits difficulties in achieving such high fidelity. Additionally, \textit{DiffuseKronA} demonstrates improved color distribution in the generated images, a feature clearly evident in the \textit{RC Car} images in ~\cref{fig:comparison_fidelity}. Moreover, it struggles to maintain fidelity to the numeral \textit{\underline{numeral \textbf{1}}} on the chest of the sitting toy. Additional examples are shown in~\cref{fig:fidelity_colour} in the Appendix.


    

\myparagraph{Text Alignment.}
\textit{DiffuseKronA} comprehends the intricacies and complexities of text prompts provided as input, producing images that align with the given text prompts, as depicted in~\cref{fig:comparison_text_alignment}. The generated image of the \textit{\underline{anime character}} in response to the prompt exemplifies the meticulous attention \textit{DiffuseKronA} pays to detail. It elegantly captures the \textit{\underline{presence of a shop in the background}}  and \textit{\underline{accompanying soup bowls.}} In contrast, LoRA-DreamBooth struggles to generate an image that aligns seamlessly with the complex input prompt. \textit{DiffuseKronA} not only generates images that align with text but is also proficient in producing a diverse range of images for a given input. More supportive examples are shown in \cref{fig:text_align} in the Appendix.




\def\thefootnote{1}
\myparagraph{Superior Stability.}
\textit{DiffuseKronA} produces images that closely align with the input images across a wide range of learning rates, which are specifically optimized for our approach. In contrast, LoRA-DreamBooth neglects the significance of input images even within its optimal range\footnote{Optimal learning rates are determined through extensive experimentation. Additionally, we have considered observations from \cite{von_Platen_Diffusers_State-of-the-art_diffusion,dreambooth} while fine-tuning LoRA-DreamBooth.} which is evident in~\cref{fig:comparison_stability}. The generated \textit{dog} images by \textit{DiffuseKronA} maintain a high degree of similarity to the input images throughout its optimal range, while LoRA-DreamBooth struggles to perform at a comparable level. Additional examples are shown in \cref{fig:lr} in Appendix.



\begin{table}[!htp]
    \vspace{-2mm}
    \centering
    \resizebox{0.38\textwidth}{!}{%
    \begin{tabular}{c|c|c|c}
    \toprule
        \textsc{Model} & \textsc{CLIP-I } ($\uparrow$) & \textsc{CLIP-T} ($\uparrow$) & \textsc{DINO} ($\uparrow$)\\\toprule
        
        \multirow{2}{*}{{\textbf{\textit{LoRA-DreamBooth}}}} & 0.785 & 0.301 & 0.661 \\

        & $\pm$ 0.062  & $\pm$ 0.027  & $\pm$ 0.127\\
        \midrule

        \cellcolor[gray]{0.9} & \cellcolor[gray]{0.9} \textbf{0.809} & \cellcolor[gray]{0.9} \textbf{0.322} & \cellcolor[gray]{0.9} \textbf{0.677} \\ 
        \multirow{-2}{*}{\cellcolor[gray]{0.9} \textbf{\textit{\textit{DiffuseKronA}}}} & \cellcolor[gray]{0.9} $\pm$ 0.052 & \cellcolor[gray]{0.9} $\pm$ 0.021 & \cellcolor[gray]{0.9} $\pm$0.100 \\
        
        \bottomrule
    \end{tabular}}
    \vspace{-2mm} 
    \caption{\textbf{Quantitative comparison} of CLIP-I, CLIP-T, and DINO scores between \textit{DiffuseKronA} and LoRA-Dreambooth. The obtained values are average across 42 datasets, with a learning rate of $5 \times 10^{-4}$ for \textit{DiffuseKronA} and $1 \times 10^{-4}$ for LoRA-DreamBooth.}
    \vspace{-3mm}  
    \label{tab:sdxl_comp}
\end{table}

\begin{figure*}[!ht]
  \centering
    \includegraphics[width=0.9\textwidth]{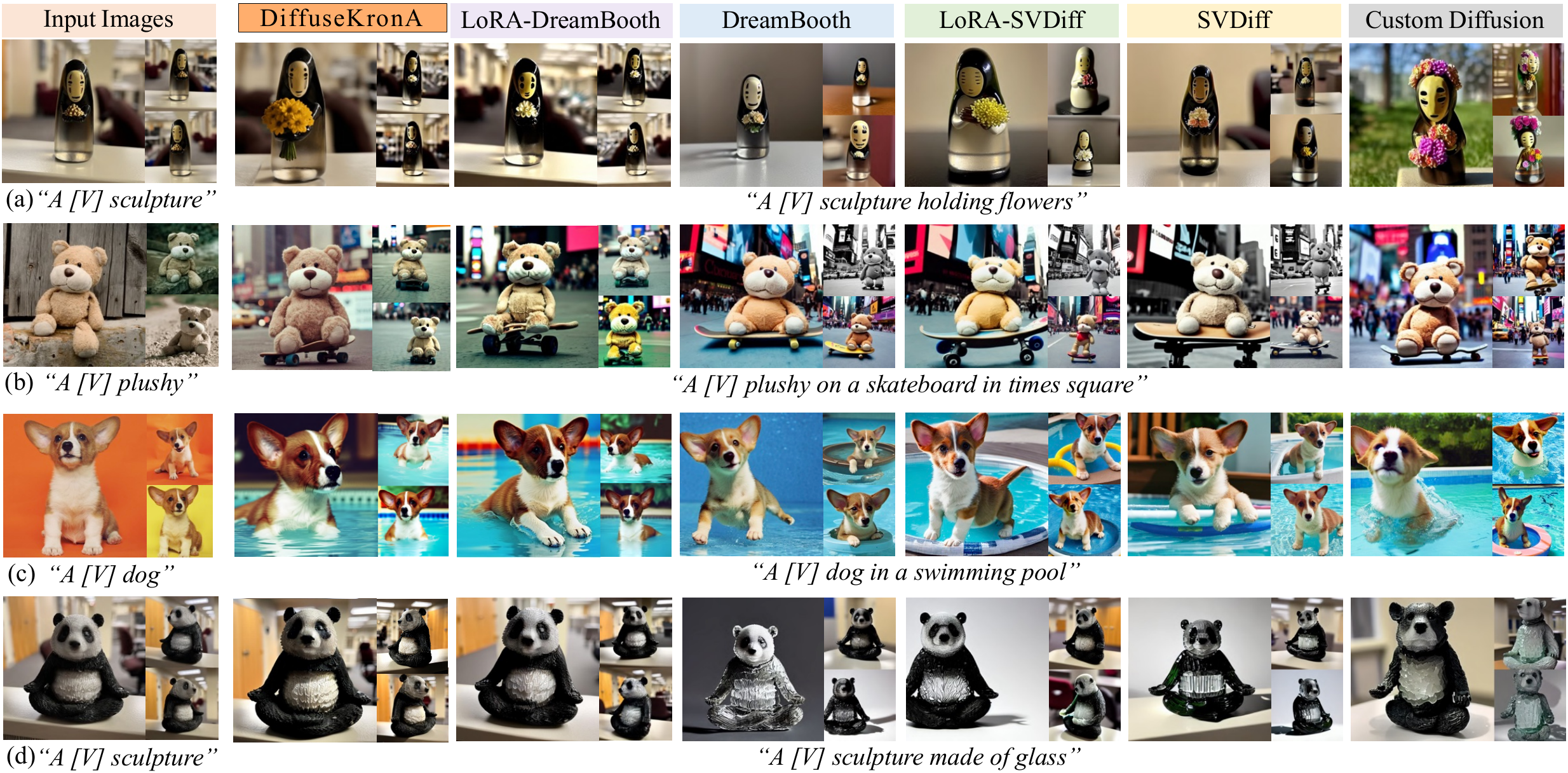}
  \vspace*{-10pt}
  \caption{\textbf{Qualitative comparison} between SVDiff, Custom Diffusion, DreamBooth, LoRA-DreamBooth, and our \textit{DiffuseKronA}. Baseline visual images are extracted from Figure 5 of SVDiff~\cite{han2023svdiff}. Notably, our methods' results are generated considering $a_2=8$. We maintain the original settings of all these methods and used the SD CompVis-1.4~\cite{compvis} variant to ensure a fair comparison.}
  \label{fig:comparison_sota}
  \vspace*{-4pt}
\end{figure*}

\begin{figure*}[!ht]
     \centering
     \begin{subfigure}[h]{0.35\textwidth}
         \centering
         \includegraphics[width=0.92\textwidth]{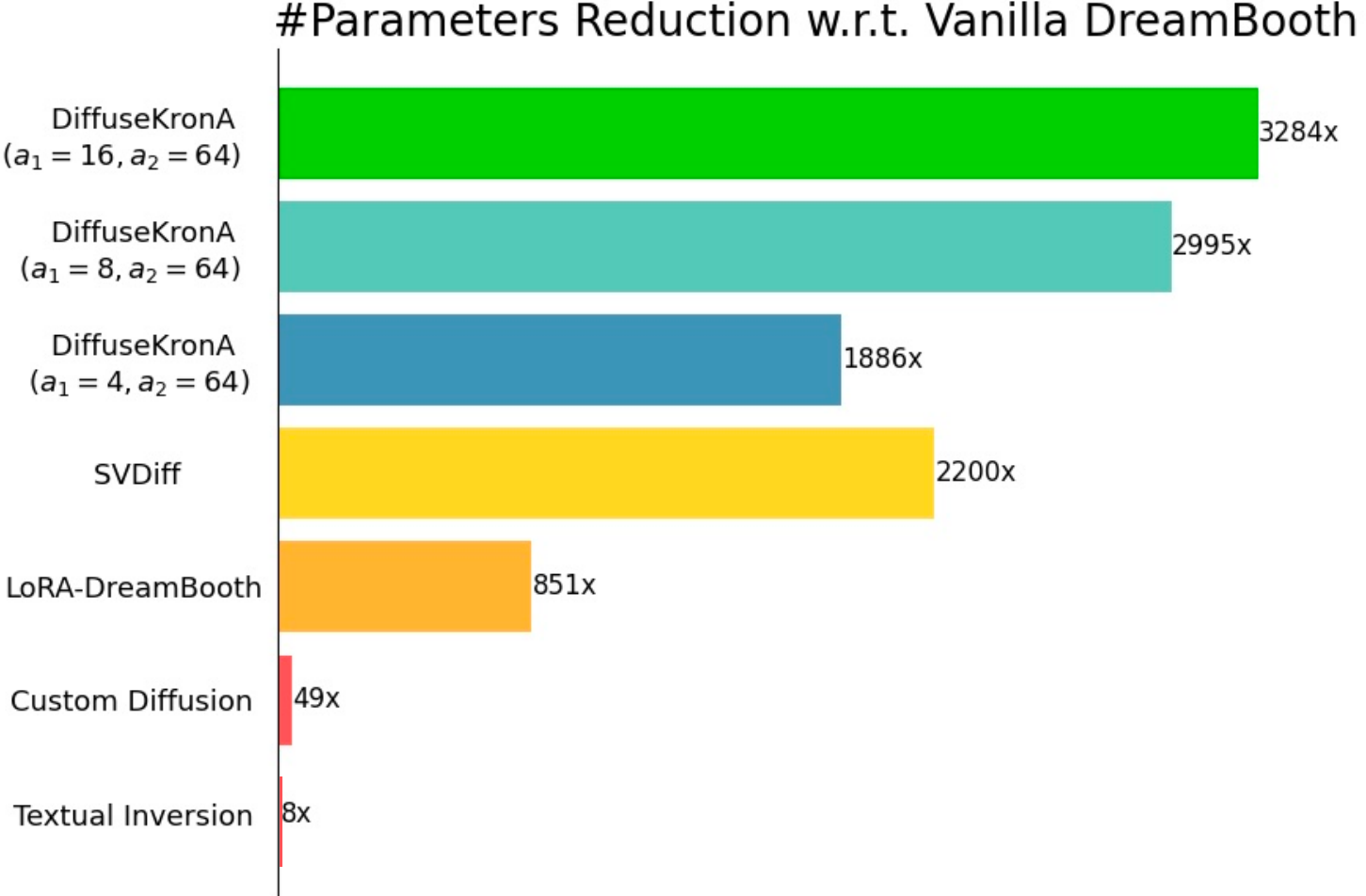}
         \label{fig:parameter_comp}
     \end{subfigure}
     \begin{subfigure}[h]{0.31\textwidth}
         \centering
         \includegraphics[width=0.9\textwidth, width=\textwidth]{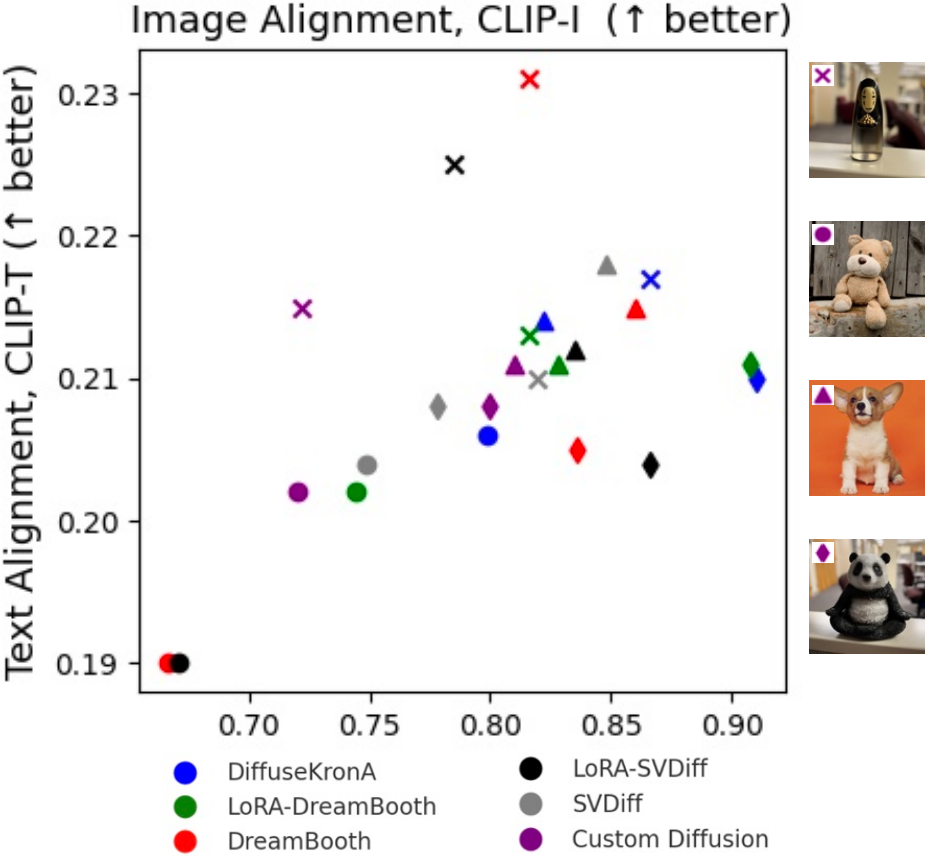}
         \label{fig:comp_clipi}
     \end{subfigure}
     \begin{subfigure}[h]{0.31\textwidth}
         \centering
         \includegraphics[width=\textwidth, width=\textwidth]{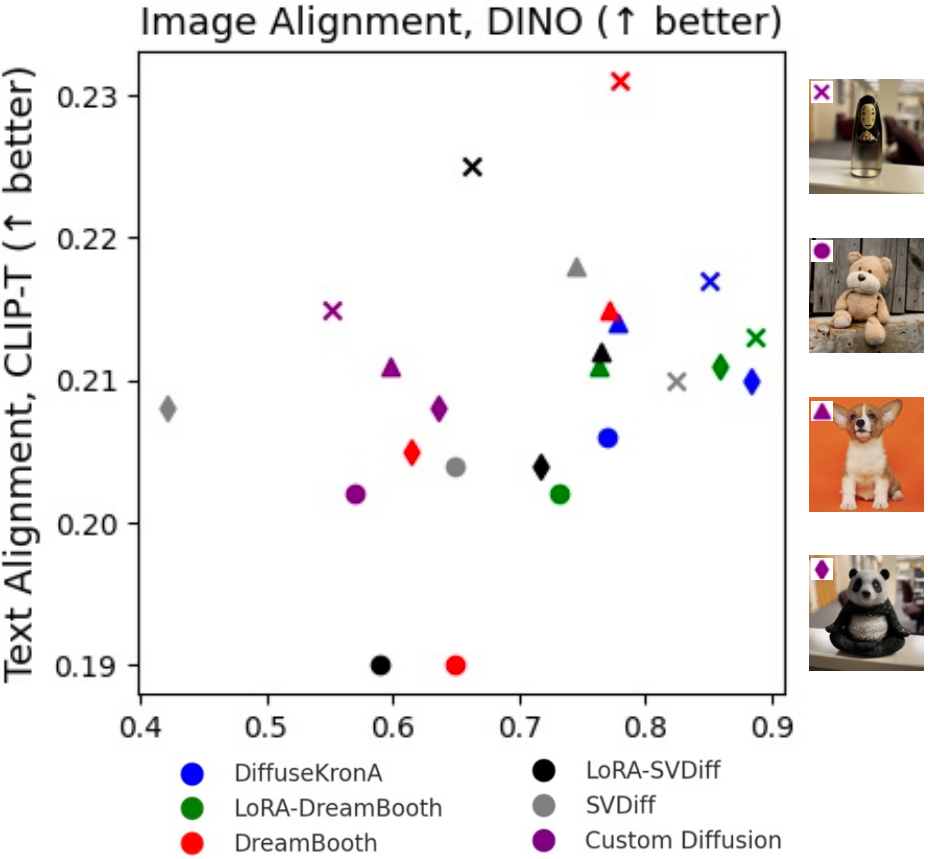}
         \label{fig:comp_dino}
     \end{subfigure}
     \vspace*{-19pt}
    \caption{\textbf{Quantitative comparison} in terms of a) parameter reduction ($\uparrow$ better), and b) text \& image alignment using CLIP-I and DINO with CLIP-T scores, independently computed for each prompt on the same set of input images shown in~\cref{fig:comparison_sota}.}
    \vspace*{-10pt}
    \label{fig:comparison_sota_quant}
\end{figure*}

\myparagraph{Complex Input images and Prompts.}
\textit{DiffuseKronA} consistently performs well, demonstrating robust performance even when presented with intricate inputs. This success is attributed to the enhanced representational power of Kronecker Adapters. As depicted in \cref{fig:teaser-dig}, \textit{DiffuseKronA} adeptly captures the features of the \textit{human face} and \textit{anime characters}, yielding high-quality images. Additionally, from the last row of \cref{fig:teaser-dig}, it is evident that \textit{DiffuseKronA} elegantly captures the semantic nuances of the text. For instance, considering the context of, \textit{\underline{``without blazer''}} and  \textit{\underline{``upset sitting under the umbrella''}}, \textit{DiffuseKronA} generates exceptional images which demonstrate that even when the input text prompt is huge, \textit{DiffuseKronA} adeptly captures various concepts mentioned as nouns in the text. It generates images that encompass all the specified concepts while maintaining a coherent and meaningful overall relationship. Furthermore, we refer the readers to~\cref{fig:human_face_anime} and ~\cref{fig:car} in the Appendix.

\myparagraph{Quantitative Results.}
The distinction in the performance of \textit{DiffuseKronA} and LoRA-DreamBooth is visually evident and is further supported by quantitative measures presented in~\cref{tab:sdxl_comp}, where our model constantly generates images with better DINO and CLIP-I scores and maintains good CLIP-T. The scores for individual datasets are present in~\cref{tab:dataset_wise_results}.
Furthermore, a detailed comparison of our method with other low-rank decomposition methods including LoKr and LoHA~\cite{lokr} are being compared qualitatively and quantitatively in~\cref{fig:comp_lora_lokr_loha} and~\cref{tab:comp_lora_lokr_loha_quant}, respectively.

\tocless{
\subsection{Comparison with State-of-the-arts}
\label{ssec:sota_comp}
}

\begin{table}[!ht]
    \centering
    \resizebox{0.48\textwidth}{!}{%
    \begin{tabular}{c|c|c|c|c}
    \toprule
        \textsc{Model} & \textsc{\# Parameters} ($\downarrow$) & \textsc{CLIP-I } ($\uparrow$) & \textsc{CLIP-T} ($\uparrow$) & \textsc{DINO} ($\uparrow$)\\\toprule
        
        \multirow{2}{*}{{\textbf{\textit{Custom Diffusion}}}} & \multirow{2}{*}{57.1 M} & 0.769 &  0.241 & 0.603 \\
        && $\pm$ 0.043 & $\pm$ 0.029 & $\pm$ 0.055\\
        \midrule
        
        \multirow{2}{*}{{\textbf{\textit{\textit{DreamBooth}}}}}
        & \multirow{2}{*}{982.5 M} & 0.796 & 0.268 & 0.701 \\ 
        && $\pm$ 0.051 & $\pm$ 0.013 & $\pm$ 0.062 \\

        \midrule 
        \multirow{2}{*}{{\textbf{\textit{LoRA-DreamBooth}}}} & \multirow{2}{*}{1.09 M} & 0.808 & 0.260 & 0.710 \\
        && $\pm$ 0.042 & $\pm$ 0.017  & $\pm$ 0.0517\\
        \midrule 
        \multirow{2}{*}{{\textbf{\textit{SVDiff}}}} & \multirow{2}{*}{0.44 M} & 0.806 & 0.265 & 0.705 \\
        && $\pm$ 0.045 & $\pm$ 0.019 & $\pm$ 0.053\\
        \midrule
        \cellcolor[gray]{0.9} & \cellcolor[gray]{0.9} & \cellcolor[gray]{0.9} \textbf{0.822} & \cellcolor[gray]{0.9} \textbf{0.269} & \cellcolor[gray]{0.9} \textbf{0.732} \\
        \multirow{-2}{*}{{\cellcolor[gray]{0.9} \textbf{\textit{DiffuseKronA}}}} & \multirow{-2}{*}{\cellcolor[gray]{0.9} \textbf{0.32 M}} & \cellcolor[gray]{0.9} $\pm$ 0.0259 & \cellcolor[gray]{0.9} $\pm$ 0.011 & \cellcolor[gray]{0.9} $\pm$ 0.039\\
        \bottomrule
    \end{tabular}}

    \vspace{-3mm}
    \caption{\textbf{Quantitative comparison} of \textit{DiffuseKronA} (used variant, $a_1=8$ and $a_2=64$) with SOTA in terms of the number of trainable parameters, text-alignment, and image-alignment scores. The scores are derived from the same set of images and prompts as depicted in~\cref{fig:comparison_sota} and~\cref{fig:comparison_sd}.}
    \vspace{1.5mm}
    \label{tab:comp_sota_quant_appendix}
\end{table}

\hypertarget{4.4}{We} compare \textit{DiffuseKronA} with four related methods, including DreamBooth~\cite{dreambooth}, LoRA-DreamBooth~\cite{von_Platen_Diffusers_State-of-the-art_diffusion}, Custom Diffusion~\cite{custom_diffusion}, SVDiff~\cite{han2023svdiff}, and LoRA-SVDiff~\cite{han2023svdiff}.  As shown in~\cref{fig:comparison_sota}, our \textit{DiffuseKronA} generates high-fidelity images that adhere to input text prompts due to the structure-preserving ability and multiplicative rank property of Kronecker product-based adaption. The images generated by LoRA-DreamBooth often require extensive fine-tuning to achieve the desired results. Methods like custom diffusion take more parameters to fine-tune the model. As compared to SVDiff our proposed approach excels in both (a) achieving superior image-text alignment, as depicted in~\cref{fig:comparison_sota_quant}, and (b) maintaining parameter efficiency.
For each method, we showcase text and image alignment scores in~\cref{fig:comparison_sota_quant} and \textit{DiffuseKronA} obtains the best alignment qualitatively and quantitatively. Additional results across a variety of datasets and prompts are presented in~\cref{fig:comparison_sd} and~\cref{fig:comparison_sd_quant}. Moreover, we present the average scores of all baseline models across 12 datasets, each evaluated with 10 prompts in~\cref{tab:comp_sota_quant_appendix}.


\tocless{
\section{Conclusion}
\label{sec: conclusion}
}

\vspace{-1mm}
We proposed a new parameter-efficient adaption module, \textit{DiffuseKronA}, to enhance text-to-image personalized diffusion models, aiming to achieve high-quality image generation with improved parameter efficiency. Leveraging the Kronecker product's capacity to capture structured relationships in weight matrices, \textit{DiffuseKronA} produces images closely aligned with input text prompts and training images, outperforming LoRA-DreamBooth in visual quality, text alignment, fidelity, parameter efficiency, and stability. 
\textit{DiffuseKronA} thus provides a new and efficient tool for advancing text-to-image personalized image generation tasks.

\newpage


\bibliography{main}
\bibliographystyle{icml2024}
\clearpage


\appendix

    


\renewcommand{\cftsecleader}{\cftdotfill{\cftdotsep}}
\renewcommand{\contentsname}{Table of Contents}
\renewcommand{\cftaftertoctitle}{\par\noindent\rule{\linewidth}{1pt}}

\renewcommand{\cftpnumalign}{l}
\setlength{\cftsecindent}{10pt}
\setlength{\cftsubsecindent}{30pt}
\setlength{\cftsecnumwidth}{17pt}
\tableofcontents
\par
\rule{\linewidth}{1pt}

\section{Background}
\label{ssec:background}

Primarily in 1998, the practical implications of the Kronecker product were introduced in \cite{nagy2003kronecker} for the task of image restoration. This study presented a flexible preconditioning approach based on Kronecker product and singular value decomposition (SVD) approximations. The approach can be used with a variety of boundary conditions, depending on what is most appropriate for the specific deblurring application.

In the realm of parameter-efficient fine-tuning (PEFT) of large-scale models in deep learning, several literature studies ~\cite{tahaei-etal-2022-kroneckerbert, edalati2022krona, he2022parameter, thakker2019compressing} have explored the efficacy of Kronecker products, illustrating their applications across diverse domains. 
\begin{tcolorbox}
    Most of the shown images in this study are generated using the SDXL~\cite{sdxl} backbone. However, for comparison figures, we have used the SD CompVis-1.4~\cite{compvis} variant and we have explicitly mentioned in the captions of these figures. 
\end{tcolorbox}
In context, COMPACTER~\cite{edalati2021compacter} was the first line of work that proposes a method for fine-tuning large-scale language models with a better trade-off between task performance and the number of trainable parameters than prior work. It builds on top of ideas from adapters~\cite{houlsby2019parameter}, low-rank optimization~\cite{li2018measuring} (by leveraging Kronecker products), and parameterized hypercomplex multiplication layers~\cite{zhang2020beyond}. KroneckerBERT~\cite{tahaei-etal-2022-kroneckerbert} significantly compressed Pre-trained Language Models (PLMs) through Kronecker decomposition and knowledge distillation. It leveraged Kronecker decomposition to compress the embedding layer and the linear mappings in the multi-head attention, and the feed-forward network modules in the Transformer layers within BERT~\cite{devlin2018bert} model. The model outperforms state-of-the-art compression methods on the GLUE and SQuAD benchmarks. In a similar line of work, KronA~\cite{edalati2022krona} proposed a Kronecker product-based adapter module for efficient fine-tuning of Transformer-based PLMs (T5~\cite{raffel2020exploring}) methods on the GLUE benchmark.

\begin{figure}[h]
    \centering
    \vspace{-2mm}
    \includegraphics[width=0.47\textwidth]{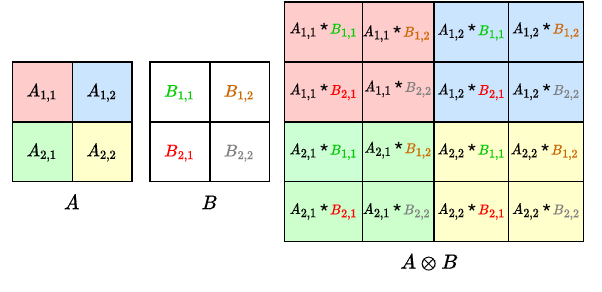}
    \vspace{-3mm}
    \caption{Demonstrating the functioning of the Kronecker product.}
    \label{fig:how_krona_works}
    \vspace{-3mm}
\end{figure}

\begin{figure*}[!ht]
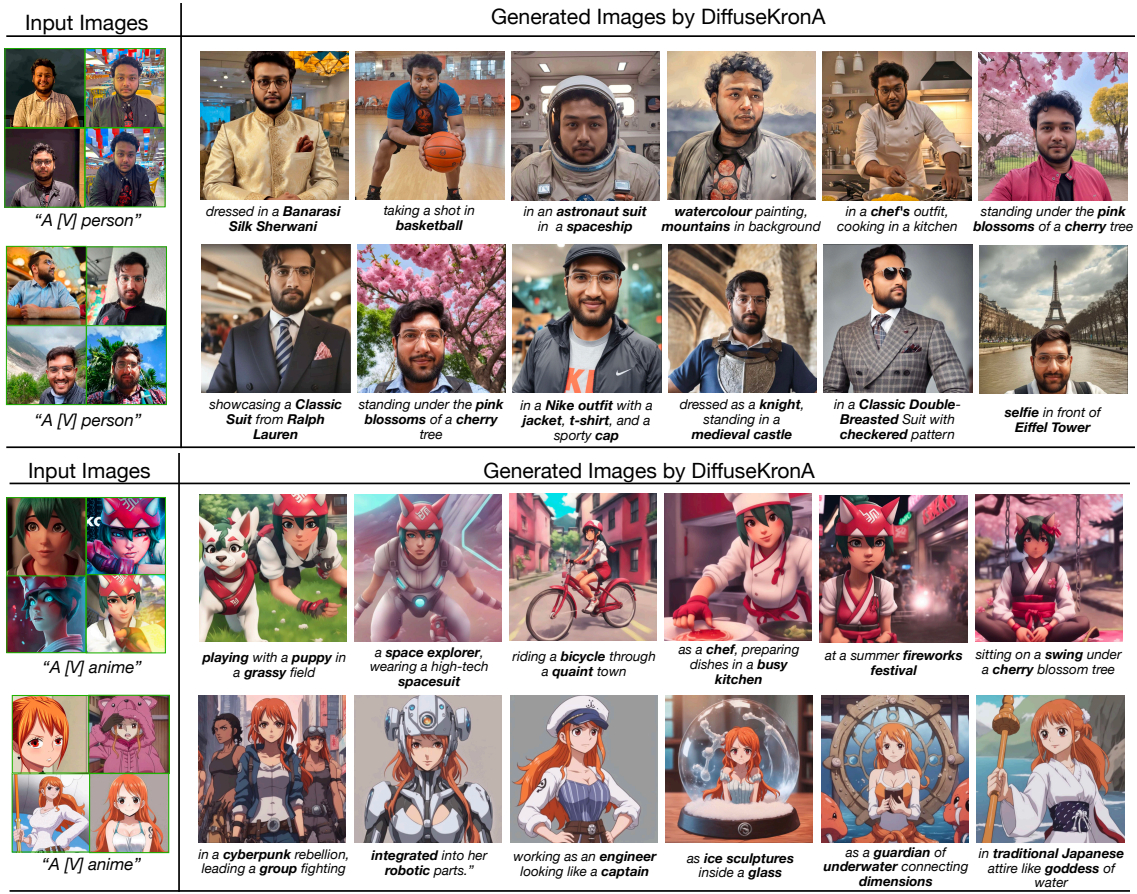

    \centering
    \captionsetup{type=figure}
    \includegraphics[width=0.88\textwidth]{supple_images/face.pdf}
    \includegraphics[width=0.873\textwidth]{supple_images/anime.pdf}
    \caption{The results for the human face and anime characters generation highlight our method's endless application in creating portraits, animes, and avatars. 
    }
    \label{fig:human_face_anime}
\end{figure*}

\begin{figure*}[!ht]
    \centering
    \captionsetup{type=figure}
    \includegraphics[width=0.88\textwidth]{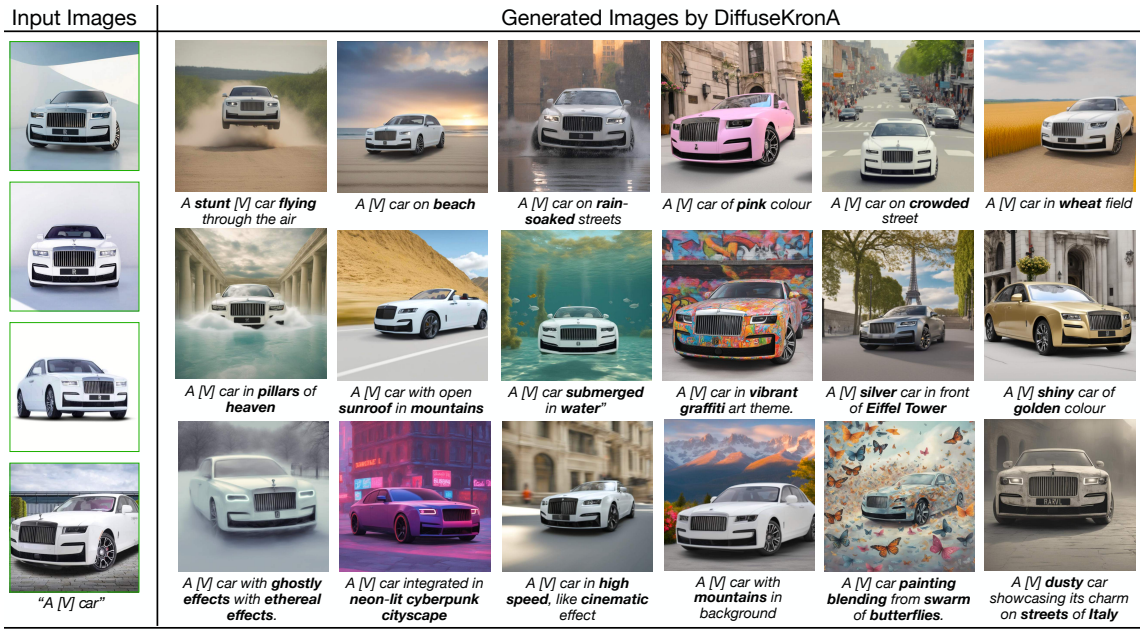}
    \caption{Results for car modifications and showcasing our method's potential application in the Automobile industry.}
    \label{fig:car}
\end{figure*}

\begin{figure*}[!ht]
  \centering
  \includegraphics[height=0.9\textwidth]{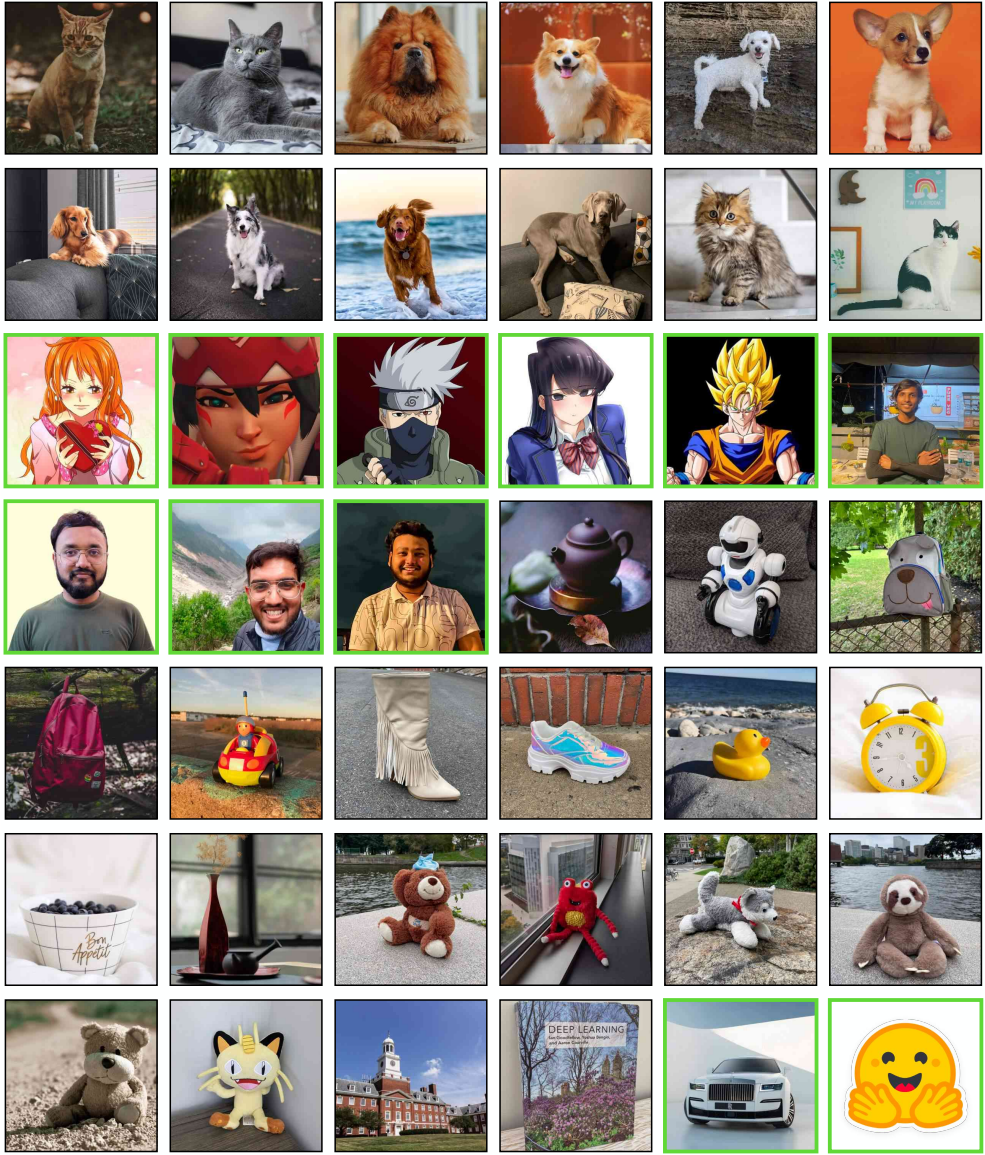}
    \vspace*{-3mm}
    \caption{A collection of sample images representing all individual subjects involved in this study. Our collected subjects are highlighted in \textcolor{green}{green}.}
    \vspace*{-5mm}
    \label{fig:dataset_all}
\end{figure*}

Apart from the efficient fine-tuning of PLMs, studies also shed some light on applying Kronecher products in the compression of convolution neural networks (CNNs) and vision transformers (ViTs).
For instance, in~\cite{hameed2021convolutional}, the authors compressed CNNs through generalized Kronecker product decomposition (GKPD) with a fundamental objective to reduce both memory usage and the required floating-point operations for convolutional layers in CNNs. This approach offers a plug-and-play module that can be effortlessly incorporated as a substitute for any convolutional layer, offering a convenient and adaptable solution. Recently proposed, KAdaptation~\cite{he2022parameter} studies parameter-efficient model adaptation strategies for ViTs on the image classification task. It formulates efficient model adaptation as a subspace training problem via Kronecker Adaptation (KAdaptation) and performs a comprehensive benchmarking over different efficient adaptation methods.

On the other hand, authors of \cite{thakker2019compressing} compressed RNNs for resource-constrained environments (e.g. IOT devices) using Kronecker product (KP) by 15-38x with minimal accuracy loss and by quantizing the resulting models to 8 bits, the compression factor is further pushed to 50x. In~\cite{wang2023kronecker}, RNNs are compressed based on a novel Kronecker CANDECOMP/PARAFAC (KCP) decomposition, derived from Kronecker tensor (KT) decomposition, by proposing two fast algorithms of multiplication between the input and the tensor-decomposed weight.\\
Besides all of the above, Kronecker decomposition is also being applied for GPT compression~\cite{edalati2022kronecker} which attempts to compress the linear mappings within the GPT-2 model. The proposed model, Kronecker GPT-2 (KnGPT2) is initialized based on the Kronecker decomposed version of the GPT-2 model. Subsequently, it undergoes a very light pre-training on only a small portion of the training data with intermediate layer knowledge distillation (ILKD).\\
From the aforementioned literature study, we have witnessed the efficacy of Kronecker products for the task of model compression within various domains including NLP, RNN, CNN, ViT, and GPT space. Consequently, it has sparked considerable interest in exploring its impact on Generative models. 

\section{Datasets Descriptions}
\label{sec:dataset_appendix}
We have incorporated a total of 25 datasets from DreamBooth~\cite{dreambooth}, encompassing images of backpacks, dogs, cats, and stuffed animals. Additionally, we integrated 7 datasets from custom diffusion~\cite{custom_diffusion} to introduce variety in our experimentation. To assess our model's ability to capture spatial features on faces, we curated a dataset consisting of 4 to 7 images each of 4 humans, captured from different angles. To further challenge our model against complex input images and text prompts, we compiled a dataset featuring 6 anime images from various sources. All datasets are categorized into four groups: \textbf{\textit{living animals}}, \textbf{\textit{non-living objects}}, \textbf{\textit{anime}}, and \textbf{\textit{human faces}}. Furthermore, the keywords utilized for fine-tuning the model remain consistent with those specified in the original papers. In~\cref{fig:dataset_all}, we present a sample image for all the considered subjects used in this study.

\myparagraph{Image Attribution.} Our collected datasets are taken from the following resources:

\begin{itemize}
    \item \textbf{Rolls Royce:}
    \begin{itemize}
    \small{
    \item \url{https://www.peakpx.com/en/hd-wallpaper-desktop-pxxec}
    \item \url{https://4kwallpapers.com/cars/rolls-royce-ghost-2020/white-background-5k-8k-2554.html} 
    \item \url{https://www.cardekho.com/Rolls-Royce/Rolls-Royce_Ghost/pictures#leadForm}
    \item \url{https://www.rolls-roycemotorcars.com/en_US/showroom/ghost-digital-brochure.html}}
    \end{itemize}
    
    \item \normalsize{\textbf{Hugging Face:}} \small{\url{https://huggingface.co/brand}}
    
    \item \normalsize{\textbf{Nami:}}
    \begin{itemize}
    \small{
    \item \url{http://m.gettywallpapers.com/nami-pfps-2/}
    \item \url{https://tensor.art/models/616615209278282245} 
    \item \url{https://www.facebook.com/NamiHotandCute/?locale=bs_BA}
    \item \url{https://k.sina.cn/article_1655152542_p62a79f9e02700nhhe.html}}
    \end{itemize}
    
    \item \normalsize{\textbf{Kiriko:}}
    \begin{itemize}
    \small{
    \item \url{https://in.pinterest.com/pin/306948530865002366/}
    \item \url{https://encrypted-tbn2.gstatic.com/images?q=tbn:ANd9GcSDSk98Uw3O2XW_RFC1jD_Kmw70JWU459euVYtU9nn1CpzPDcwS}
    \item \url{https://comisc.theothertentacle.com/overwatch+kiriko+fanart}
    \item \url{https://www.1999.co.jp/eng/11030018}}
    \end{itemize}
    
    \item \normalsize{\textbf{Shoko Komi:}}
    \small{
    \begin{itemize}
        \item \url{https://wallpaperforu.com/tag/komi-shouko-wallpaper/page/2/}
        \item \url{https://www.tiktok.com/@anime_geek00/video/7304798157894995243}
        \item \url{https://wall.alphacoders.com/big.php?i=1305702}
        \item \url{http://m.gettywallpapers.com/komi-can-t-communicate-wallpapers/}
    \end{itemize}}
    
    \item \normalsize{\textbf{Kakashi Hatake:}}
    \small{
    \begin{itemize}
        \item \url{https://www.ranker.com/list/best-kakashi-hatake-quotes/ranker-anime?page=2}; 
        \item \url{https://www.wallpaperflare.com/search?wallpaper=Hatake+Kakashi}
        \item \url{https://www.peakpx.com/en/hd-wallpaper-desktop-kiptm}
        \item \url{https://in.pinterest.com/pin/584060645404620659/}
    \end{itemize}}
    \normalsize
\end{itemize}

\begin{table*}[t]
    \centering
    \resizebox{0.95\textwidth}{!}{
    \begin{tabular}{c|c|c|c|c|c|c}
        \toprule
        \textbf{Subject} & Cat & Cat2 & Dog2 & Dog & Dog3 & Dog6 \\
        \midrule
        \textbf{CLIP-I} & 0.858 $\pm$ 0.017      & 0.826 $\pm$ 0.030 & 0.833 $\pm$ 0.023 & 0.854 $\pm$ 0.015 & 0.789 $\pm$ 0.027 & 0.845 $\pm$ 0.031 \\
        \textbf{CLIP-T} & 0.348 $\pm$ 0.033      & 0.343 $\pm$ 0.030 & 0.331 $\pm$ 0.028 & 0.349 $\pm$ 0.029 & 0.338 $\pm$ 0.025 & 0.323 $\pm$ 0.032 \\
        \textbf{DINO}  & 0.814 $\pm$ 0.025 & 0.752 $\pm$ 0.021 & 0.750 $\pm$ 0.049 & 0.856 $\pm$ 0.008 & 0.549 $\pm$ 0.060 & 0.788 $\pm$ 0.017 \\
        \midrule
        \midrule
        \textbf{Subject} & Dog5 & Dog7 & Dog8 & Doggy & Cat3 & Cat4 \\
        \midrule
        \textbf{CLIP-I}  & 0.824 $\pm$ 0.024 & 0.853 $\pm$ 0.015          & 0.829 $\pm$ 0.021 & 0.734 $\pm$ 0.031 & 0.834 $\pm$ 0.034 & 0.861 $\pm$ 0.016 \\
        \textbf{CLIP-T}  & 0.337 $\pm$ 0.026 & 0.334 $\pm$ 0.025 & 0.343 $\pm$ 0.026 & 0.329 $\pm$ 0.030 & 0.348 $\pm$ 0.029 & 0.349 $\pm$ 0.032 \\
        \textbf{DINO} & 0.761 $\pm$ 0.001 & 0.730 $\pm$ 0.049 & 0.717 $\pm$ 0.050 & 0.686 $\pm$ 0.039 & 0.744 $\pm$ 0.031 & 0.863 $\pm$ 0.030 \\
        \midrule
        \midrule
        \textbf{Subject} & Nami (Anime) & Kiriko (Anime) & Kakshi (Anime) & Shoko Komi (Anime) & Harshit (Human) & Nityanand (Human) \\
        \midrule
        \textbf{CLIP-I}  &  0.781 $\pm$ 0.035 & 0.738 $\pm$ 0.039 & 0.834 $\pm$ 0.028 & 0.761 $\pm$ 0.029 & 0.724 $\pm$ 0.018 & 0.665 $\pm$ 0.031 \\
        \textbf{CLIP-T}  & 0.337 $\pm$ 0.029 & 0.320 $\pm$ 0.032 & 0.318 $\pm$ 0.031 & 0.356 $\pm$ 0.028 & 0.297 $\pm$ 0.036 & 0.307 $\pm$ 0.030 \\
        \textbf{DINO} & 0.655 $\pm$ 0.023 & 0.483 $\pm$ 0.041 & 0.617 $\pm$ 0.061 & 0.596 $\pm$ 0.024 & 0.555 $\pm$ 0.025  & 0.447 $\pm$ 0.068 \\
        \midrule
        \midrule
        \textbf{Subject} & Shyam (Human) & Teapot & Robot Toy & Backpack & Dog Backpack & Rc Car \\
        \midrule
        \textbf{CLIP-I}  & 0.731 $\pm$ 0.015 & 0.836 $\pm$ 0.051 & 0.828 $\pm$ 0.026 & 0.907 $\pm$ 0.026 & 0.774 $\pm$ 0.037 & 0.797 $\pm$ 0.020 \\
        \textbf{CLIP-T}  & 0.297 $\pm$ 0.026 & 0.347 $\pm$ 0.025 & 0.285 $\pm$ 0.032 & 0.347 $\pm$ 0.021 & 0.333 $\pm$ 0.027 & 0.321 $\pm$ 0.027 \\
        \textbf{DINO} & 0.531 $\pm$ 0.030 & 0.528 $\pm$ 0.132 & 0.642 $\pm$ 0.023 & 0.660 $\pm$ 0.088 & 0.649 $\pm$ 0.037 & 0.651 $\pm$ 0.065 \\
        \midrule
        \midrule
        \textbf{Subject} & Shiny Shoes & Duck & Clock & Vase & Plushie1 & Monster Toy \\
        \midrule
        \textbf{CLIP-I}  & 0.806 $\pm$ 0.025 & 0.845 $\pm$ 0.023 & 0.825 $\pm$ 0.062 & 0.827 $\pm$ 0.013 & 0.897 $\pm$ 0.014 & 0.782 $\pm$ 0.041 \\
        \textbf{CLIP-T} & 0.308 $\pm$ 0.023 & 0.303 $\pm$ 0.016 & 0.308 $\pm$ 0.035 & 0.332 $\pm$ 0.026 & 0.308 $\pm$ 0.030 &  0.308 $\pm$ 0.029 \\
        \textbf{DINO} & 0.735 $\pm$ 0.090 & 0.682 $\pm$ 0.049 & 0.590 $\pm$ 0.158 & 0.705 $\pm$ 0.025 & 0.813 $\pm$ 0.027 & 0.573 $\pm$ 0.060 \\
        \midrule
        \midrule
        \textbf{Subject} & Plushie2 & Plushie3 & Building & Book & Car & HuggingFace \\
        \midrule
        \textbf{CLIP-I}  & 0.803 $\pm$ 0.022 & 0.792 $\pm$ 0.015 & 0.852 $\pm$ 0.013 & 0.695 $\pm$ 0.023 & 0.830 $\pm$ 0.024 & 0.810 $\pm$ 0.002 \\
        \textbf{CLIP-T}  & 0.324 $\pm$ 0.024 & 0.337 $\pm$ 0.031 & 0.268 $\pm$ 0.023 & 0.301 $\pm$ 0.022 & 0.299 $\pm$ 0.032 & 0.288 $\pm$ 0.042 \\
        \textbf{DINO} & 0.728 $\pm$ 0.020 & 0.766 $\pm$ 0.033 & 0.742 $\pm$ 0.019 & 0.579 $\pm$ 0.040 & 0.684 $\pm$ 0.036 & 0.692 $\pm$ 0.001 \\
        \bottomrule
  \end{tabular}}
  \label{tab:dataset_wise_results}
  \vspace*{-2mm}
  \caption{Average metrics (CLIP-I, CLIP-T, and DINO scores) from various prompt runs for each subject using our proposed method.}
  \vspace*{-2mm}
\end{table*}

\section{Evaluation Metrics}
\label{sec:evaluation_metrics_appendix}
We utilize metrics introduced in DreamBooth~\cite{dreambooth} for evaluation: DINO and CLIP-I scores measure subject fidelity, while CLIP-T assesses image-text alignment. The DINO score is the normalized pairwise cosine similarity between the \texttt{ViT-S/16 DINO} embeddings of the generated and input (real) images. Similarly, the CLIP-I score is the normalized pairwise \texttt{CLIP ViT-B/32 image} embeddings of the generated and input images. Meanwhile, the CLIP-T score computes the normalized cosine similarity between the given text prompt and generated image CLIP embeddings.

Let's denote the pre-trained CLIP Image encoder as $\mathcal{I}$, the CLIP text encoder as $\mathcal{T}$, and the DINO model as $\mathcal{V}$. We measure cosine similarity between two embeddings $x$ and $y$ as $sim(x,\:y) = \frac{x.y}{||x||\cdot||y||}$. Given two sets of images, we represent the input image set as $\mathcal{R} = \left\{R_{i}\right\}_{i=1}^{n}$ and generated image set as $\mathcal{G} = \left\{G_{i}\right\}_{i=1}^{m}$ corresponding to the prompt set $\mathcal{P} = \left\{P_{i}\right\}_{i=1}^{m}$, where $m$ and $n$ represents the number of generated and input images, respectively and $R, G \in \mathbb{R}^{3\times H \times W}$ ($H$ and $W$ is the height and width of the image). Then, CLIP-I image-to-image and CLIP-T text-to-image similarity scores would be computed as $S_{CLIP}^{I}$ and $S_{CLIP}^{T}$, respectively.
\begin{equation}
    S_{CLIP}^{I} =  \frac{1}{mn}\sum \limits_{i=1}^{n} \sum \limits_{j=1}^{m} sim (\mathcal{I}\left(R_{i}\right),\:\mathcal{I}\left(G_{j}\right))
\end{equation}
  
\begin{equation}
    S_{CLIP}^{T} =  \frac{1}{m}\sum \limits_{i=1}^{m} sim (\mathcal{I}\left(G_{i}\right),\:\mathcal{T}\left(P_{i}\right))
\end{equation}
Similarly, the DINO image-to-image similarity score would be computed as \begin{equation}
    S_{DINO} =  \frac{1}{mn}\sum \limits_{i=1}^{n} \sum \limits_{j=1}^{m} sim (\mathcal{V}\left(R_{i}\right),\:\mathcal{V}\left(G_{j}\right)).
\end{equation}

Notably, the DINO score is preferred to assess subject fidelity owing to its sensitivity to differentiate between subjects within a given class. In personalized T2I generations, all three metrics should be considered jointly for evaluation to avoid a biased conclusion. For instance, models that copy training set images will have high DINO and CLIP-I scores but low CLIP-T scores, while a vanilla T2I generative model like SD and SDXL without subject knowledge will produce high CLIP-T scores with poor subject alignment. As a result, both models are not considered desirable for the subject-driven T2I generation.
In Table-5, we showcase mean subject-specific CLIP-I, CLIP-T, and DINO scores along with standard deviations computed across 36 datasets, with a total of around 1600 generated images and prompts.

\section{\textit{DiffuseKronA} Ablations Study}
\label{sec:supple_ablation}
As outlined in \hyperlink{4.2}{4.2} of the main paper, we explore various trends and observations derived from extensive experimentation on the datasets specified in ~\cref{fig:dataset_all}.

\begin{figure*}[!ht]
  \centering
  \vspace*{-2mm}
  \includegraphics[width=0.7\textwidth]{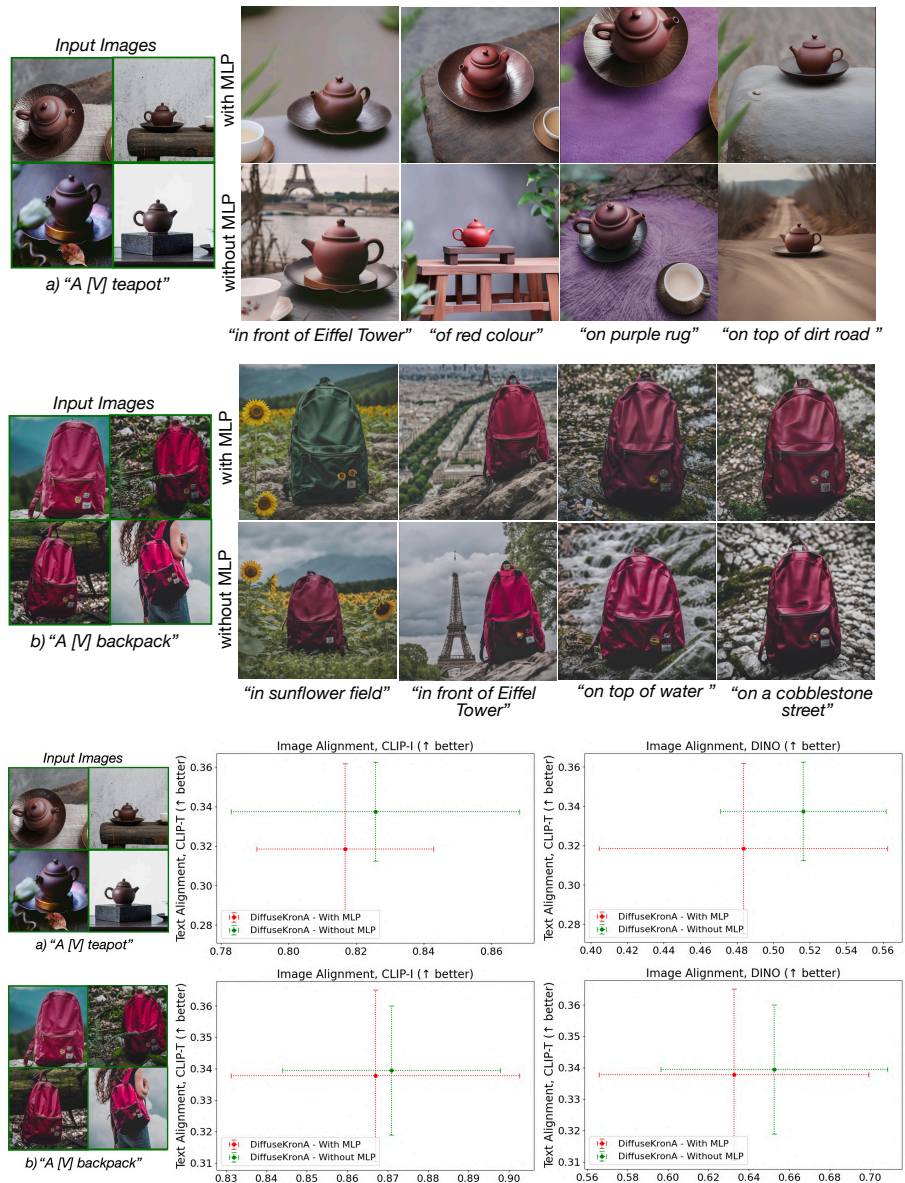}
  \vspace*{-5mm}
  \caption{Qualitative and Quantitative comparison between fine-tuning with MLP and w/o MLP. Fine-tuning MLP layers introduces more parameters and doesn't enhance image generation compared with fine-tuning solely attention-weight matrices. So, the best outcomes and efficient use of parameters occur when only attention weight (without MLP) matrices are fine-tuned.}
  \label{fig:with_without_MLP}
\end{figure*}

\subsection{Choice of modules to fine-tune the model}
\label{ssec:without_with_MLP}
Within the UNet network's transformer block, the linear layers consist of two components: a) attention matrices and b) a feed-forward network (FFN). Our investigation focuses on discerning the weight matrices with the highest importance for fine-tuning, aiming for efficiency in parameter utilization.

Our findings reveal that fine-tuning only the attention weight matrices, namely $\left(W_K, W_Q, W_V, W_O\right)$, proves to be the most impactful and parameter-efficient strategy. Conversely, fine-tuning the FFN layers does not significantly enhance image synthesis quality but substantially increases the parameter count, approximately doubling the computational load. Refer to ~\cref{fig:with_without_MLP} for a visual representation comparing synthesis image quality with and without fine-tuning FFN layers on top of attention matrices. This graph unequivocally demonstrates that incorporating MLP layers does not enhance fidelity in the results. On the contrary, it diminishes the quality of generated images in certain instances, such as \underline{\textit{``A [V] backpack in sunflower field''}}, while concurrently escalating the number of trainable parameters substantially, approximately 2x times.

This approach of exclusively fine-tuning attention layers not only maximizes efficiency but also helps maintain a lower overall parameter count. This is particularly advantageous when computational resources are limited, ensuring computational efficiency in the fine-tuning process.

\begin{figure*}[!ht]
  \centering
  \vspace*{-1mm}
  \includegraphics[width=0.85\textwidth]{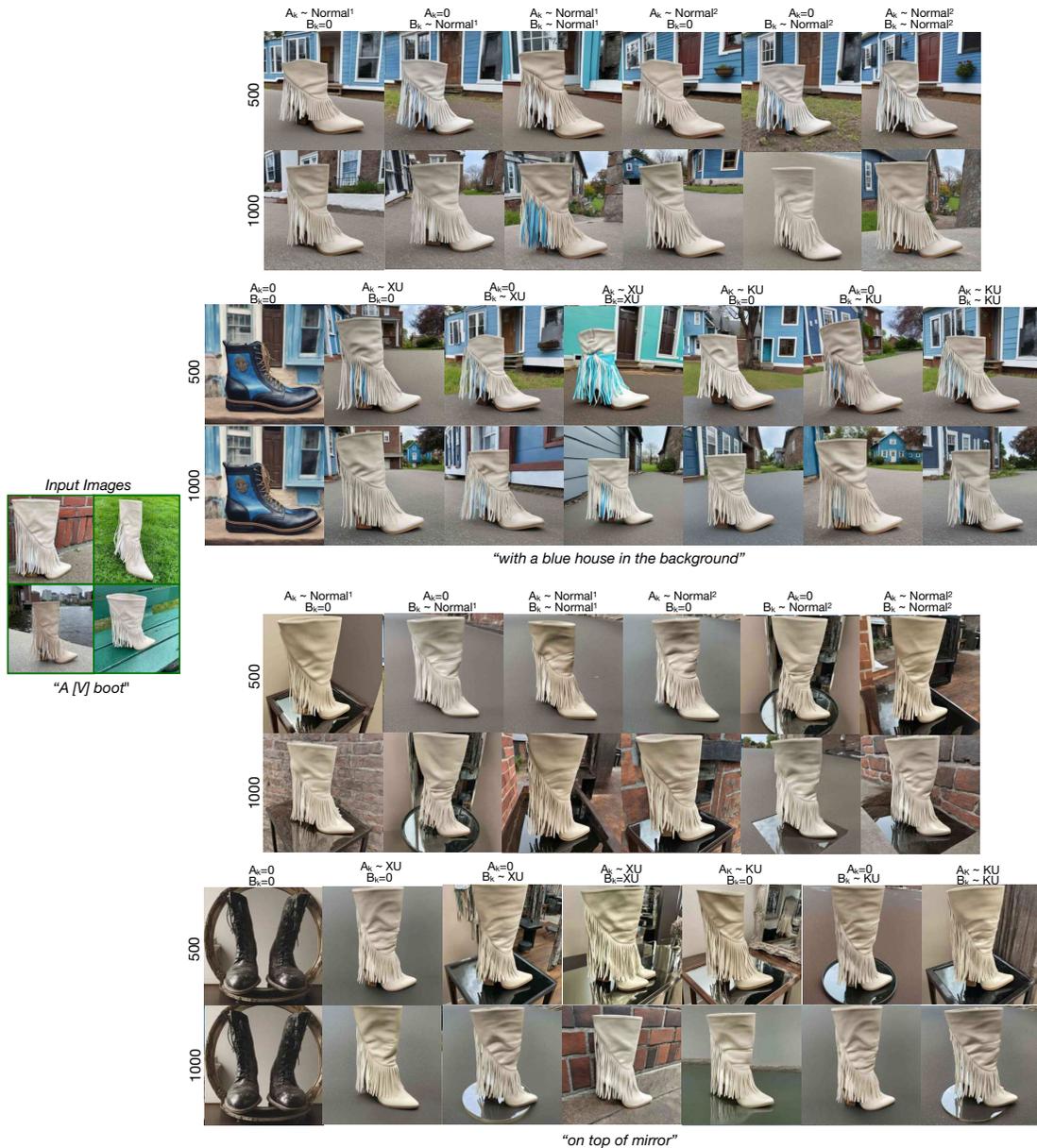}
  \vspace*{-4.2mm}
  \caption{Impact of different initialization strategies: optimal outcomes are achieved when initializing $B_k$ to zero while initializing $A_k$ with either a Normal or Kaiming uniform distribution.}
  \vspace*{-2mm}
  \label{fig:init}
\end{figure*}

\subsection{Effect of Kronecker Factors}
\label{ssec:Effect_Kronecker_Factors}

\begin{figure*}[!ht]
    \centering
    \begin{subfigure}{\textwidth}
        \centering
        \includegraphics[width=0.9\textwidth]{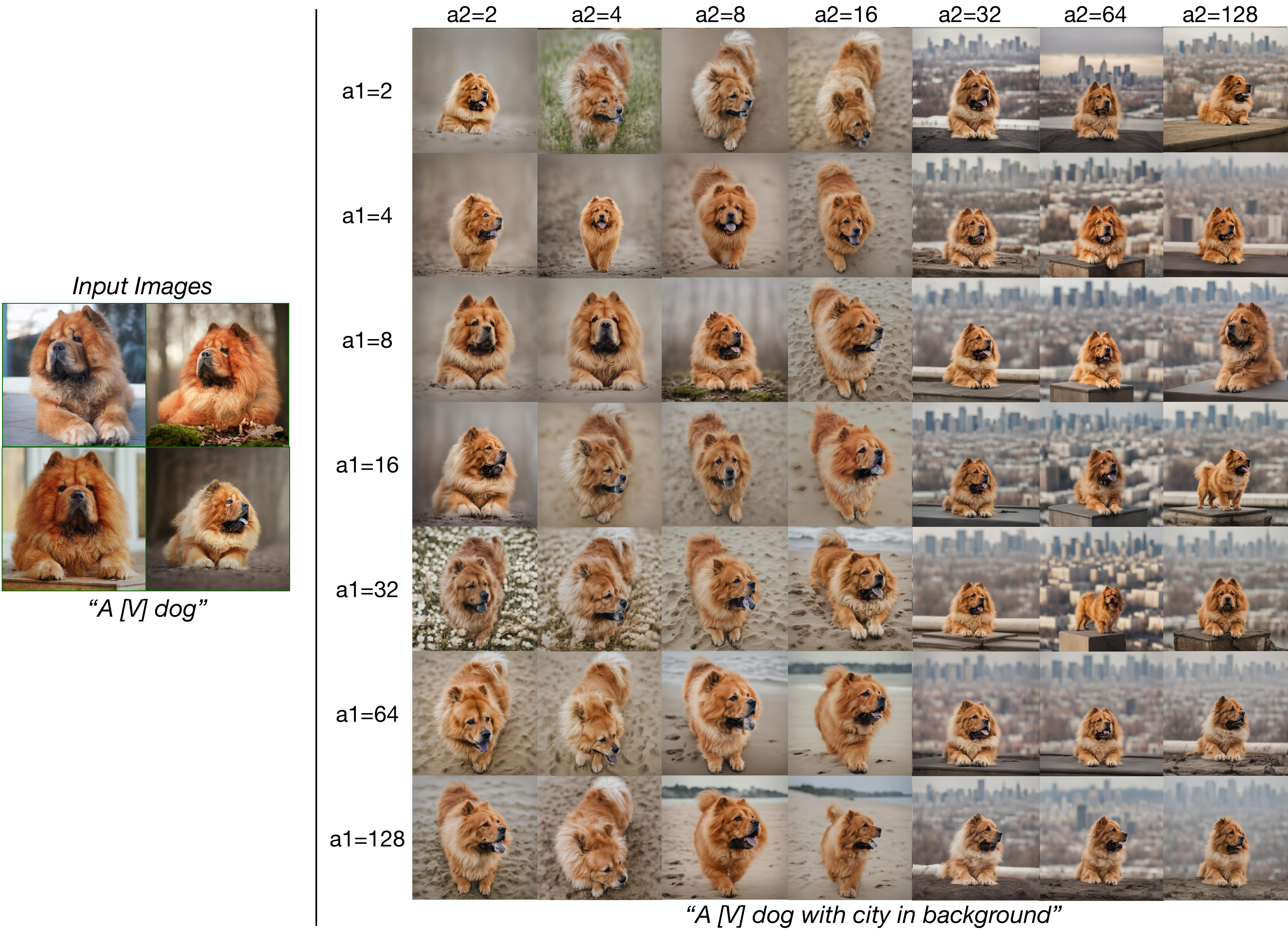}
    \end{subfigure}
    \begin{subfigure}{\textwidth}
        \centering
        \includegraphics[width=0.9\textwidth]{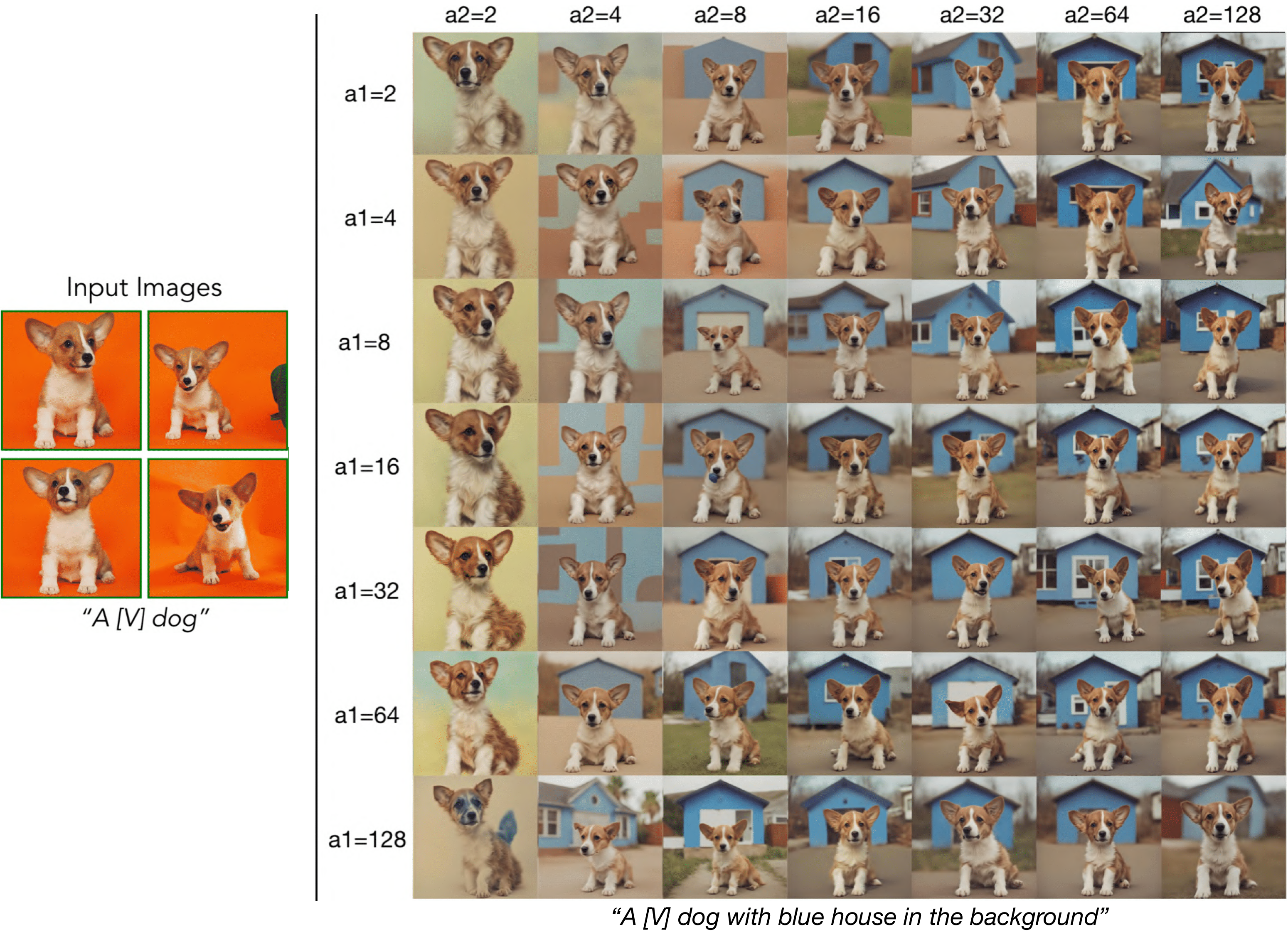}
    \end{subfigure}
\end{figure*}

\begin{figure*}[!ht]
    \ContinuedFloat
    \centering
    \begin{subfigure}{0.84\textwidth}
        \centering
        \includegraphics[width=\textwidth]{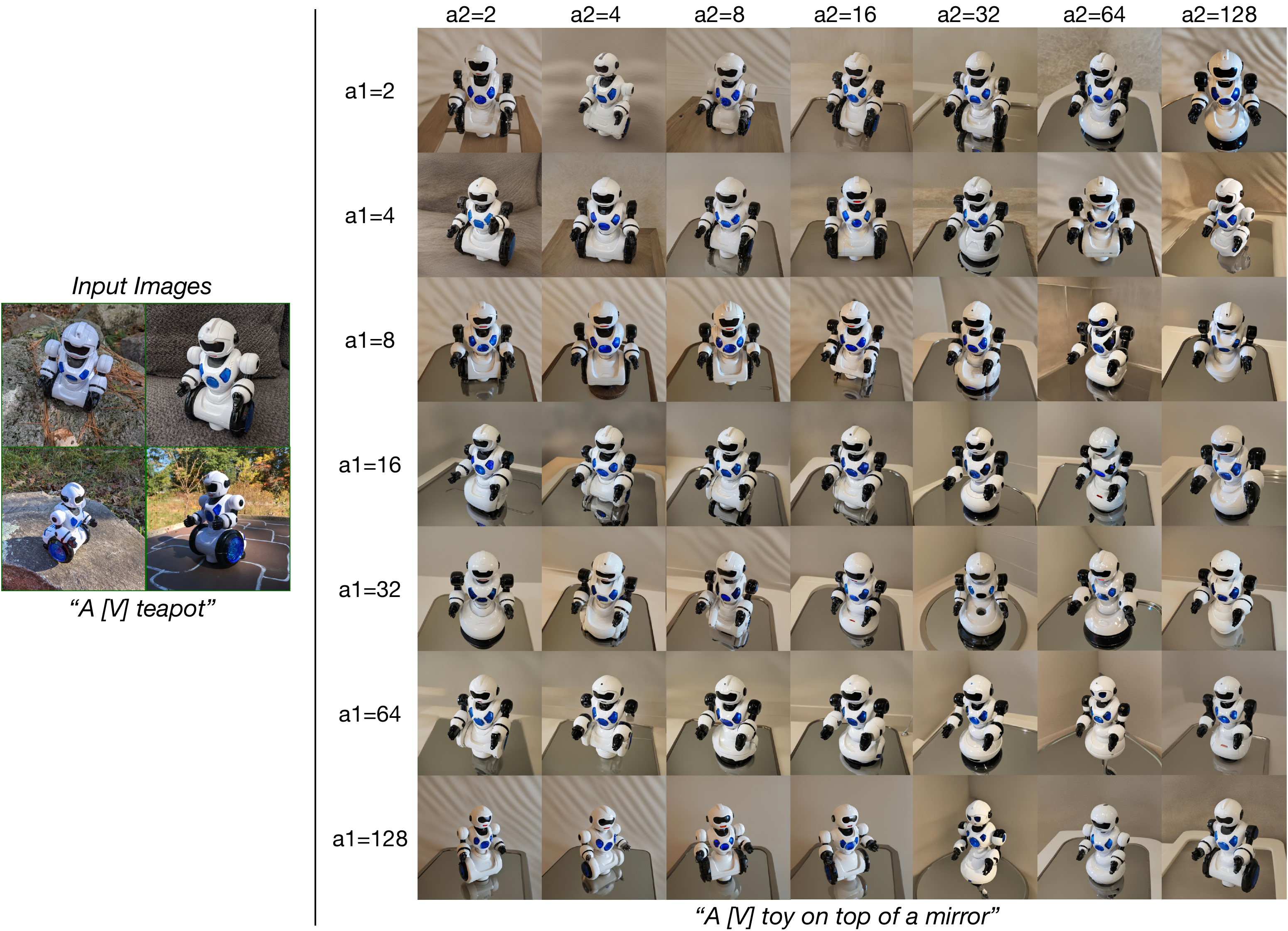}
    \end{subfigure}
    \begin{subfigure}{0.84\textwidth}
        \centering
        \includegraphics[width=\textwidth]{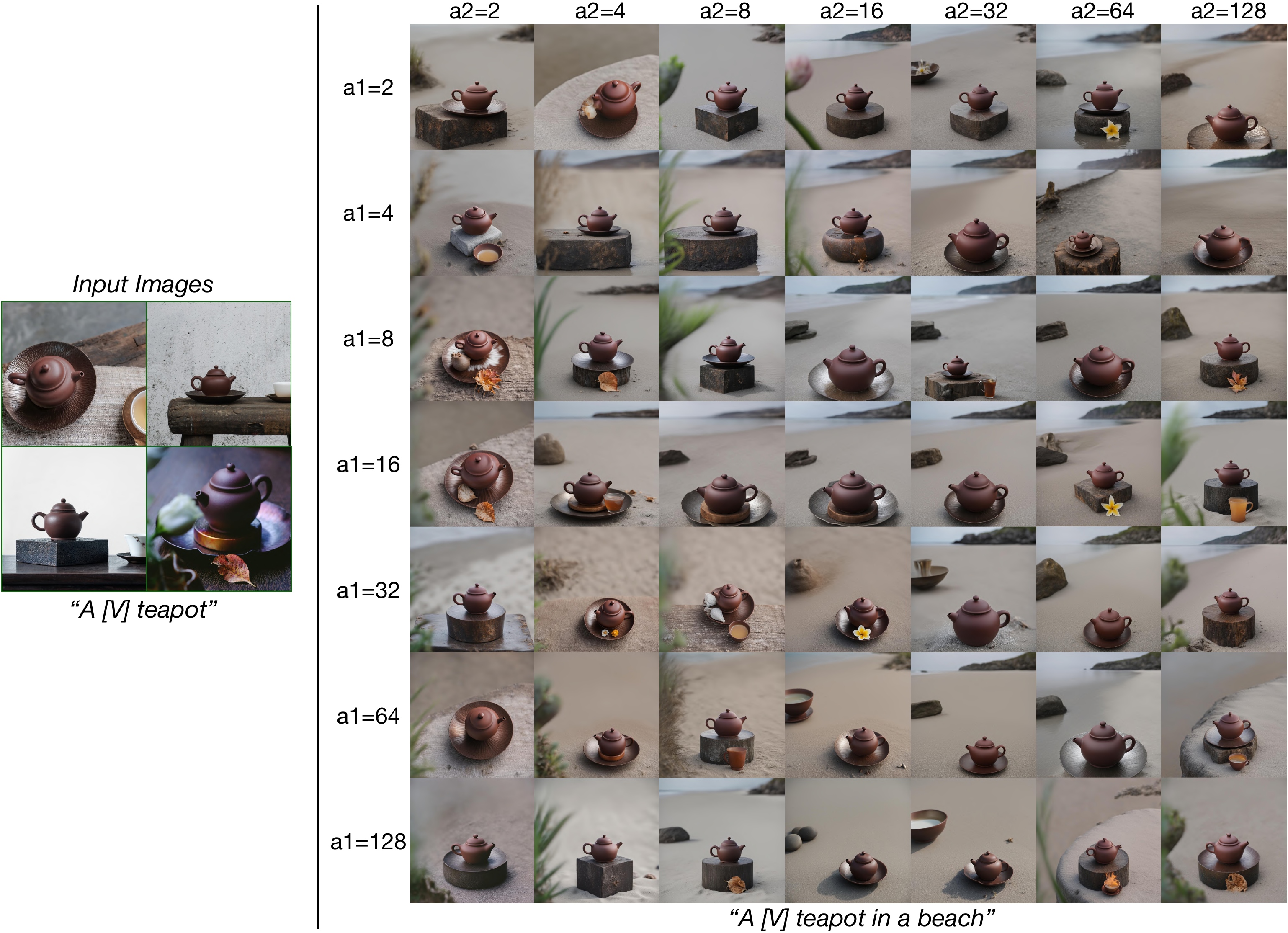}
    \end{subfigure}
    \caption{Effect of Kronecker factors \emph{i.e.}, $a_1$ and $a_2$ in image generations. Optimal selection of $a_1$ and $a_2$ considers \textbf{image fidelity} and \textbf{parameter count}. Following this, we choose $a_1$ and $a_2$ as 4 and 64, respectively, interpreting that the lower Kronecker factor ($A$) should have a lower dimension compared to the upper Kronecker factor ($B$).}
    \label{fig:kronecker_factors}
\end{figure*}

\myparagraph{How to initialize the Kronecker factors?} Initialization plays a crucial role in the fine-tuning process. Networks that are poorly initialized can prove challenging to train. Therefore, having a well-crafted initialization strategy is crucial for achieving effective fine-tuning. In our experiments, we explored three initialization methods: Normal initialization, Kaiming Uniform initialization~\cite{he2015init}, and Xavier initialization. These methods were applied to initialize the Kronecker factors $A_k$ and $B_k$. We observed that initializing both factors with the same type of initialization failed to preserve fidelity. Surprisingly, initializing $B_{k}$ with zero yielded the best results in the fine-tuning process.

As illustrated in~\cref{fig:init}, images initialized with ($A_k = \text{Normal}^{s}$, $B_k = 0$) and ($A_k = \text{KU}$, $B_k = 0$) produce the most favorable results, while images initialized with ($A_k = \text{Normal}^{s}$, $B_k = \text{Normal}^{s}$) and ($A_k = \text{XU}$, $B_k = \text{XU}$) result in the least satisfactory generations. Here, $s \in {1,2}$ denotes two different normal distributions - $\mathcal{N}\left(0, 1/a_2\right)$ and $\mathcal{N}\left(0, \sqrt{min(d, h)}\right)$ respectively, where $d$ and $h$ represents in features and out features dimension.

\myparagraph{Effect of size of Kronecker Factors.}
The size of the Kronecker factors significantly influences the images generated by \textit{DiffuseKronA}. Larger Kronecker factors tend to produce images with higher resolution and more detailing, while smaller Kronecker factors result in lower-resolution images with less detailing. Images generated with larger Kronecker factors tend to look more realistic, while those generated with smaller Kronecker factors appear more abstract. Varying the Kronecker factors can result in a wide range of images, from highly detailed and realistic to more abstract and lower resolution.

In ~\cref{fig:kronecker_factors} when both $a_1$ and $a_2$ are set to relatively high values (8 and 64 respectively), the generated images are of very high fidelity and detail. The features of the dog and the house in the background are more defined and realistic with the house having a blue colour as mentioned in the prompt. When $a_1$ is halved (4) while maintaining the same (64) results in images where the dog and the house are still quite detailed due to the high value of $a_2$, but perhaps less so than in the previous case due to the smaller value of $a_1$.
However, when the factors are small $\le$ 8, not only the generated images do not adhere to the prompt, but the number of trainable parameters increases drastically.

In ~\cref{tab:param_krona}, we present the count of trainable parameters corresponding to different Kronecker factors.

\begin{table*}[!ht]
  \centering
  \begin{tabular}{cccc}
  \resizebox{0.45\columnwidth}{!}{
    \begin{tabular}{c|c|c}
      \toprule
      \textbf{$a_{1}$} & \textbf{$a_{2}$} & \textbf{\# Parameters} \\
      \midrule
      \multirow{7}{*}{1} & 2 & 119399520\\
      & 4 & 59701440\\
      & 8 &  29854080\\
      & 16 & 14933760\\
      & 32 & 7480320\\
      & 64 & 3767040\\
      & 128 & 1937280\\
      \bottomrule
    \end{tabular}}
    &
      \resizebox{0.45\columnwidth}{!}{
    \begin{tabular}{c|c|c}
      \toprule
      \textbf{$a_{1}$} & \textbf{$a_{2}$} & \textbf{\# Parameters} \\
      \midrule
      \multirow{7}{*}{2} & 2 & 238799040\\
      & 4 & 119402880 \\
      & 8 & 59708160 \\
      & 16 & 29867520 \\
      & 32 & 14960640\\
      & 64 & 7534080\\
      & 128 & 3874560\\
      \bottomrule
    \end{tabular}} &
      \resizebox{0.45\columnwidth}{!}{
    \begin{tabular}{c|c|c}
    \toprule
      \textbf{$a_{1}$} & \textbf{$a_{2}$} & \textbf{\# Parameters} \\
      \midrule
      \multirow{7}{*}{4} & 2 & 119402880\\
      & 4 & 59708160\\
      & 8 & 29867520\\
      & 16 & 14960640\\
      & 32 & 7534080\\
      & 64 & 3874560\\
      & 128 & 2152320\\
      \bottomrule
    \end{tabular}} &
      \resizebox{0.45\columnwidth}{!}{
    \begin{tabular}{c|c|c}
    \toprule
      \textbf{$a_{1}$} & \textbf{$a_{2}$} & \textbf{\# Parameters} \\
      \midrule
      \multirow{7}{*}{8} & 2 & 59708160\\
      & 4 & 29867520\\
      & 8 & 14960640\\
      & 16 & 7534080\\
      & 32 & 3874560\\
      & 64 & 2152320\\
      & 128 & 1506240\\
      \bottomrule
    \end{tabular}} \\
      \resizebox{0.45\columnwidth}{!}{
    \begin{tabular}{c|c|c}
      \textbf{$a_{1}$} & \textbf{$a_{2}$} & \textbf{\# Parameters} \\
      \midrule
      \multirow{7}{*}{16} & 2 & 29867520\\
      & 4 & 14960640\\
      & 8 & 7534080\\
      & 16 & 3874560\\
      & 32 & 2152320\\
      & 64 & 1506240\\
      & 128 & 1613280\\
      \bottomrule
    \end{tabular}} &
      \resizebox{0.45\columnwidth}{!}{
    \begin{tabular}{c|c|c}
      \textbf{$a_{1}$} & \textbf{$a_{2}$} & \textbf{\# Parameters} \\
      \midrule
      \multirow{7}{*}{32} & 2 & 14960640\\
      & 4 & 7534080 \\   
      & 8 & 3874560\\
      & 16 & 2152320\\
      & 32 & 1506240\\
      & 64 & 1613280\\
      & 128 & 2526960\\
      \bottomrule
    \end{tabular}} &
      \resizebox{0.45\columnwidth}{!}{
    \begin{tabular}{c|c|c}
      \textbf{$a_{1}$} & \textbf{$a_{2}$} & \textbf{\# Parameters} \\
      \midrule
      \multirow{7}{*}{64} & 2 & 7534080\\
      & 4 & 3874560\\
      & 8 & 2152320\\
      & 16 & 1506240\\
      & 32 & 1613280\\
      & 64 & 2526960\\
      & 128 & 4704120\\
      \bottomrule
    \end{tabular}} &
      \resizebox{0.45\columnwidth}{!}{
    \begin{tabular}{c|c|c}
      \textbf{$a_{1}$} & \textbf{$a_{2}$} & \textbf{\# Parameters} \\
      \midrule
      \multirow{7}{*}{\footnotesize{128}} & 2 & 3874560\\
      & 4 & 2152320\\
      & 8 &  1506240\\
      & 16 & 1613280\\
      & 32 & 2526960\\
      & 64 & 4704120\\
      & 128 & 9233340\\
      \bottomrule
    \end{tabular}}
  \end{tabular}
  \vspace*{-2mm}
  \caption{Effect of the size of Kronecker factors (\emph{i.e.} $a_{1}$ \& $a_2$) in terms of trainable parameter count.}
  \label{tab:param_krona}
\end{table*}

\begin{figure*}[!htbp]
  \centering
  \includegraphics[width=1\textwidth]{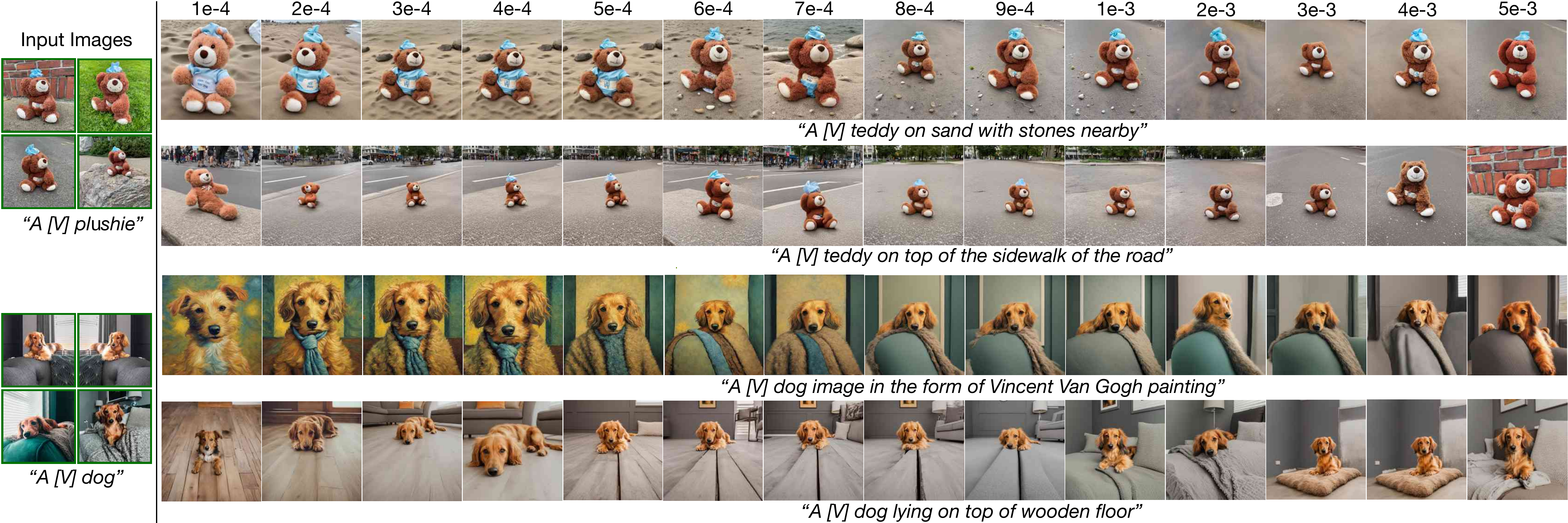}
  \vspace*{-8mm}
  \caption{Effect of learning rate on subject fidelity and text adherence. The most favorable results are obtained using learning rate $5\times10^{-4}$.}
  \label{fig:lr}
\end{figure*}

\subsection{Effect of learning rate}
\label{ssec:effect_learning_rate}
The learning rate factor influences the alignment of generated images towards both text prompts and input images. Our approach yields better results when using learning rates near $5 \times 10^{-4}$. Higher learning rates, typically around $10^{-3}$, compel the model to overfit, resulting in images closely mirroring the input images and largely ignoring the input text prompts. Conversely, lower learning rates, below $10^{-4}$, cause the model to overlook the input images, concentrating solely on the provided input text.

In ~\cref{fig:lr}, for \underline{\textit{``A [V] teddy on sand with stones nearby''}} when the learning rate is $\ge 1 \times 10^{-3}$, the generated teddy bears closely resemble the input images. Additionally, the sand dunes in the images vanish, along with the removal of stones. Conversely, for learning rates in the intermediate ranges, the sand dunes and pebbles remain distinctly visible.
In the context of \underline{\textit{``A [V] dog image in the form of}} \underline{\textit{a Vincent Van Gogh painting''}} in ~\cref{fig:lr}, images close to the rightmost edge lack a discernible painting style, appearing too similar to the input images. Conversely, images near the leftmost side exhibit a complete sense of Van Gogh's style but lack the features present in the input images. Notably, in the images positioned in the middle, there is an excellent fusion of the painting style and the features of the input images.

\begin{figure*}[!ht]
    \centering
    \vspace*{-1mm}
    \includegraphics[width=0.9\textwidth]{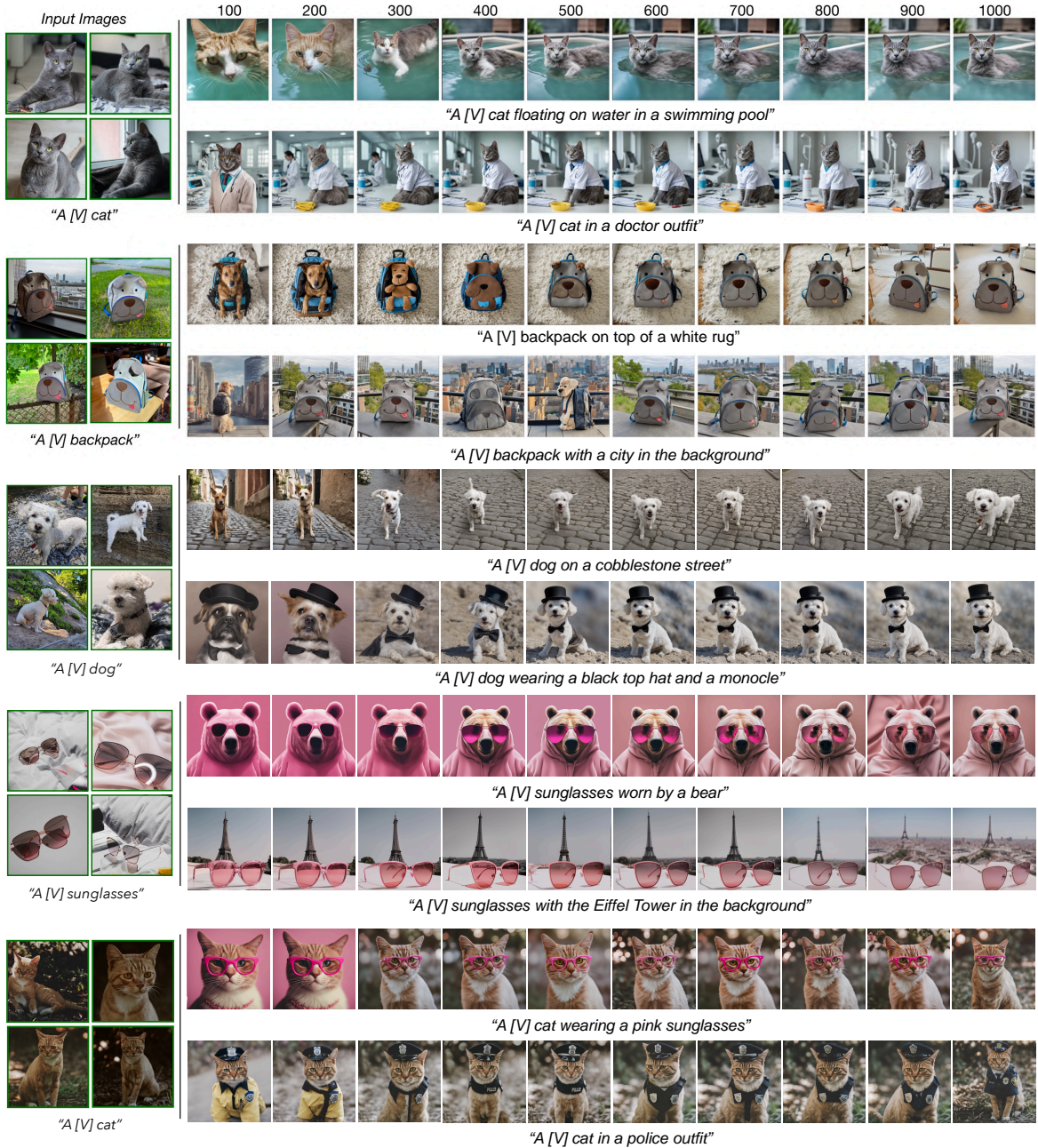}
    \vspace*{-4mm}
    \caption{Effect of training steps in image generation on SDXL. In the case of simple prompts (row 1), \textit{DiffuseKronA} consistently delivers favorable results between steps 500 and 1000. Conversely, for more complex prompts (row 2), reaching the desired outcome might necessitate waiting until after step 1000.}
  \label{fig:iter_qual}
\end{figure*}

\begin{figure*}[!ht]
  \centering
  \vspace*{-1.5mm}
  \includegraphics[width=0.93\textwidth]{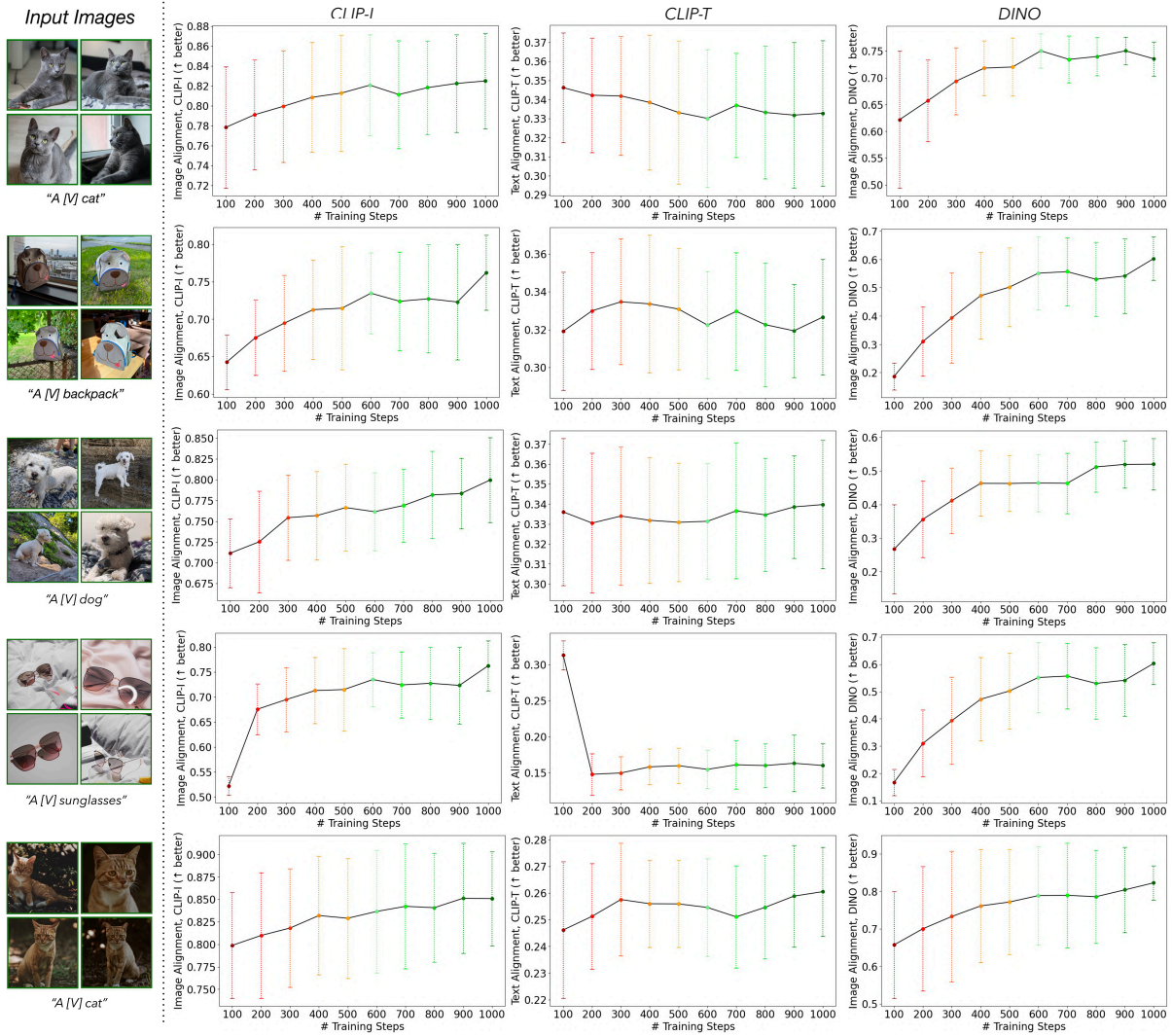}
  \vspace*{-3.5mm}
  \caption{Plots depicting image alignment, text alignment, and DINO scores against training iterations. The scores are computed from the same set of images and prompts as depicted in~\cref{fig:iter_qual}.}
  \label{fig:iter_quant}
\end{figure*}

\begin{figure*}[!h]
    \centering
    \hspace*{5mm}
    \includegraphics[width=0.85\textwidth]{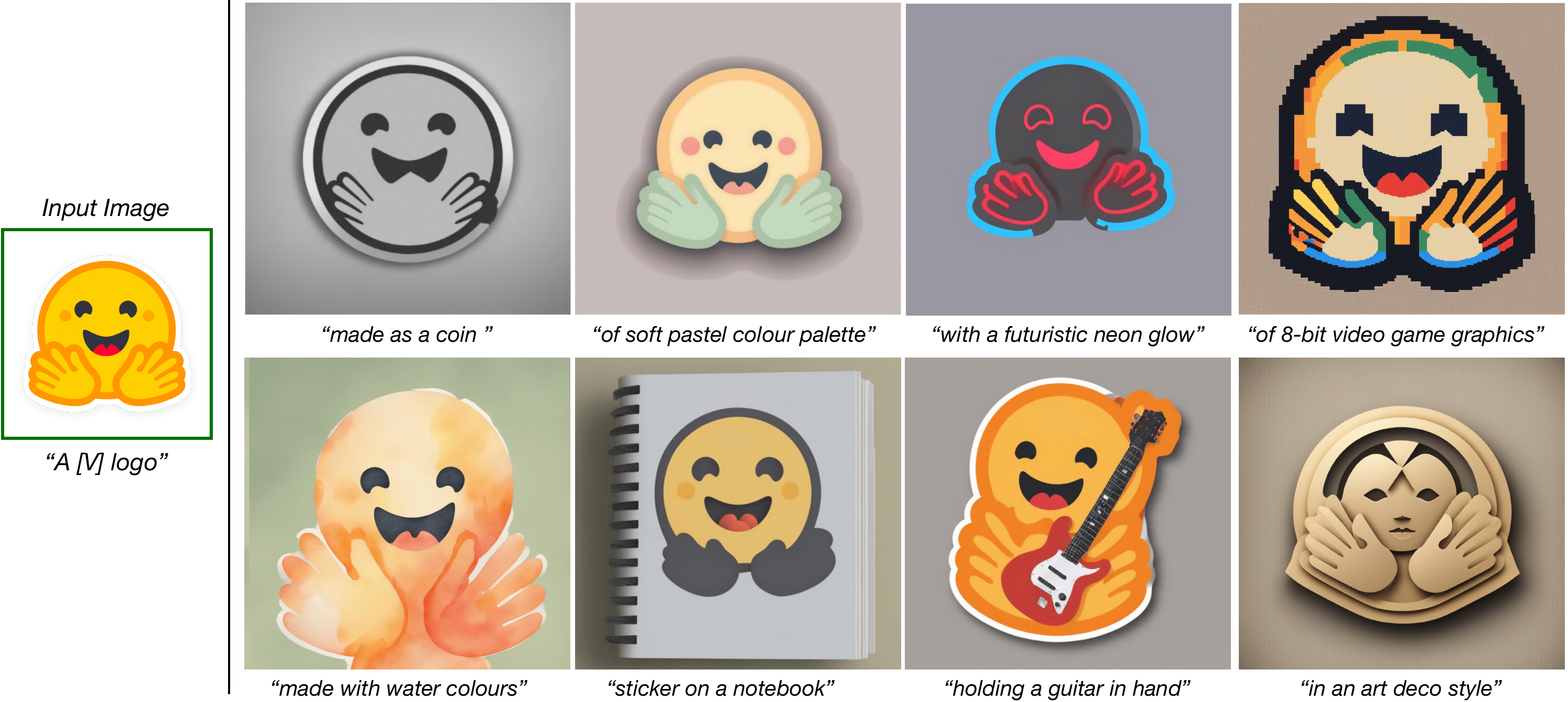}
    \vspace*{-3mm}
    \caption{One-shot image generation results showcase the remarkable effectiveness of \textit{DiffuseKronA} while preserving high fidelity and better text alignment.}
    \vspace*{-2mm}
  \label{fig:one-shot}
\end{figure*}

\subsection{Effect of training steps}
\label{ssec:effect_training_steps} 
In T2I personalization, the timely attainment of satisfactory results within a specific number of iterations is crucial. This not only reduces the overall training time but also helps prevent overfitting to the training images, ensuring efficiency and higher fidelity in image generation. With SDXL, we successfully generate desired-fidelity images within 500 iterations, if the input images and prompt complexity are not very high. However, in cases where the input image complexity or the prompt complexity requires additional refinement, it is better to extend the training up to 1000 iterations as depicted in ~\cref{fig:iter_qual} and~\cref{fig:iter_quant}.\\
The images generated by \textit{DiffuseKronA} show a clear progression in quality with respect to different steps. As the steps increase, the model seems to refine the details and improve the quality of the images. This iterative process allows the model to gradually improve the image, adding more details and making it more accurate to the prompt.

In ~\cref{fig:iter_qual} for instance, \underline{\textit{``A cat floating on water in a}} \underline{\textit{swimming pool''}}, in the initial iterations, the model generates a basic image of a cat floating on water. As the iterations progress and reach 500, the model refines the image, adding more details such as the color and texture of the cat, the ripples in the water, and the details of the swimming pool. At 1000 steps the image is a detailed and realistic representation of the prompt. 

In ~\cref{fig:iter_qual}, \underline{\textit{``A backpack on top of a white rug''}}, the early iterations produce a simple image of a backpack on a white surface. However, as the iterations increase, the model adds more details to the backpack, such as the zippers, pockets, and straps. It also starts to add texture to the white rug, making it look more realistic. By the final iteration, the white rug gets smoother in texture producing a fine image.

\begin{figure*}[!ht]
    \centering
    \includegraphics[width=0.7\textwidth]{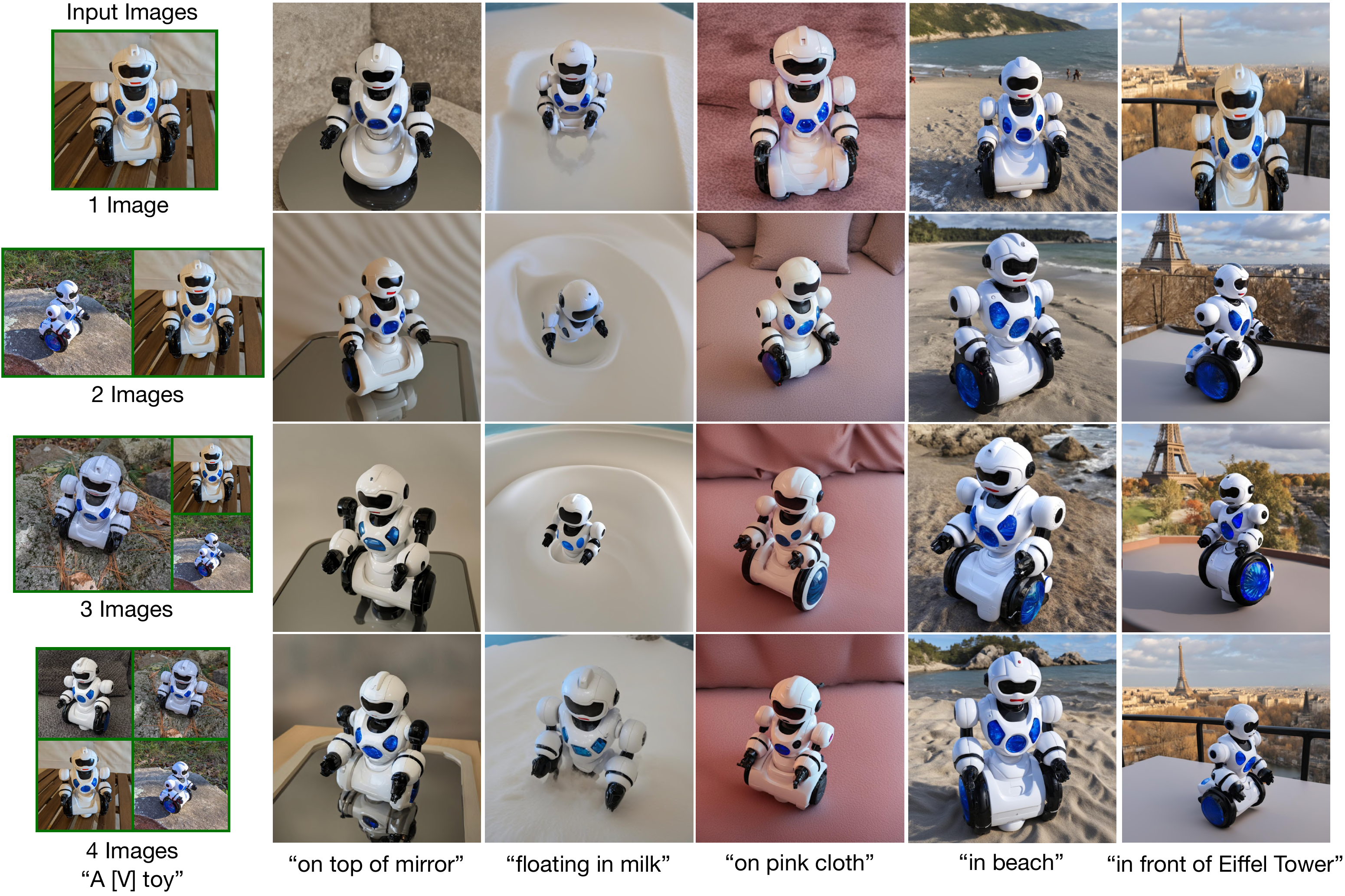}
    \hspace*{2pt} 
    \includegraphics[width=0.7\textwidth]
    {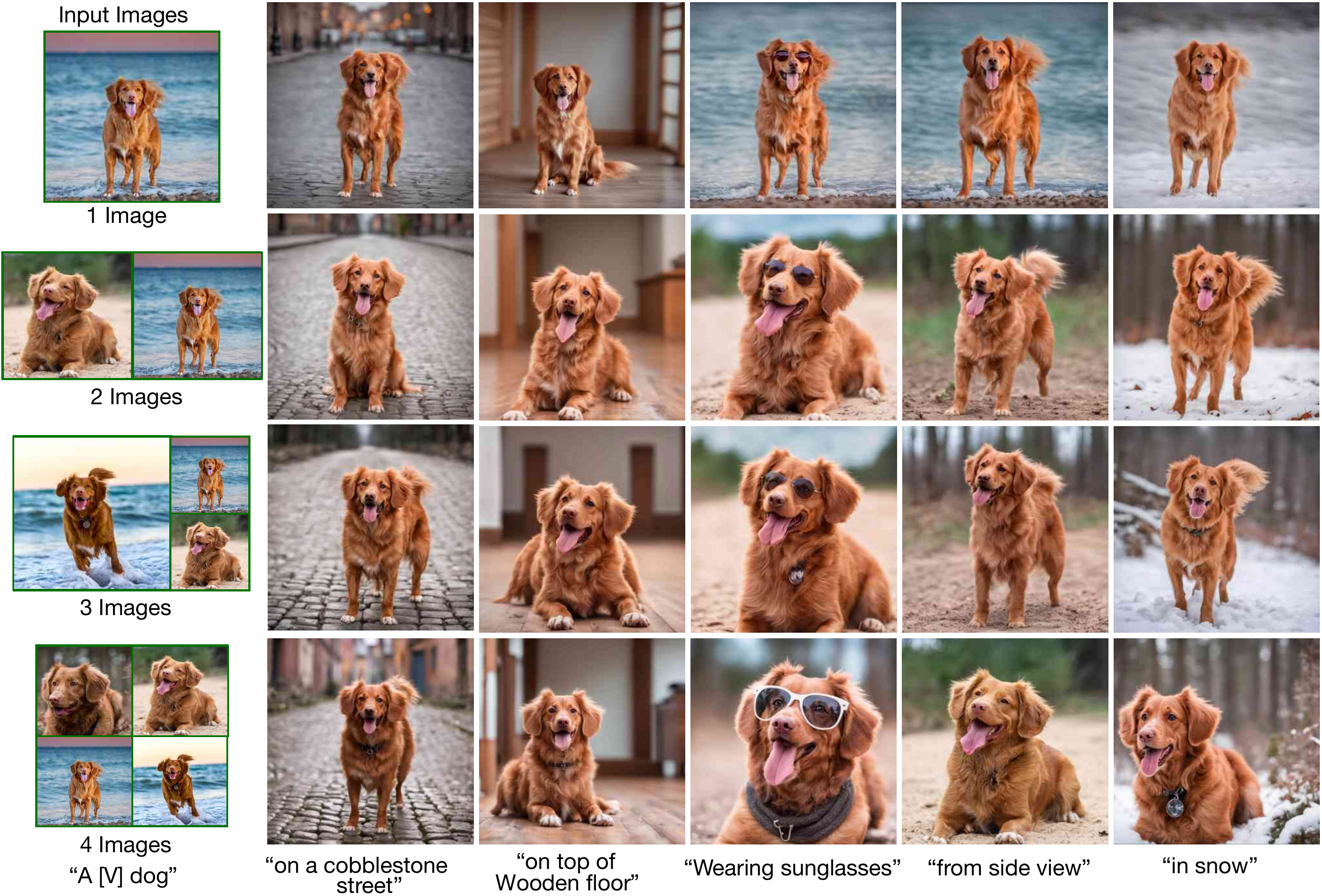}
    \vspace*{-3mm}
    \caption{The influence of training images on fine-tuning. Even though \textit{DiffuseKronA} produces impressive results with a single image, the generation of images with a broader range of perspectives is enhanced when more training images are provided with variations.}
  \label{fig:train_images}
\end{figure*}

\subsection{Effect of the number of training images}
\label{ssec:effect_training_images}
\subsubsection{One shot image generation}
\label{sssec:one_shot_performance}
The images are high-quality and accurately represent the text prompts. They are clear and well-drawn, and the content of each image matches the corresponding text prompt perfectly.
For instance, in ~\cref{fig:one-shot}, the image of the \underline{\textit{``A [V] logo''}} is a yellow smiley face with hands. The \underline{\textit{``made as a coin''}} prompt resulted in a grey ghost with a white border, demonstrating the model’s ability to incorporate abstract concepts. The \underline{\textit{``futuristic neon glow''}} and \underline{\textit{``made with watercolours''}} prompts resulted in a pink and a yellow octopus respectively, showcasing the model’s versatility in applying different artistic styles. The model’s ability to generate an image of a guitar-playing octopus on a grey notebook from the prompt \underline{\textit{``sticker on a notebook''}} is a testament to its advanced capabilities.

The images are diverse in terms of style and content which is impressive, especially considering that these images were generated in a one-shot setting which makes it suitable for image editing tasks. While our model demonstrates remarkable proficiency in generating compelling results with a single input image, it encounters challenges when attempting to generate diverse poses or angles. However, when supplied with multiple images (2, 3, or 4), our model adeptly captures additional spatial features from the input images, facilitating the generation of images with a broader range of poses and angles. Our model can effectively use the information from multiple input images to generate more accurate and detailed output images as depicted in~\cref{fig:train_images}.

\subsection{Effect of Inference Hyperparameters}
\label{ssec:effect_inference_hyperparameters}
\myparagraph{Guidance Score ($\alpha$).}
The guidance score, denoted as $\alpha$, regulates the variation and distribution of colors in the generated images. A lower guidance score produces a more subdued version of colors in the images, aligning with the description provided in the input text prompt. In contrast, a higher guidance score results in images with more vibrant and pronounced colors. Guidance scores ranging from 7 to 10 generally yield images with an appropriate and well-distributed color palette.

\begin{figure*}[!ht]
  \centering
  \includegraphics[width=0.88\textwidth]{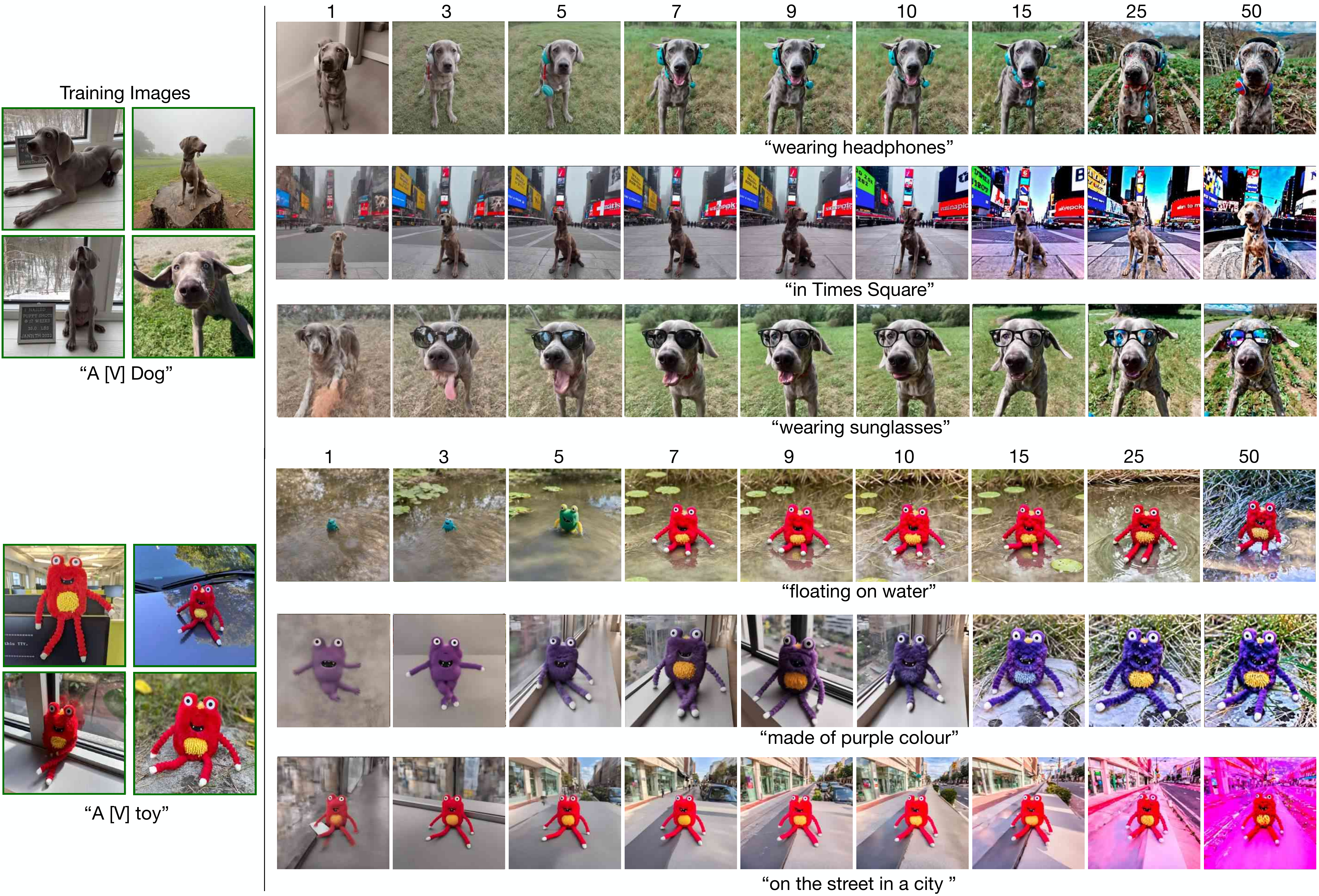}
  \vspace*{-4mm}
  \caption{Images produced by adjusting the guidance score ($\alpha$) reveal that a score of 7 produces the most realistic results. Increasing the score beyond 7 significantly amplifies the contrast of the images.}
  \label{fig:guidance}
\end{figure*}

In the example of \underline{\textit{``A [V] toy''}} in ~\cref{fig:guidance}, when the prompt is \underline{\textit{``made of purple color''}}, it is evident that a reddish lavender hue is generated for a guidance score of 1 or 3. Conversely, with a guidance score exceeding 15, a mulberry shade is produced. For guidance scores close to 8, images with a pure purple color are formed.

\begin{figure*}[!h]
  \centering
    \includegraphics[width=0.88\textwidth]{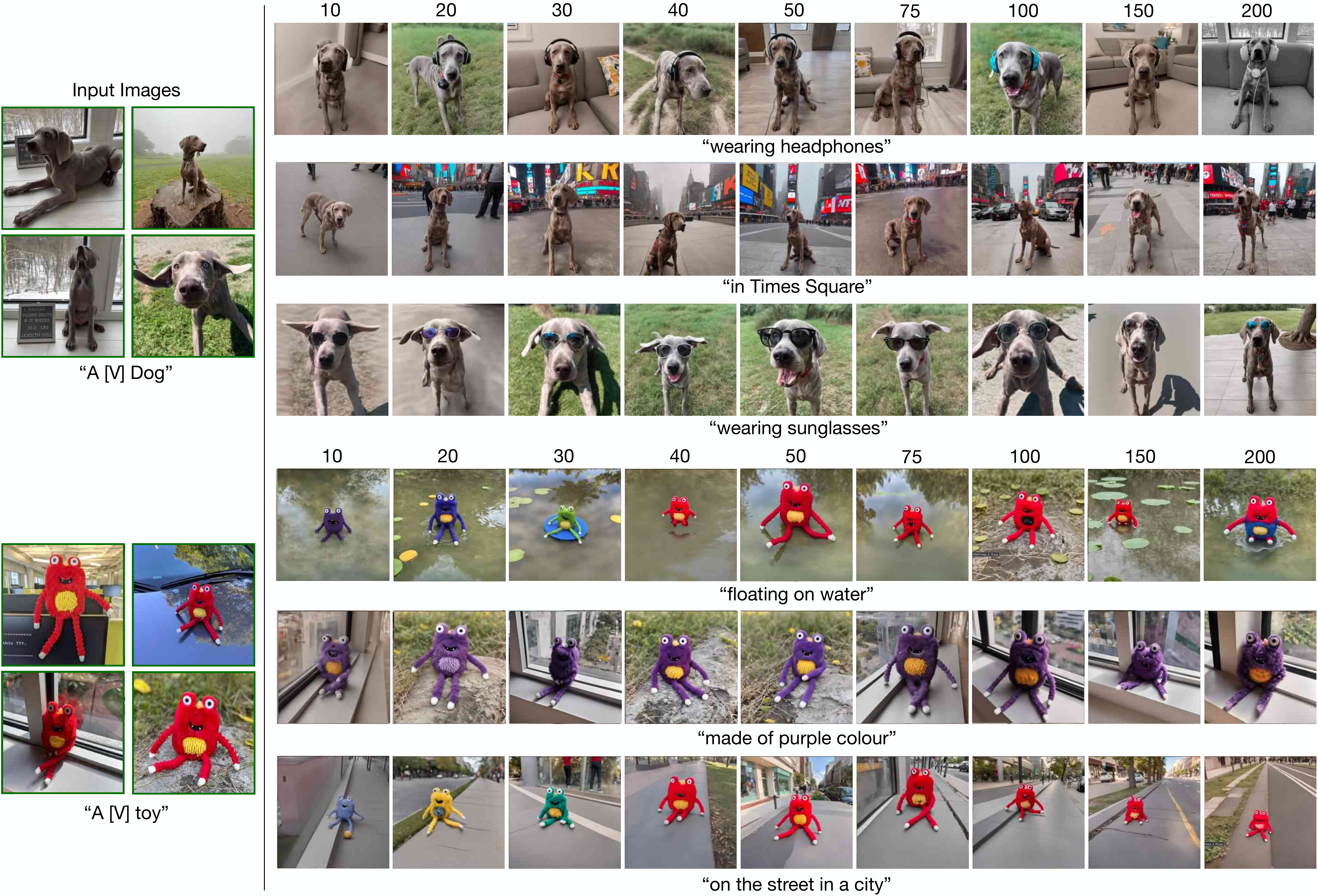}
    \vspace*{-4mm}
    \caption{The influence of inference steps on image generation. Optimal results are achieved in the range of 50-70 steps, striking a balance between textual input and subject fidelity. Here, we opted for 50 inference steps to minimize inference time.}
  \label{fig:inf_step}
\end{figure*}

\myparagraph{Number of inference Steps.}
The number of steps plays a crucial role in defining the granularity of the generated images. As illustrated in ~\cref{fig:inf_step}, during the initial steps, the model creates a subject that aligns with the text prompt and begins incorporating features from the input image. With the progression of generation, finer details emerge in the images. Optimal results, depending on the complexity of prompts, are observed within the range of 30 to 70 steps, with an average of 50 steps proving to be the most effective. However, exceeding 100 steps results in the introduction of noise and a decline in the quality of the generated images.

The quality of the generated images appears to improve with an increase in the number of inference steps. For instance, the images for the prompt \underline{\textit{``a toy''}} and \underline{\textit{``wearing sunglasses''}} appear to be of higher quality at 50 and 75 inference steps respectively, compared to at 10 inference steps.

\section{Detailed study on LoRA-DreamBooth vs \textit{DiffuseKronA}}
\label{sec:detailed_lora_vs_our}
In this section, we expand our analysis of model performance (from ~\cref{ssec:lora_vs_krona}), comparing LoRA-DreamBooth and \textit{DiffuseKronA} across various aspects, including fidelity, color distribution, text alignment, stability, and complexity.

\subsection{Multiplicative Rank Property and Gradient Updates}
Let $A$ and $B$ be $m \times n$ and $p \times q$ matrices respectively. Suppose that $A$ has rank $r$ and $B$ has rank $s$.  

\begin{theorem}
Ranks for dot product are bound by the rank of multiplicand and multiplier, i.e. $rank(A \cdot B) = \min(rank(A),\:rank(B)) = \min(r,\:s)$.
\end{theorem}

\begin{theorem} 
Ranks for Kronecker products are multiplicative i.e. $rank(A \otimes B) = rank(A) \times rank(B) = r \times s$.
\end{theorem}

Since the Kronecker Product has the advantage of multiplicative rank, it has a better representation of the underlying distribution of images as compared to the dot product.

Another notable difference between the Low-rank decomposition (LoRA) and the Kronecker product is when computing the derivatives, denoted by $d\left(\cdot\right)$. In the case of LoRA, $d\left(A \cdot B\right) = d\left(A\right) \cdot B + A \cdot d\left(B\right)$. But in the case of the Kronecker product, $d\left(A \otimes B\right) = d\left(A\right) \otimes d\left(B\right)$. The gradient updates in LoRA are direct without a structured relationship, whereas the Kronecker product preserves the structure during an update. While a dot product is simpler and LoRA updates each parameter independently, a Kronecker product introduces structured updates that can be beneficial when preserving relationships between parameters stored in $A$ and $B$.

\subsection{Fidelity \& Color Distribution}
\textit{DiffuseKronA} generates images of superior fidelity as compared to LoRA-DreamBooth in lieu of the higher representational power of Kronecker Products along with its ability to capture spatial features.

\begin{figure*}[!ht]
    \centering
    \includegraphics[scale=0.15]{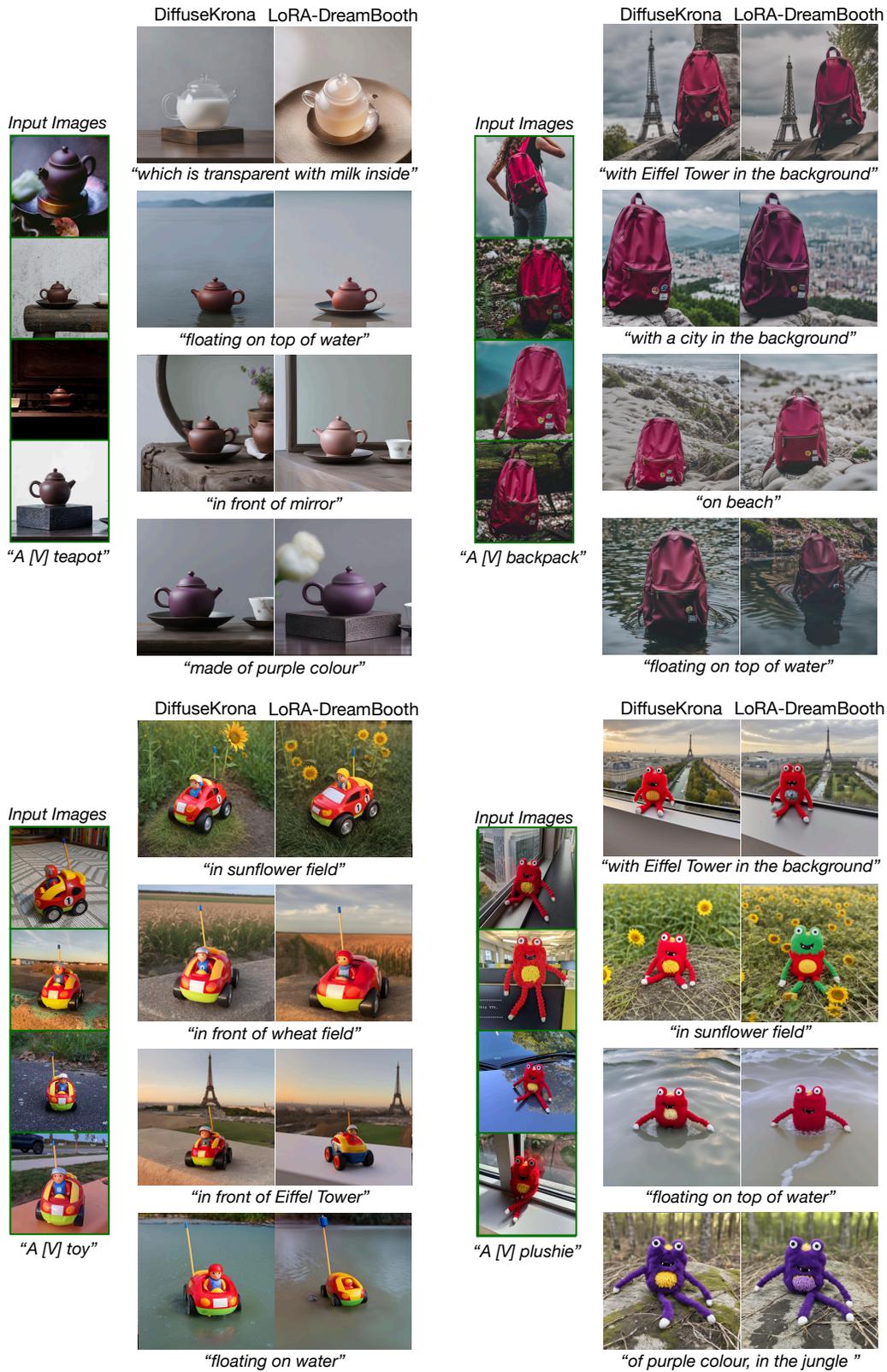}
    \vspace*{-3mm}
    \caption{Comparison of fidelity and color preservation in \textit{DiffuseKronA} and LoRA-DreamBooth.}
    \label{fig:fidelity_colour}
\end{figure*}

In the example of \underline{\textit{``A [V] backpack''}} in ~\cref{fig:fidelity_colour}, the following observations can be made:

(1) \textbf{\textit{``with the Eiffel Tower in the background''}}: The backpack generated by \textit{DiffuseKronA} is pictured with the Eiffel Tower in the background, creating a striking contrast between the red of the backpack and the muted colors of the cityscape, which LoRA-DreamBooth fails to do.

(2) \textbf{\textit{``city in background''}}: The backpack generated by \textit{DiffuseKronA} is set against a city backdrop, where the red color stands out against the neutral tones of the buildings, whereas, LoRA-DreamBooth does not generate high contrast between images.

(3) \textbf{\textit{``on the beach''}}: The image generated by \textit{DiffuseKronA} shows the backpack on a beach, where the red contrasts with the blue of the water and the beige of the sand.

\begin{figure*}[!ht]
    \centering
    \includegraphics[width=0.98\textwidth]{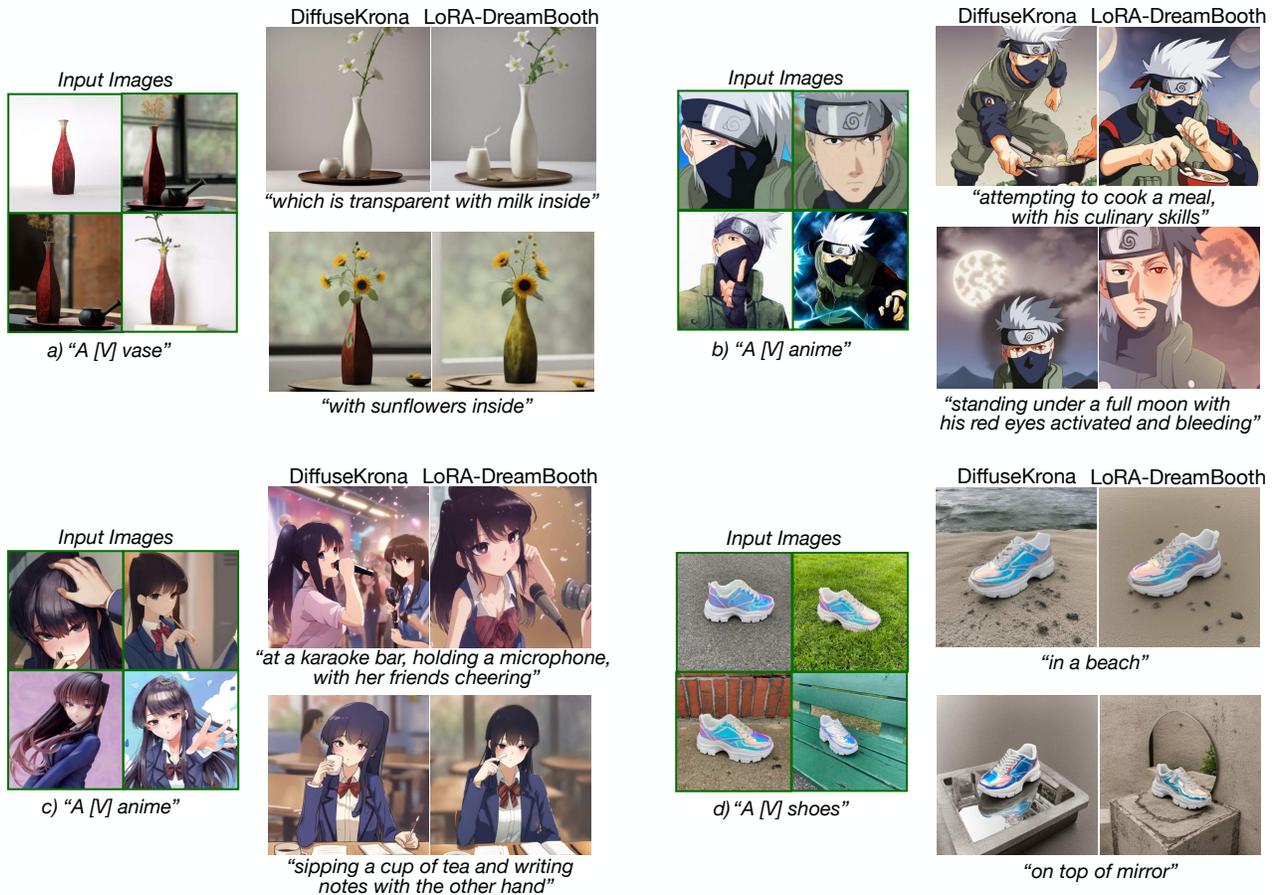}
    \vspace*{-3mm}
    \caption{Comparison of text alignment in generated images by our proposed \textit{DiffuseKronA} and LoRA-DreamBooth.}
  \label{fig:text_align}
\end{figure*}

\begin{figure*}[!ht]
    \centering
    \includegraphics[width=1\textwidth]{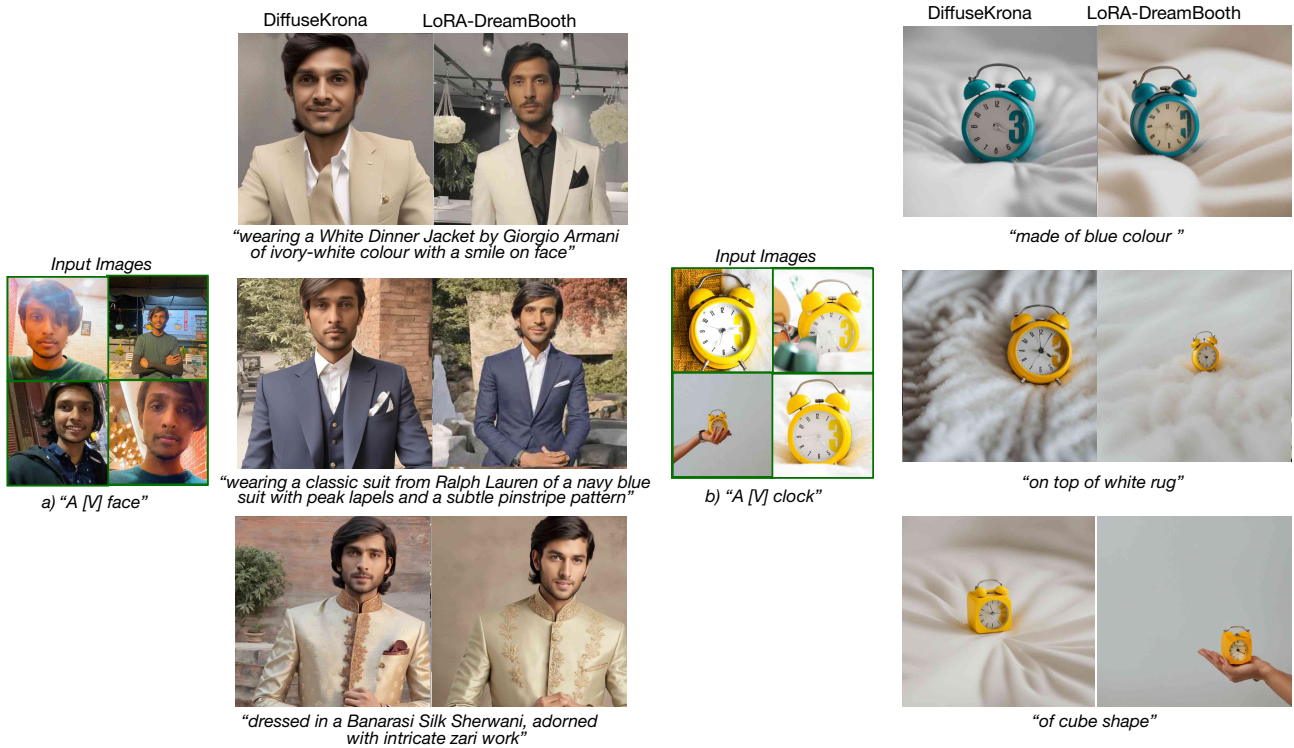}
    \caption{Comparison of image generation on complex prompts and input images by \textit{DiffuseKronA} and LoRA-DreamBooth.}
    \label{fig:complex_im_tx}
\end{figure*}

\subsection{Text Alignment}
\textit{DiffuseKronA} is more accurate in aligning text with images compared to the Lora-DreamBooth. For instance, in the first row, \textit{DiffuseKronA} correctly aligns the text with \underline{\textit{``sunflowers inside''}} with the image of a vase with sunflowers, whereas LoRA-DreamBooth fails to align the sunflower in the vase of the same color as of input images.

In more complex input examples like in ~\cref{fig:text_align}, such as the one involving anime in \underline{\textit{``A [V] character''}}, the generated images by LoRA-DreamBooth lack the sense of cooking a meal and a karaoke bar, whereas \textit{DiffuseKronA} consistently produces images that closely align with the provided text prompts.

\subsection{Complex Input images and Prompts}
\textit{DiffuseKronA} demonstrates a notable emphasis on capturing nuances within text prompts and excels in preserving intricate details from input images to the highest degree. In contrast, LoRA-DreamBooth lacks these properties. This distinction is evident in ~\cref{fig:complex_im_tx}, where, for the prompt \underline{\textit{``A [V] face''}}, \textit{DiffuseKronA} successfully generates an ivory-white blazer and a smiling face, while LoRA-DreamBooth struggles to maintain both the color and the smile on the face.

Similarly, for the prompt \underline{\textit{``A [V] clock''}} in ~\cref{fig:complex_im_tx}, \textit{DiffuseKronA} accurately reproduces detailed numbers, particularly 3, from the input images. Although it encounters challenges in preserving the structure of numbers while creating a clock of cubical shape, it still maintains a strong focus on text details— a characteristic lacking in LoRA-DreamBooth.


\subsection{Qualitative and Quantitative comparison}
We have assessed the image generation capabilities of \textit{DiffuseKronA} and LoRA-DreamBooth on SDXL~\cite{sdxl}. Our findings reveal that \textit{DiffuseKronA} excels in generating images with high fidelity, more accurate color distribution, and greater stability compared to LoRA-DreamBooth.

\section{Comparison with other Low-Rank Decomposition methods}
\label{sec:low-rank-other}

In this section, we compare our \textit{DiffuseKronA} with low-rank methods other than LoRA, specifically with LoKr~\cite{lokr} and LoHA~\cite{lokr}. We also note that our implementation is independent of the \texttt{LyCORIS} project~\cite{lokr}, and we did not use LoKr nor LoHA in \textit{DiffuseKronA}\footnote{To ensure a fair comparison, we have incorporated LoKr and LoHA into the SDXL backbone.}. We summarize the key differences between \textit{DiffuseKronA} and these methods as follows:

\ding{182} \textit{DiffuseKronA} has 2 controllable parameters ($a_1$ and $a_2$), which are chosen manually through extensive experiments (refer to~\cref{fig:kronecker_factors} and~\cref{tab:param_krona}), whereas LoKr~\cite{lokr} follows the procedure mentioned in the \textsc{factorization} function (see right) which depends on input dimension and another hyper-parameter called \textit{factor}. Following the descriptions on the implementation of Figure 2 in~\cite{lokr}, and we quote ``we set the factor to 8 and do not perform further decomposition of the second block'', the default implementation makes $A$ a square matrix of dimension $\left(\textit{factor}\times\textit{factor}\right)$. Notably, for any factor, $f>0$, $A$ would always be a square matrix of shape $\left(f\times f\right)$ which is a special case (a subset) of \textit{DiffusKronA} (diagonal entry in~\cref{fig:kronecker_factors}) but for $f=-1$, $A$ matrix size would be completely dependent upon dimension, and it would not be a square matrix always.
\vspace{-2mm}
\begin{lstlisting}[language=Python, caption={This code snippet is extracted from the official \texttt{LyCORIS} codebase (\href{https://github.com/KohakuBlueleaf/LyCORIS/blob/main/lycoris/modules/lokr.py\#L10}{Link}).}]
def factorization(dim: int, factor: int = -1) -> tuple[int, int]:

    if factor > 0 and (dim % factor) == 0:
        m = factor
        n = dim // factor
        if m > n:
            n, m = m, n
        return m, n
    if factor < 0:
        factor = dim
    m, n = 1, dim
    length = m + n
    while m < n:
        new_m = m + 1
        while dim % new_m != 0:
            new_m += 1
        new_n = dim // new_m
        if new_m + new_n > length or new_m > factor:
            break
        else:
            m, n = new_m, new_n
    if m > n:
        n, m = m, n
    return m, n
\end{lstlisting}

These attributes make our way of performing Kronecker decomposition a superset of LoKr, offering greater control and flexibility compared to LoKr. On the other hand, LoHA has only one controllable parameter, \emph{i.e.}, rank, similar to LoRA.


\ding{183} LoKr takes the generic form of $\Delta W = A \otimes (B \cdot C)$, and LoHA adopts $\Delta W = (A\cdot B) \odot (C \cdot D) $, where $\odot$ denotes the Hadamard product. For more details, we refer the readers to Figure 1 in~\cite{lokr}. Based on the definition, LoHA does not explore the benefits of using Kronecker decomposition.

\ding{184} \citet{lokr} provided the first use of Kronecker decomposition in Diffusion model fine-tuning but limited analysis in the few-shot T2I personalization setting. In our study, we conducted detailed analysis and exploration to demonstrate the benefits of using Kronecker decomposition. Our new insights include large-scale analysis of parameter efficiency, enhanced stability to hyperparameters, and improved text alignment and fidelity, among others.

\begin{figure*}[!htbp]
  \centering
  \includegraphics[width=1\textwidth]{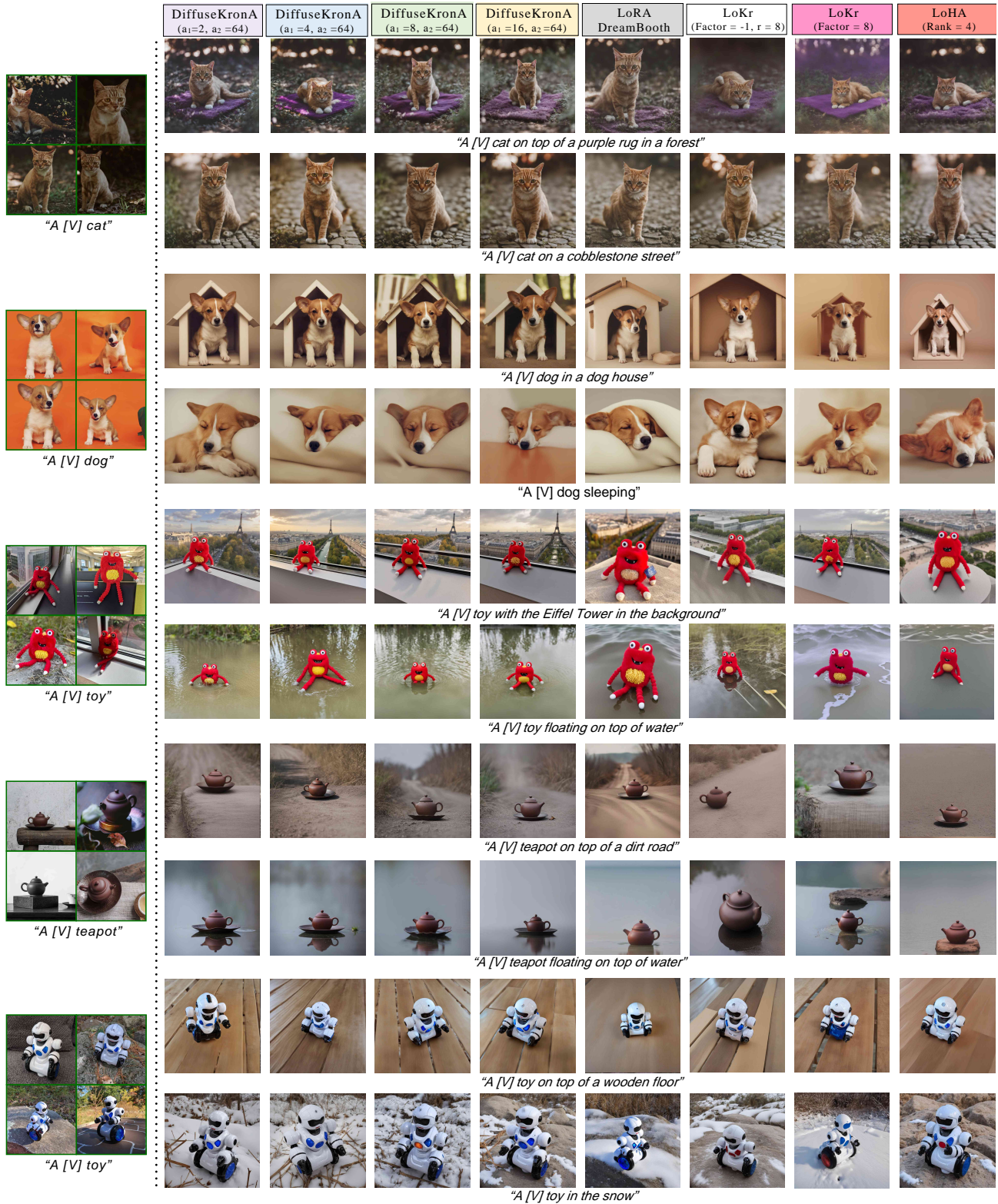}
  \vspace{-5mm}
  \caption{\textbf{Qualitative comparison} of four variants of \textit{DiffusekronA} with other low-rank methods including LoRA, LoKr, and LoHA. Learning rates: \textit{DiffusekronA} ($5\times\mathrm{10}^{-4}$), LoRA ($1\times\mathrm{10}^{-4}$), LoKr ($1\times\mathrm{10}^{-3}$) \& LoHA ($1\times\mathrm{10}^{-4}$).}
  \label{fig:comp_lora_lokr_loha}
\end{figure*}

\begin{figure*}[!htbp]
    \centering
    \includegraphics[width=0.8\textwidth]{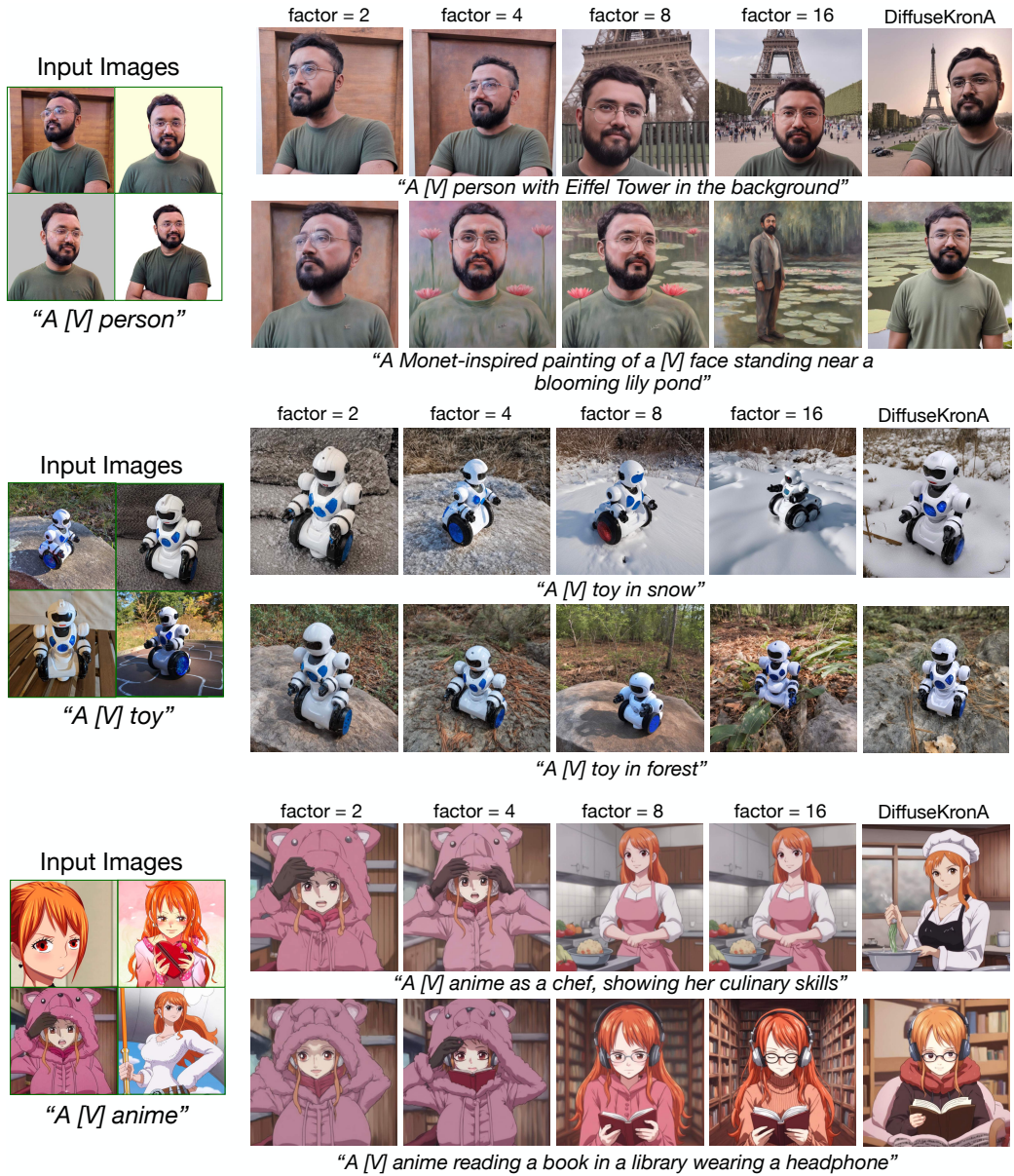}
    \vspace{-3mm}
    \caption{\textbf{Qualitative comparison.} Results are shown for the default factors given by the LoKr implementation, with the varying factors being 2, 4, 8, and 16.}
    \vspace{-2mm}
  \label{fig:comp_lokr_default}
\end{figure*}

\ding{185} We further compare our \textit{DiffuseKronA} with LoKr and LoHA using the default implementations from \cite{lokr} in~\cref{fig:comp_lora_lokr_loha} and~\cref{fig:comp_lokr_default}, respectively. However, the default settings were used in the SD variant, and it is also evident that personalized T2I generations are very sensitive to model settings and hyper-parameter choices. Bearing these facts, we also explored the hyperparameters in both adapters. In~\cref{fig:ablation_lokr_factor_rank}, we have presented the ablation study examining the factors and ranks for LoKr utilizing SDXL, while in~\cref{fig:ablation_lokr_lr}, we showcase an ablation study on the learning rate. Moreover, \cref{fig:ablation_loha} features an ablation study on the learning rate and rank for LoHA using SDXL. These analyses reveal that for LoKr, the optimal factor is -1 and the optimal rank is 8, with a learning rate of $1\times10^{-3}$; while for LoHA, the optimal rank is 4, with a learning rate of $1\times10^{-4}$. 

Additionally, quantitative comparisons are conducted, encompassing parameter count alongside image-to-image and image-to-text alignment scores, as detailed in \cref{tab:comp_lora_lokr_loha_quant} and \cref{tab:comp_lokr_quant}. The results in~\cref{tab:comp_lora_lokr_loha_quant} indicate that although LoKr marginally possesses fewer parameters still \textit{DiffuseKronA} with $a_1=16$ achieves superior CLIP-I, CLIP-T, and DINO scores. This contrast is readily noticeable in the visual examples depicted in~\cref{fig:comp_lora_lokr_loha}. For the prompt \underline{\textit{``A [V] toy with the Eiffel Tower in the background''}}, LoKr fails to construct the \textit{Eiffel Tower} in the background, unlike \textit{DiffuseKronA} ($a_1=16$). Similarly, in the case of \underline{\textit{``A [V] teapot floating on top of water''}} LoKr distorts the teapot's spout, whereas \textit{DiffuseKronA} maintains fidelity. In the case of \underline{\textit{``A [V] toy''}} (last row), the results of \textit{DiffuseKronA} are much more aligned as compared to LoKr for both prompts. Conversely, for \textit{dog} and \textit{cat} examples, all the methods demonstrate similar visual appearance in terms of fidelity as well as textual alignment. Consequently, it's evident that while LoKr reduces parameter count, it struggles with complex input images or text prompts with multiple contexts. Hence, \textit{DiffusekronA} achieves efficiency in parameters while upholding average scores across CLIP-I, CLIP-T, and DINO metrics. Hence, achieving a better trade-off between parameter efficiency and personalized image generation. 


    

\begin{table}[h]
    \centering
    \resizebox{0.48\textwidth}{!}{%
    \begin{tabular}{c|c|c|c|c}
    \toprule
        \textsc{Model} & \textsc{\# Parameters} ($\downarrow$) & \textsc{CLIP-I } ($\uparrow$) & \textsc{CLIP-T} ($\uparrow$) & \textsc{DINO} ($\uparrow$)\\\toprule
        
        \textbf{\textit{DiffuseKronA}} & \multirow{2}{*}{3.8 M} & 0.799 &0.267 &0.648 \\ 
        $a_1=2$ && $\pm$0.073 & $\pm$0.048 & $\pm$0.122 \\
        \midrule

        \textbf{\textit{DiffuseKronA}} & \multirow{2}{*}{7.5 M} & 0.809 & 0.268 &0.651 \\
        $a_1=4$ && $\pm$0.086 & $\pm$0.055 & $\pm$0.142 \\
        \midrule

        \textbf{\textit{DiffuseKronA}} & \multirow{2}{*}{2.1 M} &0.815 &0.313 &0.649 \\ 
        $a_1=8$ && $\pm$0.074 & $\pm$0.024 & $\pm$0.139 \\
        \midrule

        \cellcolor[gray]{0.9} \textbf{\textit{DiffuseKronA}} & \cellcolor[gray]{0.9} & \cellcolor[gray]{0.9} 0.817 & \cellcolor[gray]{0.9} 0.301 & \cellcolor[gray]{0.9} 0.654 \\ 
        \cellcolor[gray]{0.9} $a_1=16$ & \multirow{-2}{*}{\cellcolor[gray]{0.9} 1.5 M} & \cellcolor[gray]{0.9} $\pm$0.078 & \cellcolor[gray]{0.9} $\pm$0.038 & \cellcolor[gray]{0.9} $\pm$0.127 \\
        \midrule 
        
        \textbf{\textit{LoRA-DreamBooth}} & \multirow{2}{*}{5.8 M} &0.807 & 0.288 & 0.635 \\
        $rank=4$ && $\pm$0.077 & $\pm$0.033 & $\pm$0.136 \\
        \midrule 
        
        \textbf{\textit{LoKr}} & \multirow{2}{*}{\textbf{1.36 M}} & 0.801  & 0.287  & 0.646 \\
        $f=-1$, $rank=8$ && $\pm$0.065 & $\pm$0.049 & $\pm$0.147 \\
        \midrule
        
        \textbf{\textit{LoKr}} & \multirow{2}{*}{14.9 M} & 0.812 & 0.277 & 0.639\\
        $f=8$ && $\pm$ 0.069& $\pm$0.042 & $\pm$0.111 \\
        \midrule
        
        \textbf{\textit{LoHA}} & \multirow{2}{*}{20.9 M} & 0.818 & 0.299 & 0.641\\
        $rank=4$ && $\pm$0.064  & $\pm$0.041 & $\pm$0.120\\
        \bottomrule
    \end{tabular}}

    \caption{\textbf{Quantitative comparison} of \textit{DiffuseKronA} with low-rank decomposition methods namely LoRA, LoKr, and LoHA in terms of the number of trainable parameters, text-alignment, and image-alignment scores. The scores are computed from the same set of images and prompts as depicted in~\cref{fig:comp_lora_lokr_loha}.}
    \label{tab:comp_lora_lokr_loha_quant}
\end{table}

\begin{table}[h]
    \centering
    \resizebox{0.48\textwidth}{!}{%
    \begin{tabular}{c|c|c|c|c}
    \toprule
        \textsc{Model} & \textsc{\# Parameters} ($\downarrow$) & \textsc{CLIP-I } ($\uparrow$) & \textsc{CLIP-T} ($\uparrow$) & \textsc{DINO} ($\uparrow$)\\\toprule
        
        \textit{\textbf{LoKr}} & \multirow{2}{*}{238.7 M} & 0.825 & 0.244 & 0.727\\
        $f=2$ && $\pm$0.037 & $\pm$0.024 & $\pm$0.036 \\
        \midrule
        
        \textbf{\textit{LoKr}} & \multirow{2}{*}{59.7 M} & 0.784 & 0.246 & 0.683\\
        $f=4$ && $\pm$ 0.063 & $\pm$0.030 & $\pm$0.051 \\
        \midrule
        
        \textbf{\textit{LoKr}} & \multirow{2}{*}{14.9 M} & 0.749 & 0.292 & 0.568\\
        $f=8$ && $\pm$0.067 & $\pm$0.064 & $\pm$0.075 \\
        \midrule

        \textbf{\textit{LoKr}} &\multirow{2}{*}{3.8 M} & 0.707 & 0.231  & 0.472\\
        $f=16$ && $\pm$ 0.121 & $\pm$0.025& $\pm$0.160 \\
        \midrule
        
        \cellcolor[gray]{0.9} \textbf{\textit{DiffuseKronA}} & \cellcolor[gray]{0.9} & \cellcolor[gray]{0.9} 0.806 & \cellcolor[gray]{0.9} 0.281 & \cellcolor[gray]{0.9} 0.653\\
        \cellcolor[gray]{0.9} $a_1=8$& \multirow{-2}{*}{\cellcolor[gray]{0.9} \textbf{2.1 M}} & \cellcolor[gray]{0.9} $\pm$ 0.028 & \cellcolor[gray]{0.9} $\pm$ 0.070 & \cellcolor[gray]{0.9} $\pm$ 0.045\\
        \bottomrule
    \end{tabular}}
    \caption{\textbf{Quantitative comparison} of \textit{DiffuseKronA} with varying factors (\emph{i.e.} 2, 4, 8, 16) of LoKr in terms of the number of trainable parameters, text-alignment, and image-alignment scores. The scores are computed from the same set of images and prompts as depicted in~\cref{fig:comp_lokr_default}.}
    \label{tab:comp_lokr_quant}
\end{table}

\clearpage

\begin{figure*}[!htbp]
    \centering
    \vspace*{-2mm}
    \includegraphics[width=1\textwidth]{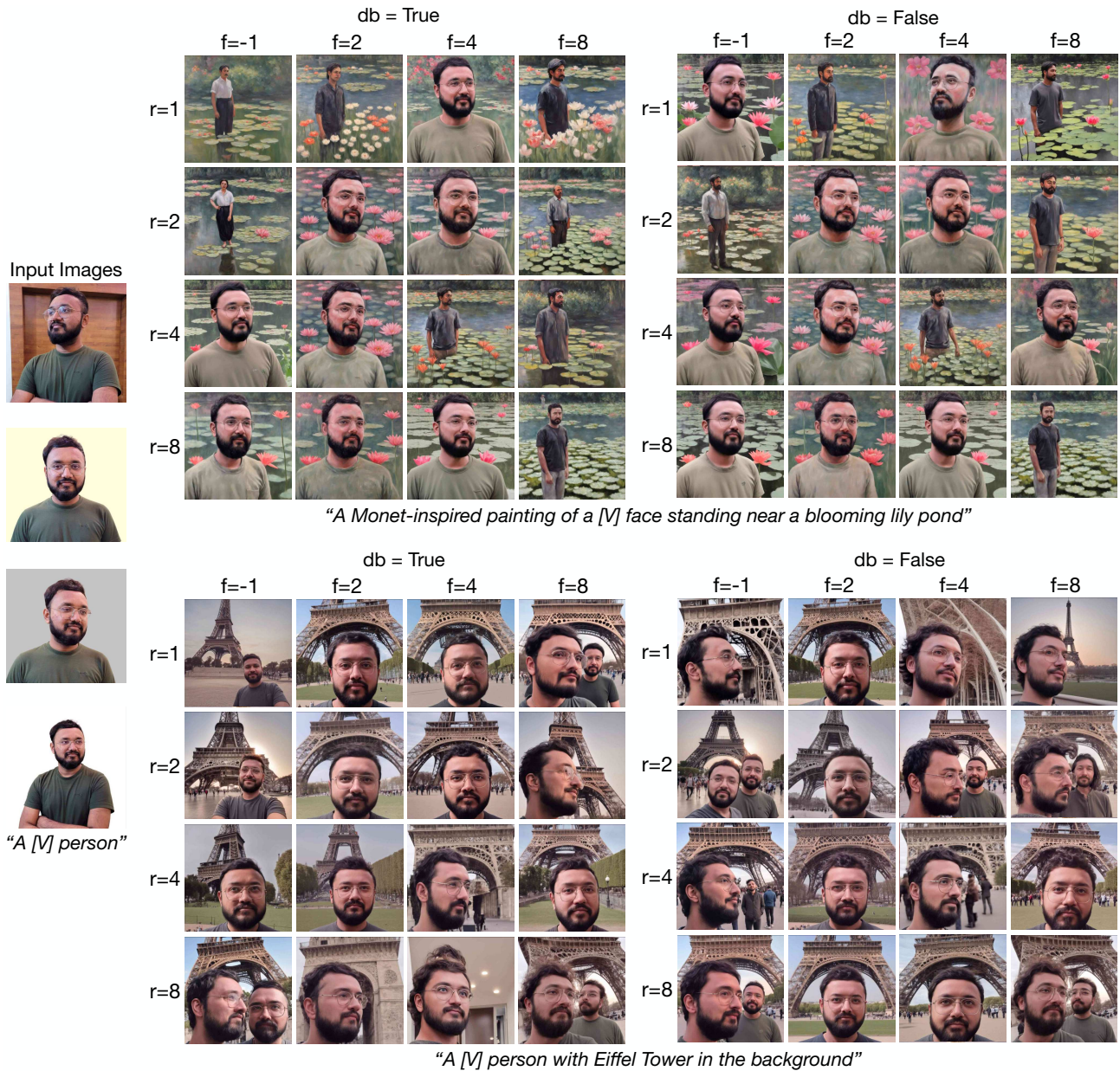}
    \vspace*{-7mm}
    \caption{\textbf{Ablation study on factor and rank} for LoKr using SDXL, with a learning rate of $1\times\mathrm{10}^{-3}$. We found that the optimal factor and rank are -1 and 8, respectively. We also experimented with db=True, which indicates further low-rank decomposition of both matrices $A$ and $B$, whereas db=False means only matrix $B$ is decomposed further. (\textbf{continued}..)}
\end{figure*}

\begin{figure*}[!htbp]
    \centering
    \ContinuedFloat
    \includegraphics[width=1\textwidth]{supple_images/Latest_learning_rate_fix_2.pdf}
    \caption{\textbf{Ablation study on factor and rank} for LoKr using SDXL, with a learning rate of $1\times\mathrm{10}^{-3}$. We found that the optimal factor and rank are -1 and 8, respectively. We also experimented with db=True, which indicates further low-rank decomposition of both matrices $A$ and $B$, whereas db=False means only matrix $B$ is decomposed further. (\textbf{continued}..)}
\end{figure*}

\begin{figure*}[!htbp]
    \centering
    \ContinuedFloat
    \includegraphics[width=1\textwidth]{supple_images/Latest_learning_rate_fix_3.pdf}
    \caption{\textbf{Ablation study on factor and rank} for LoKr using SDXL, with a learning rate of $1\times\mathrm{10}^{-3}$. We found that the optimal factor and rank are -1 and 8, respectively. We also experimented with db=True, which indicates further low-rank decomposition of both matrices $A$ and $B$, whereas db=False means only matrix $B$ is decomposed further. (\textbf{end})}
  \label{fig:ablation_lokr_factor_rank}
\end{figure*}

\begin{figure*}[!htbp]
    \centering
    \includegraphics[width=1\textwidth]{supple_images/Laest_fixed_factor_1.pdf}
    \caption{\textbf{Ablation study on factor and learning rate} for LoKr using SDXL, with a fixed factor of -1. We found that the optimal learning rate and rank are $1\times\mathrm{10}^{-3}$ and 8, respectively. We also experimented with db=True, which indicates further low-rank decomposition of both matrices $A$ and $B$, whereas db=False means only matrix $B$ is decomposed further. (\textbf{continued}..)}
\end{figure*}

\begin{figure*}[!htbp]
    \centering
    \ContinuedFloat
    \includegraphics[width=1\textwidth]{supple_images/Laest_fixed_factor_2.pdf}
    \caption{\textbf{Ablation study on factor and learning rate} for LoKr using SDXL, with a fixed factor of -1. We found that the optimal learning rate and rank are $1\times\mathrm{10}^{-3}$ and 8, respectively. We also experimented with db=True, which indicates further low-rank decomposition of both matrices $A$ and $B$, whereas db=False means only matrix $B$ is decomposed further. (\textbf{continued}..)}
\end{figure*}

\begin{figure*}[!htbp]
    \centering
    \ContinuedFloat
    \includegraphics[width=1\textwidth]{supple_images/Laest_fixed_factor_3.pdf}
    \caption{\textbf{Ablation study on factor and learning rate} for LoKr using SDXL, with a fixed factor of -1. We found that the optimal learning rate and rank are $1\times\mathrm{10}^{-3}$ and 8, respectively. We also experimented with db=True, which indicates further low-rank decomposition of both matrices $A$ and $B$, whereas db=False means only matrix $B$ is decomposed further. (\textbf{end})}
  \label{fig:ablation_lokr_lr}
\end{figure*}

\begin{figure*}[!htbp]
  \centering
    \includegraphics[width=0.73\textwidth]{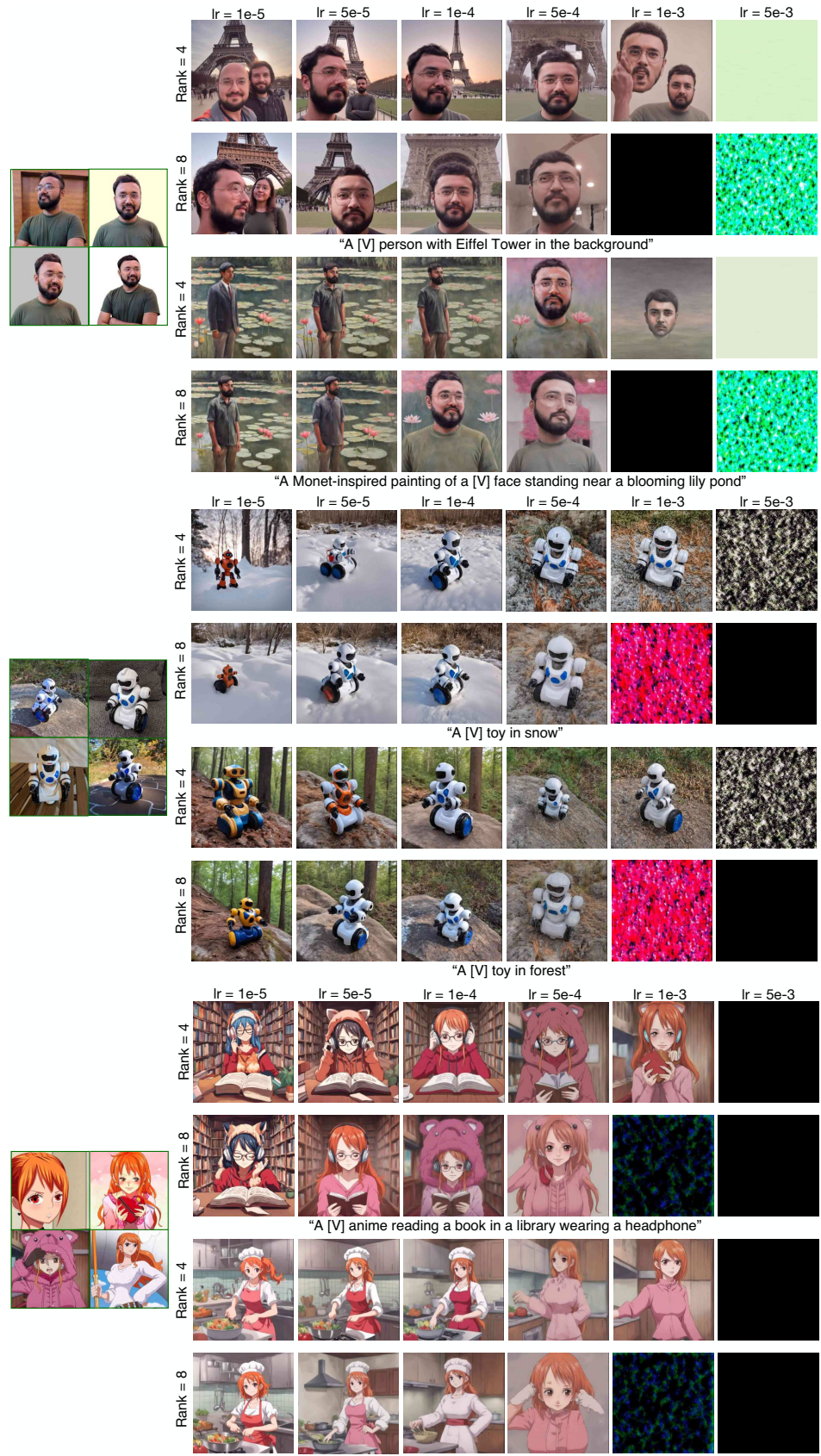}
  \caption{Ablation study on learning rate and rank for LoHA using SDXL. We found the optimal learning rate and rank to be $1\times\mathrm{10}^{-4}$ and 4, respectively.}
  \label{fig:ablation_loha}
\end{figure*}

\clearpage
\section{Comparison with state-of-the-arts}
\myparagraph{Qualitative Comparison.}
\label{sec:comparison}
In this section, we extend upon \hyperlink{4.4}{4.4} of the main paper, comparing \textit{DiffuseKronA} with State-of-the-art text-to-image personalization models including DreamBooth, LoRA-DreamBooth, SVDiff, Textual Invention, and Custom Diffusion.

(\textbf{1}) Textual Inversion~\cite{gal2022textual} is a fine-tuning method that optimizes a placeholder embedding to reconstruct the training set of subject images. Learning a new concept requires 3,000 steps, which takes around 30 minutes on an A100 GPU~\cite{blip_diffusion}.

(\textbf{2}) DreamBooth~\cite{dreambooth} refines the entire network through additional preservation loss as a form of regularization, leading to enhancements in visual quality that exhibit promising results. Updating DreamBooth for a new concept typically requires about 6 minutes on an A100 GPU~\cite{blip_diffusion}.

(\textbf{3}) LoRA-DreamBooth~\cite{cloneofsimo} explores low-rank adaptation for parameter-efficient fine-tuning attention-weight matrices of the text-to-image diffusion model. Fine-tuning LoRA-DreamBooth for a new concept typically takes about 5 minutes on a single 24GB NVIDIA RTX-3090 GPU.

\begin{figure*}[!ht]
  \centering
    \includegraphics[width=1\textwidth]{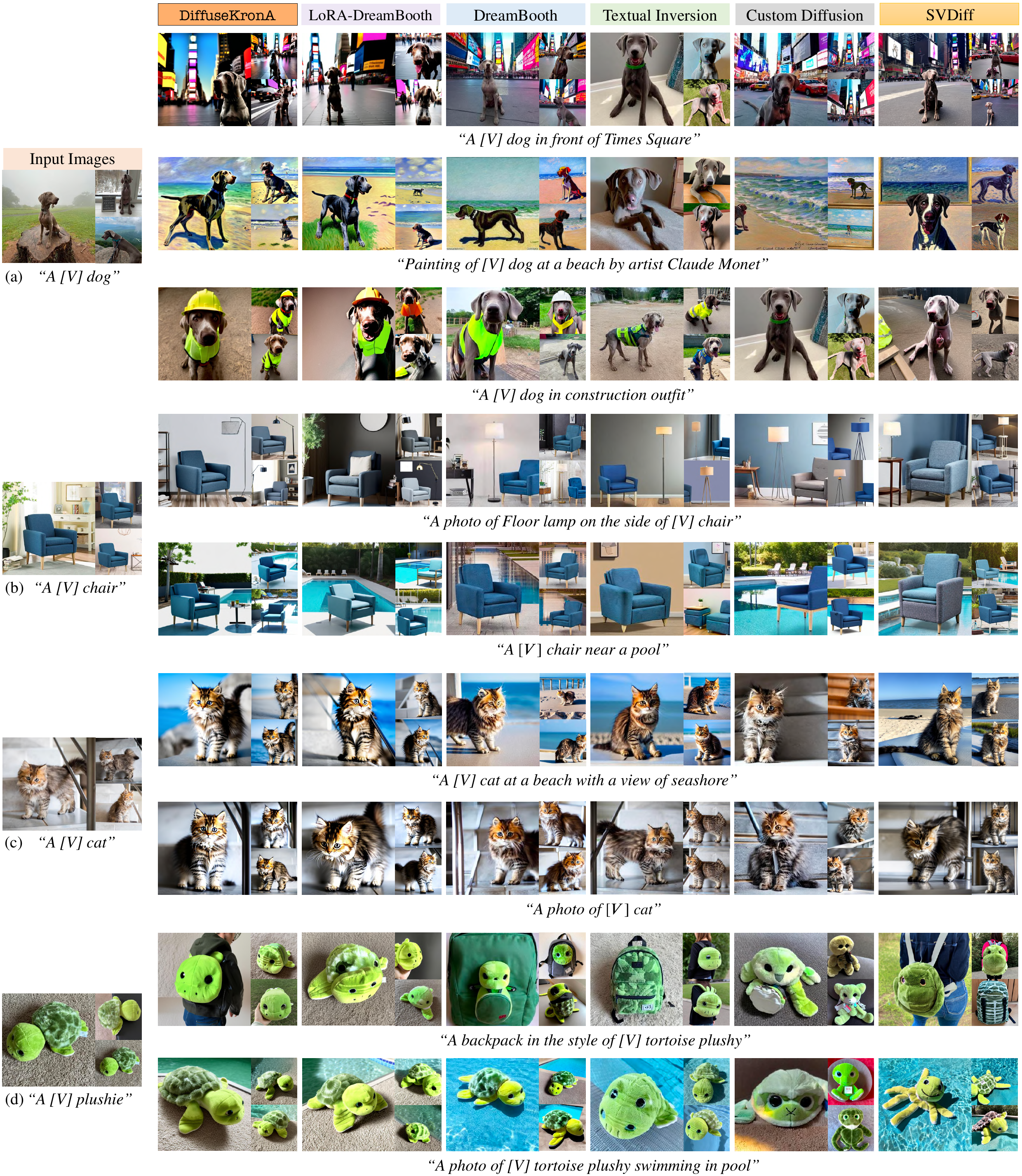}
  \caption{\textbf{Qualitative comparison} between generated images by \textit{DiffuseKronA}, LoRA-DreamBooth, Textual Inversion, DreamBooth, and Custom Diffusion. Notably, our methods’ results are generated considering $a_2 = 8$. We maintained the original settings of all these methods and used the SD CompVis-1.4~\cite{compvis} variant to ensure a fair comparison.}
  \label{fig:comparison_sd}
\end{figure*}

(\textbf{4}) SVDiff~\cite{han2023svdiff} involves fine-tuning the singular values of the weight matrices, leading to a compact and efficient parameter space that reduces the risk of overfitting and language drifting. It took around 15 minutes on a single 24GB NVIDIA RTX-3090 GPU\footnote{SVDiff did not release official codebase, we used open-source \href{https://github.com/mkshing/svdiff-pytorch}{code} for SVDiff results in~\cref{fig:comparison_sd}.}.

(\textbf{5}) Custom diffusion~\cite{custom_diffusion} involves selective fine-tuning of weight matrices through a conditioning mechanism, enabling parameter-efficient refinement of diffusion models. This approach is further extended to encompass multi-concept fine-tuning. The fine-tuning time of Custom diffusion is around 6 minutes on 2 A100 GPUs.

\myparagraph{Qualitative Comparison.}
\textit{DiffuseKronA} consistently produces images closely aligned with the input images and consistently integrates features specified in the input text prompt. The enhanced fidelity and comprehensive comprehension of the input text prompts can be attributed to the structure-preserving capability and improved expressiveness facilitated by Kronecker product-based adaptation. The images generated by LoRA-DreamBooth are not of high quality and demand extensive experimentation for improvement, as depicted in ~\cref{fig:comparison_sd}. As depicted in the figure, \textit{DiffuseKronA} not only generates well-defined images but also has a better color distribution as compared to Custom Diffusion.

\section{Practical Implications}

\begin{itemize}
    \item Content Creation: It can be used to generate photorealistic content from text prompts.
    \item Image Editing and In-painting: The model can be used to edit images or fill in missing parts of an image. 
    \item Super-Resolution: It can be used to enhance the resolution of images. 
    \item Video Synthesis: The model can be used to generate videos from text prompts. 
    \item 3D Assets Production: It can be used to create 3D assets from text prompts. 
    \item Personalized Generation: The model can be used in personalized generation with DreamBooth fine-tuning. 
    \item Resource Efficiency: The model is resource-efficient and can be trained with limited resources. 
    \item Model Compression: The model allows for architectural compression, reducing the number of parameters, MACs per sampling step, and latency.
\end{itemize}

\begin{figure*}[!ht]
  \centering
    \includegraphics[width=1\textwidth]{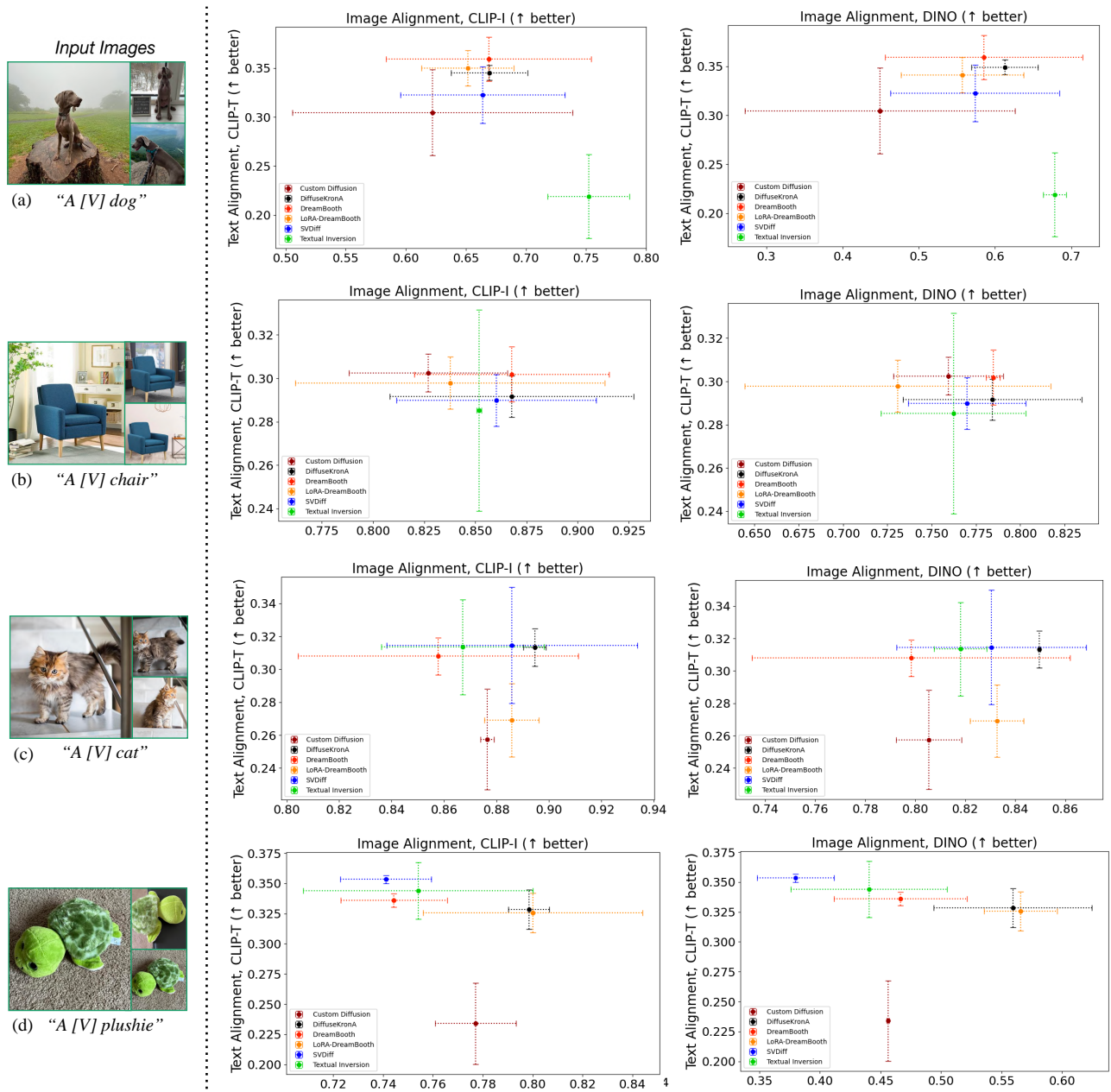}
  \caption{\textbf{Quantitative comparison} of \textit{DiffuseKronA} with SOTA on Text-Image Alignment. The scores are computed from the same set of images and prompts as depicted in~\cref{fig:comparison_sd}.}
  \label{fig:comparison_sd_quant}
\end{figure*}

\end{document}